\documentclass[11pt,letterpaper]{article}
\usepackage[margin=1in]{geometry}
\usepackage{adjustbox}
\ifdefined\pdfinfoomitdate \pdfinfoomitdate=1 \fi
\ifdefined\pdfsuppressptexinfo \pdfsuppressptexinfo=-1 \fi
\ifdefined\pdftrailerid \pdftrailerid{} \fi
\title{When Does Pooling Pay? Credibility and Resolution\\
under Forgetting in Intermittent-Demand Forecasting}
\author{Zong-Han Bai\thanks{Both authors contributed equally to this research.}\\
National Tsing Hua University\\
\texttt{hu110048138@gapp.nthu.edu.tw}
\and Po-Yen Chu\footnotemark[1]\\
Stanford University\\
\texttt{poyenchu@stanford.edu}}
\date{}
\usepackage{bold-extra}       
\usepackage{amsmath,amssymb,amsthm}
\usepackage{booktabs}
\usepackage{graphicx}
\usepackage[expansion=false]{microtype}
\usepackage[numbers,sort&compress]{natbib}
\usepackage{xcolor}
\usepackage{float}
\usepackage[colorlinks=true,linkcolor=blue,citecolor=blue,urlcolor=blue]{hyperref}
\usepackage{tikz}
\usetikzlibrary{arrows.meta,positioning,calc}

\theoremstyle{definition}
\newtheorem{proposition}{Proposition}
\newtheorem{corollary}{Corollary}
\newtheorem{lemma}{Lemma}

\newcommand{\EBB}{\textsc{Ebb}}
\newcommand{\E}{\mathbb{E}}
\newcommand{\Var}{\mathrm{Var}}
\newcommand{\ind}[1]{\mathbf{1}\{#1\}}
\newcommand{\best}[1]{\textbf{#1}}
\newcommand{\secondbest}[1]{\underline{#1}}

\hypersetup{pdftitle={When Does Pooling Pay? Credibility and Resolution under Forgetting in Intermittent-Demand Forecasting},pdfauthor={Zong-Han Bai and Po-Yen Chu}}

\usepackage[section]{placeins}
\begin{document}

\maketitle

\begin{abstract}
Forecasting many sparse series requires two choices: how much of each
series' own past to retain, and how much to borrow from other series.
We show that when one exponential recency operator is applied to item-
and group-level statistics alike in a hierarchical empirical-Bayes
hurdle model, the two become one decision: forgetting preserves the
leverage of a shared prior while raising the noise floor against which
fine cross-series structure must be resolved. We formalize this
two-sided effect as a fitting-window diagnostic, a credibility bound
and a resolution condition for candidate pools, and derive from it an
ex ante screen for regimes where refinement cannot pay. On five public
intermittent-demand panels comprising over 19{,}000 series, where
item-level evidence is sparsest, the diagnostic is right in both
directions. Where truncation lowers credibility the shared prior pays:
switching it off costs $5\%$ and $12\%$ at the shortest histories on
the two panels it helps most, against about $1\%$ at full length, and
the gain sits in the occurrence block. On four panels the diagnostic
reports at most $0.6\%$ of room for refinement and realized gains are
no larger; on the one panel with room the learned mixture does not
collect it, which locates the open problem in the partition objective.
The resulting forecaster, \EBB{}, stays within about $3\%$ of a
per-series Gaussian-process specialist at fixed origin and within $1\%$
under walk-forward evaluation where both run, at two to three orders of
magnitude lower cost.
\end{abstract}

\section{Introduction}\label{sec:intro}

Forecasting a collection of related time series requires a basic choice:
how much should each series rely on its own history, and how much should
it borrow from others? In practice, this locality--globality choice is
resolved empirically by fitting competing local and pooled models
\citep{monteromanso2021,januschowski2020,hewamalage2022}. We ask how
much of this choice can instead be diagnosed before fitting increasingly
refined pooling structures. The question is especially sharp for
intermittent demand, where long runs of zeros leave little item-level
evidence, and it is entangled with a second choice: how much historical
evidence to retain when the process may drift or become obsolete.

These two choices are one decision. With exponential forgetting, an
item's effective history stays bounded, so a shared prior retains
non-vanishing leverage even after a long observation sequence; the same
discounting raises the noise floor against which cross-series
heterogeneity must be estimated. Stronger forgetting therefore makes
borrowing strength more consequential while making fine pooling
structure harder to resolve. The coupling is easy to miss because the
two are usually set apart: pooled models estimate their prior on the
full history, and forgetting, where it is used, is applied to each
item's own statistics. Once one operator governs both, the same
discounted evidence determines item-level credibility and the
information available for estimating the shared prior, and both effects
leave measurable signatures in the fitting window. Our answer to the
title question is correspondingly two-sided: low credibility creates
the leverage for pooling to matter, and finer pooling pays only when
the discounted data also contain resolvable cross-series structure.

We study the coupling in the simplest model that exhibits it.
\EBB{}\footnote{Empirical Bayes with Bounded memory: under the recency
operator an item's effective history stays bounded
(Section~\ref{sec:theory}), so evidence ebbs rather than accumulates.}
is a recency-weighted hierarchical empirical-Bayes forecaster with
Beta--Binomial occurrence and log-normal size blocks, in which the same
recency weights form each item's posterior and estimate the group-level
hyperparameters. It crosses a small grid of discount factors with three
pooling candidates---a single global pool, the ADI/CV$^2$ taxonomy, and
BIC-selected mixture partitions---and selects the pair on a
leakage-safe chronological validation tail. Apart from the mixture
search, estimation is deterministic and linear in the number of
observations, and online updating touches only sufficient statistics.

\paragraph{Contributions.}
\textbf{(1) A two-sided theory of pooling under forgetting.}
Proposition~\ref{prop:twosided} bounds the credibility an item retains
under forgetting and characterizes, under a Gaussian moment
approximation, the probability that a candidate pool's variance
component collapses. Corollary~\ref{cor:resolution} converts the latter
into a fitting-window condition for whether finer group structure can be
stably resolved (Section~\ref{sec:theory}).

\textbf{(2) A diagnostic that is right in both directions.}
Where item-level credibility is low the shared prior pays: switching it
off within \EBB{} costs about $1\%$ of mean SPL on Auto and RAF at full
length and $5\%$ and $12\%$ with six periods of history, the gain sits
in the occurrence block, and it accounts for most of \EBB{}'s advantage
over the per-series TweedieGP specialist there. Where there is no room
to refine, none is realized: a controlled surface over leverage and
group separation places four panels where even the true partition is
worth at most $0.6\%$, and their realized gains are no larger. Carparts
alone has room, $1.7$--$2.0\%$, which the learned mixture fails to
recover in data and in simulation alike, locating the open problem in
the partition objective. On M5 forgetting is worth an order of
magnitude more than refinement, and a pre-fit screen built from
single-pool quantities rules out, in held-out simulation, regimes where
refinement cannot pay (Section~\ref{sec:pays}).

\textbf{(3) An efficient forecaster whose wins and losses the
diagnostic orders.}
At fixed origin \EBB{} is statistically indistinguishable from or
better than the per-series TweedieGP specialist on four of five panels
and trails it by $2.9\%$ on M5, and likewise under walk-forward
evaluation wherever both run, at two to three orders of magnitude lower
cost. Against a default DeepAR, occurrence credibility orders the
outcome on all five panels: \EBB{} leads by $7.8\%$ and $2.8\%$ where
credibility is lowest, ties in the middle, and trails by $4\%$ and
$9$--$18\%$ on the two daily panels where it is highest and the prior
has no leverage to offer. The quantity that says when pooling pays thus
also indicates, before any model is trained, where a seconds-scale
empirical-Bayes fit is enough (Section~\ref{sec:accuracy}).

We also document two evaluation cautions for sparse forecasting:
absolute-error metrics can reward the all-zero forecast, and static
conformal wrappers understate adaptive ones
(Section~\ref{sec:cautions}).

\section{Related Work}\label{sec:related}

\textbf{Locality, globality, and cross-series pooling.}
Forecasting a collection of related series spans a continuum from local
models, which estimate each series separately, to global models, which
share statistical strength across a collection
\citep{januschowski2020}. \citet{monteromanso2021} show that a global
model can always match local ones in principle, and
\citet{hewamalage2022} map when it does so in practice; in both, the
choice is settled by fitting and comparing. Recent work learns
increasingly flexible forms of cross-series specialization: in
hierarchical forecasting, \citet{cini2024} learn a graph-based
aggregation hierarchy aligned with the forecasting objective, while
Moirai-MoE \citep{liu2025moirai} uses token-level sparse-expert routing
to specialize a time-series foundation model. Our object is simpler, an explicit exchangeability partition
whose members share a prior, and the restriction is what makes the
interaction between pooling and finite information tractable: the value
of refinement can be screened before the finer partition is fitted.

\textbf{Intermittent-demand forecasting.}
Intermittent demand provides a particularly sparse setting in which this
locality--globality decision is consequential. Croston's decomposition
\citep{croston1972}, its SBA correction
\citep{syntetos2005accuracy}, and TSB \citep{teunter2011} estimate
occurrence and positive size locally; TSB additionally discounts the
occurrence estimate under obsolescence. ADIDA
\citep{nikolopoulos2011} and IMAPA \citep{kourentzes2016} reduce
intermittency through temporal aggregation. Likelihood-based alternatives
include inventory count models \citep{snyder2012}, iETS
\citep{svetunkov2023iets}, Tweedie Gaussian processes
\citep{damato2025}, and deep renewal processes
\citep{turkmen2019}. Hierarchical approaches instead borrow strength
across related series: \citet{chapados2014} pools count series through a
supplied product hierarchy, \citet{seeger2016} fit per-item state-space
models with shared feature effects at scale, and \citet{pitkin2024}
develops a hierarchical hurdle model for retail. In each, the pooling
structure is fixed in advance and the temporal dynamics are specified
separately from it. \EBB{} studies their interaction explicitly by
applying the same recency operator to the item-level evidence and to
the empirical-Bayes statistics that define the shared prior, then
selecting forgetting and pooling jointly.

\textbf{Discounting, credibility, and model selection.}
Exponential discounting is classical in Bayesian dynamic models
\citep{westharrison1997}, and power priors down-weight historical
likelihood contributions in a related way \citep{ibrahimchen2000}.
Likewise, $\lambda_i=n_i/(n_i+\phi_g)$ is the classical actuarial
credibility form \citep{buhlmann1967}, restated in
Appendix~\ref{app:supporting}. Every ingredient used here is classical:
credibility weights, exponential discounting, empirical-Bayes
hyperparameters, validation-based selection. What is new is the
coupling and its consequence: one recency operator determines both the
effective item-level evidence and the empirical-Bayes evidence used to
estimate the shared prior, which ties how much a prior can move a
forecast to how well its structure can be learned
(Section~\ref{sec:theory}).

\section{\EBB{}: Recency-Weighted Hierarchical EB}\label{sec:method}

\EBB{} couples two choices inside one occurrence--size model: a recency
discount $w$ determines how much historical evidence remains informative,
and a partition $g$ determines which items share an empirical-Bayes prior.
The key construction is that the \emph{same} recency-weighted sufficient
statistics are used both to estimate the group prior and to update each
item posterior. This shared representation makes forgetting and pooling
jointly selectable and exposes the credibility quantities analyzed in
Section~\ref{sec:theory} (Figure~\ref{fig:pipeline}).

\begin{figure}[H]
\centering
\begin{tikzpicture}[
  font=\scriptsize,
  box/.style={draw=black!55, rounded corners=1.5pt, line width=0.4pt,
              inner sep=2.2pt, align=center, minimum height=5mm,
              text width=3.6cm},
  hub/.style={box, fill=black!6, line width=0.6pt, text width=4.6cm},
  fc/.style={box, fill=blue!5},
  dg/.style={box, fill=red!5},
  arr/.style={-{Stealth[length=1.6mm]}, line width=0.5pt, black!65},
  lab/.style={font=\tiny, text=black!60, inner sep=1pt, align=center}
]
\node[hub] (S) at (0,0) {discounted statistics $S_i(w)$ \eqref{eq:discounted-stats}\\
  one recency operator, used twice};
\node[lab, above=1.2mm of S] {series $y_{i,t}$, discount $w$, pool $r\in\mathcal R$};
\draw[arr] ($(S.north)+(0,2.8mm)$) -- (S.north);

\node[fc] (prior) at (-2.05,-1.05) {group prior $\widehat\eta_g(w)$ per pool};
\node[fc, below=2.2mm of prior] (post) {item posterior, weight $\lambda_i(w)$ \eqref{eq:credibility-pair}};
\node[fc, below=2.2mm of post] (sel) {quantiles; validation tail selects $(\hat r,\hat w)$};

\node[dg] (res) at (2.05,-1.05) {single-pool fit: $\hat\tau_g^2$, $\hat z_g$, $\lambda_i$};
\node[dg, below=2.2mm of res] (cor) {resolution condition (Cor.~\ref{cor:resolution})};
\node[dg, below=2.2mm of cor] (scr) {pre-fit screen: room / no room for refinement};
\draw[arr] (S.south) -- ++(0,-1.5mm) -| (prior.north);
\draw[arr] (S.south) -- ++(0,-1.5mm) -| (res.north);
\draw[arr] (prior) -- (post);
\draw[arr] (post) -- (sel);
\draw[arr] (res) -- (cor);
\draw[arr] (cor) -- (scr);
\node[lab, below=0.8mm of sel] {forecast path:\\ leverage persists (Prop.~\ref{prop:twosided}i)};
\node[lab, below=0.8mm of scr] {diagnostic path:\\ resolution falls (Prop.~\ref{prop:twosided}ii)};
\end{tikzpicture}
\caption{\textbf{One fit, two readings.} The discounted sufficient
statistics $S_i(w)$ enter both the group prior and the item posterior
(forecast path, left) and, from the single-pool fit alone, yield the
credibility and resolution quantities of Section~\ref{sec:theory}
(diagnostic path, right), before any finer partition is fitted.}
\label{fig:pipeline}
\end{figure}

\FloatBarrier
\subsection{Occurrence--size model}\label{sec:base}

For item $i$, let $z_{i,t}=\ind{y_{i,t}>0}$ and
$\ell_{i,t}=\log y_{i,t}$ when $z_{i,t}=1$. Conditional on group
membership $g_i=g(i)$, we use the hurdle model
\begin{equation}
\begin{aligned}
\pi_i &\sim \mathrm{Beta}(\alpha_{g_i},\beta_{g_i}), &
z_{i,t}\mid\pi_i &\sim \mathrm{Bern}(\pi_i),\\
\mu_i &\sim \mathcal{N}(\mu_{0,g_i},\tau^2_{g_i}), &
\sigma_i^2 &\sim \mathrm{ScaleInv}\text{-}\chi^2(\nu,s^2_{0,g_i}),\\
\ell_{i,t} &\;\big|\; z_{i,t}{=}1,\mu_i,\sigma_i^2
\sim \mathcal{N}(\mu_i,\sigma_i^2).\span\span
\end{aligned}
\label{eq:model}
\end{equation}
Here $s^2_{0,g}$ is parameterized so that the prior mean of
$\sigma_i^2$ equals the group estimate $\sigma_g^2$, while $\nu$ is
fixed. Group hyperparameters are estimated by empirical Bayes:
Beta--Binomial marginal likelihood for occurrence, and an intercept-only
random-effects fit with conjugate variance shrinkage for positive
log-size. Appendix~\ref{app:implementation} gives the numerical
objectives and fallback rules.

Without forgetting, inference depends on four per-item sufficient
statistics,
\[
S_i(1)=(n_i,c_i,u_i,v_i)
=\sum_{t=1}^{T_i}
  \bigl(1,z_{i,t},z_{i,t}\ell_{i,t},z_{i,t}\ell_{i,t}^2\bigr).
\]
Let $n_i^+=c_i$, $\bar p_i=c_i/n_i$, and
$\bar\ell_i=u_i/n_i^+$. The posterior means for occurrence and positive
log-size have parallel credibility forms,
\begin{align}
\hat\pi_i
&=\lambda_i^{(o)}\bar p_i
 +(1-\lambda_i^{(o)})\mu_{\pi,g_i},
&\lambda_i^{(o)}
&=\frac{n_i}{n_i+\phi_{g_i}},\nonumber\\
\hat\mu_i
&=\lambda_i^{(+)}\bar\ell_i
 +(1-\lambda_i^{(+)})\mu_{0,g_i},
&\lambda_i^{(+)}
&=\frac{n_i^+}{n_i^++\kappa_{i,g_i}},
\label{eq:credibility-pair}
\end{align}
where $\phi_g=\alpha_g+\beta_g$, $\mu_{\pi,g}=\alpha_g/\phi_g$, and
$\kappa_{i,g}=\hat\sigma_i^2/\tau_g^2$. Thus, for these posterior
locations, changing the partition changes the prior target and the
strength with which each item is shrunk toward it. The resolution
analysis of Section~\ref{sec:resolution} replaces the item-level ratio
by the group surrogate $\kappa_g=\sigma_g^2/\tau_g^2$; that is the
quantity entering Proposition~\ref{prop:twosided}.

The resulting plug-in predictive distribution is zero-inflated
log-normal:
\begin{equation}
\begin{aligned}
\Pr(Y_i=0\mid\mathcal D)&=1-\hat\pi_i,\\
\log Y_i\mid(Y_i>0,\mathcal D)
&\sim\mathcal N(\hat\mu_i,\hat\sigma_i^2+V_i),
\end{aligned}
\label{eq:predictive-law}
\end{equation}
where $V_i$ is posterior uncertainty in $\mu_i$. Predictive quantiles
are analytic. For point evaluation we use
$\hat y_i=\hat\pi_i\exp(\hat\mu_i+\hat\sigma_i^2/2)$; this conditional
plug-in mean is distinct from the mean of
\eqref{eq:predictive-law}, which additionally integrates $V_i$.

\FloatBarrier
\subsection{Shared recency weighting}\label{sec:discount}

For $w\in(0,1]$, we replace $S_i(1)$ by
\begin{equation}
S_i(w)=\sum_{t=1}^{T_i}w^{T_i-t}
  \bigl(1,z_{i,t},z_{i,t}\ell_{i,t},z_{i,t}\ell_{i,t}^2\bigr).
\label{eq:discounted-stats}
\end{equation}
These fractional counts define the corresponding weighted likelihoods.
Crucially, $S_i(w)$ is used at both levels of the hierarchy:
\begin{equation}
\begin{aligned}
\widehat\eta_g(w)
&=\arg\max_{\eta_g}
  \sum_{i:g_i=g}\log m\{S_i(w)\mid\eta_g\},\\
q_i^{(w)}(\theta_i)
&=q_i\!\left(
  \theta_i\mid S_i(w),\widehat\eta_{g_i}(w)
  \right),
\end{aligned}
\label{eq:shared-recency}
\end{equation}
where $\eta_g$ collects the group hyperparameters and $q_i$ denotes the
item posterior. The shared prior therefore adapts to recent
cross-sectional evidence rather than remaining a stationary prior
applied to discounted item data. Setting $w=1$ recovers the stationary
estimator. The credibility forms \eqref{eq:credibility-pair} carry over
with $n_i(w)$, $n_i^+(w)$, $\phi_g(w)$, and $\kappa_{i,g}(w)$ in place
of their undiscounted counterparts; these discounted weights are the
quantities analyzed in Section~\ref{sec:resolution}.

\FloatBarrier
\subsection{Candidate pooling structures}\label{sec:mixture}

We consider a deliberately small family $\mathcal R$ of pooling
structures, ordered from coarse to fine. The \emph{global} candidate
uses one prior for the entire panel. The \emph{taxonomy} candidate uses
the four classical ADI/CV$^2$ demand-pattern classes
\citep{syntetos2005categorization}. The \emph{learned} candidate uses
hard assignments from a mixture of occurrence--size prior components,
with $K\in\{1,\dots,6\}$ chosen by BIC. The mixture is fit by EM with
analytic responsibility updates and deterministic initialization;
details are given in Appendix~\ref{app:demoted},
Table~\ref{tab:pool-candidates}. Every data-dependent label is
constructed from fitting data only.

\FloatBarrier
\subsection{Leakage-safe joint selection}\label{sec:select}

Each training series is split chronologically into a fitting head
$\mathcal H_i$ (first $80\%$) and validation tail $\mathcal V_i$ (last
$20\%$). For every $(r,w)\in\mathcal R\times\mathcal W$, where
\[
\mathcal W=\{1,.999,.997,.995,.99,.98,.95,.90\},
\]
the partition for $r$, the discounted EB prior, and the item posteriors
are fit using only $\mathcal H=\cup_i\mathcal H_i$. With
$n_V=\sum_i|\mathcal V_i|$ validation observations, the candidate is
scored by
\begin{equation}
\widehat R(r,w)
=\frac{1}{n_V|\mathcal Q|}\sum_{i}
  \sum_{t\in\mathcal V_i}\sum_{q\in\mathcal Q}
  \frac{\rho_q\bigl(y_{i,t},
  \widehat Q_{i,t}^{(r,w)}(q)\bigr)}{a_i},
\label{eq:selection-risk}
\end{equation}
with $\mathcal Q=\{.5,.75,.9\}$ and pinball loss $\rho_q$, a proper
scoring rule \citep{gneiting2007} whose upper quantiles matter for
asymmetric inventory decisions \citep{kolassa2016}.
The scale $a_i$ is the mean absolute demand on the fitting head, with a
panel-level fallback for an all-zero head, so high-volume series do not
dominate selection; every validation observation carries equal weight.

We select
\[
(\hat r,\hat w)=\arg\min_{r,w}\widehat R(r,w).
\]
The chosen partition rule is then relearned on the complete training
window at $\hat w$, followed by one final EB fit before external
evaluation. Thus the validation tail selects a \emph{construction rule}
and discount, rather than freezing labels learned on only part of the
training data. External targets are never used in selection. The grid
contains 24 forecast candidates. The same fitting-window statistics
yield, for every candidate, the credibility weights and resolution
statistic of Section~\ref{sec:resolution} at no extra cost
(Appendix~\ref{app:demoted}).

\FloatBarrier
\subsection{Forecasts and computation}\label{sec:complexity}

Selection uses the analytic plug-in quantiles from
\eqref{eq:predictive-law}. For reported probabilistic forecasts, we also
average quantile estimates across $B=20$ cross-item bootstrap refits of
the group hyperparameters, a quantile ensemble. The predictive law is
horizon-invariant: absent new observations, the same distribution is
issued for every step of the evaluation span, and under walk-forward
evaluation it changes only through the sufficient-statistic update
after each revealed block (Section~\ref{sec:setup}).

One pass per $w$ computes the discounted statistics in $O(\sum_i T_i)$;
each EB fit is a low-dimensional deterministic optimization within
groups, mixture EM costs $O(NK)$ per iteration, and prediction is
$O(NQ)$ for $Q$ quantiles.

\section{Resolution under Forgetting}\label{sec:theory}

The shared recency operator in \eqref{eq:shared-recency} induces a
tension between \emph{leverage} and \emph{resolution}. At the item level,
discounting caps the effective evidence, so the influence of a shared
prior need not vanish with a long history. At the group level, the same
cap raises the noise floor against which cross-series heterogeneity must
be estimated. Throughout, resolution means the ability to estimate the
variance component of a candidate pool under the working
empirical-Bayes model. Supporting results connecting
\EBB{} to TSB, drift tracking, finite-class selection, and classical
credibility are collected in Appendix~\ref{app:supporting}.

\FloatBarrier
\subsection{Persistent prior leverage}

Fix an item and suppress its subscripts. For $w<1$,
$n(w)=\sum_{t=1}^{T}w^{T-t}\le(1-w)^{-1}$, so the occurrence
credibility weight satisfies
\[
\lambda^{(o)}(w)
=\frac{n(w)}{n(w)+\phi(w)}
\le
\{1+(1-w)\phi(w)\}^{-1}<1,
\]
whereas without forgetting $\lambda^{(o)}(1)\to1$ as $T\to\infty$ under
fixed prior strength (Appendix~\ref{app:demoted},
Proposition~\ref{prop:active}). Forgetting thus keeps prior influence
bounded away from zero even for long-lived series. This is a statement
about \emph{leverage}, not benefit: a prior that retains influence can
improve or worsen the forecast depending on its target.

\FloatBarrier
\subsection{Variance-component resolution}\label{sec:resolution}

The other side appears when the shared prior itself must be estimated.
All quantities in this subsection are the discounted ones of
Section~\ref{sec:discount}; we suppress the argument $(w)$. Consider
the positive-size block within group $g$ under the Gaussian working
approximation,
\[
\bar\ell_i\mid\mu_i \sim \mathcal N(\mu_i,s_i),
\quad
s_i=\frac{\sigma_g^2}{n_i^+},
\quad
\mu_i\sim\mathcal N(\mu_{0,g},\tau_g^2),
\]
so that $\lambda_i^{(+)}=n_i^+/(n_i^++\kappa_g)$ with
$\kappa_g=\sigma_g^2/\tau_g^2$. Let $m_g$ be the number of items in the
group, $S_g^2$ the unbiased sample variance of their discounted means,
$\bar s_g=m_g^{-1}\sum_{i\in g}s_i$ the working noise floor, and
$\hat\tau_g^2=\max(0,S_g^2-\bar s_g)$ the moment estimate of the
variance component, which initializes the REML fit \EBB{} uses and
shares its boundary when the $s_i$ are equal.

\begin{proposition}[Two-sided effect of forgetting]
\label{prop:twosided}
For the working model above and $w<1$:

\textbf{(i) Persistent leverage.}
$\lambda_i^{(+)}\le\{1+(1-w)\kappa_g\}^{-1}<1$.

\textbf{(ii) Reduced resolution.}
$\E[S_g^2]=\tau_g^2+\bar s_g$ exactly, and under the Gaussian moment
approximation to $S_g^2$,
\begin{equation}
\Pr(\hat\tau_g^2=0)\simeq\Phi(-z_g),
\qquad
z_g=\sqrt{\tfrac{m_g-1}{2}}\,\rho_g,
\label{eq:collapse-prob}
\end{equation}
with $\rho_g=\tau_g^2/(\tau_g^2+\bar s_g)$. Since $n_i^+\le(1-w)^{-1}$,
$\bar s_g\ge(1-w)\sigma_g^2$ and hence
$\rho_g\le\tau_g^2/\{\tau_g^2+(1-w)\sigma_g^2\}$. On the boundary event
$\hat\tau_g^2=0$ the plug-in variance ratio diverges and
$\lambda_i^{(+)}=0$: every item is shrunk fully to the group mean.
\end{proposition}

Equation~\eqref{eq:collapse-prob} exposes the resolution coordinate
$z_g$: resolution improves with heterogeneity relative to the noise
floor ($\rho_g$) and with pool size ($m_g$). Neither is known, but the
plug-in $\hat z_g=(S_g^2-\bar s_g)/\{\sqrt{2/(m_g-1)}\,S_g^2\}$ is
nonpositive exactly on the boundary estimate $\hat\tau_g^2=0$.

\begin{corollary}[Resolution condition]
\label{cor:resolution}
Fix a collapse probability $\delta\in(0,1/2)$ and let
$c_\delta=\sqrt2\,\Phi^{-1}(1-\delta)$. Under \eqref{eq:collapse-prob},
$\Pr(\hat\tau_g^2=0)\le\delta$ requires
\begin{equation}
\rho_g\sqrt{m_g-1}\ge c_\delta,
\quad\text{i.e.}\quad
\tau_g^2\ge\frac{c_\delta}{\sqrt{m_g-1}-c_\delta}\,\bar s_g
\label{eq:resolution-condition}
\end{equation}
whenever $\sqrt{m_g-1}>c_\delta$. Because $\bar s_g\ge(1-w)\sigma_g^2$,
the heterogeneity a pool must exhibit to be resolved grows at least in
proportion to $(1-w)\sigma_g^2$: forgetting raises the resolution
threshold.
\end{corollary}

The corollary makes the structural cost of refinement explicit. Moving
to finer pools reduces $m_g$ and the residual within-pool heterogeneity
$\tau_g^2$, while stronger forgetting raises $\bar s_g$; all three
changes reduce $z_g$ and push the variance-component estimate toward
its boundary.

\paragraph{What is structural and what is specific.}
Two parts of the argument do not depend on the estimator. The cap
$n_i^+\le(1-w)^{-1}$ on effective evidence, and with it the persistence
of leverage, holds for any credibility form built on discounted counts;
and the noise floor is a property of the discounted means themselves:
an exponentially weighted mean has sampling variance at least
$\sigma_g^2(1-w)/(1+w)$ whatever the length of the history
(Lemma~\ref{lem:floor}, Appendix~\ref{app:proofs}). What is specific to \EBB{} is the
collapse probability and the threshold $c_\delta$, which depend on the
moment estimator and the Gaussian approximation. Whether a resolvable
partition also improves forecasts depends further on between-group
separation and on how the partition is learned; Section~\ref{sec:pays}
tests that step directly. Proofs, the accuracy of the approximation
(Figure~\ref{fig:mechanism}), and a controlled test under known labels
are in Appendices~\ref{app:proofs} and~\ref{app:demoted}.

\section{Experiments}\label{sec:experiments}

\FloatBarrier
\subsection{Experimental design}\label{sec:setup}

\textbf{Panels.}
Two daily panels, \textbf{Online Retail} \citep{chen2015} (3{,}649
series) and a seed-controlled 5{,}000-series \textbf{M5} sample
\citep{makridakis2022}, and three monthly panels from
\citet{damato2025}, \textbf{Auto} (3{,}000), \textbf{Carparts}
(2{,}509), and \textbf{RAF} (5{,}000): over 19{,}000 series spanning
long daily histories and short, highly sparse monthly ones
(Appendix~\ref{app:datasets}).

\textbf{Comparators and protocols.}
Classical intermittent-demand methods (Croston, SBA, TSB, ADIDA,
IMAPA), AutoARIMA and AutoTheta, an autoregressive Tweedie GLM, iETS,
and the official per-series TweedieGP \citep{damato2025} at its released
defaults, the current per-series state of the art on these panels and
the source of the three monthly ones.
DeepAR \citep{salinas2020} is the neural global comparator: one untuned
model per panel with a negative-binomial output, trained on the
initialization window and never refitted, and run at fixed origin on
the monthly panels only (Appendix~\ref{app:deepar}).
Point forecasters receive
chronological split-conformal wrappers \citep{lei2018}; under
walk-forward evaluation we add adaptive conformal inference (ACI)
\citep{gibbs2021} on ADIDA and on \EBB{}, and a tuned-TSB control
receives \EBB{}'s selection budget. The fixed-origin experiment fits once on the initialization
window and scores the full evaluation span; strict walk-forward
evaluation reveals seven-day blocks on the daily panels and one-month
blocks on the monthly ones, after which \EBB{} updates its discounted
sufficient statistics and other methods refit per their protocol
(expensive baselines every four blocks on Online Retail; TweedieGP not
run on M5). DeepState and Chronos-Bolt ran on subsets under other
protocols and are reported in Appendix~\ref{app:neural}.

\textbf{Metrics.}
The primary criterion is scaled pinball loss (SPL; lower is better)
averaged over $q\in\{0.1,0.25,0.5,0.75,0.9\}$. MAE and RMSSE are
secondary (Section~\ref{sec:cautions}). Raw \EBB{} is reported without
post-hoc calibration; adaptive variants appear as separate rows. Point,
interval, far-tail, significance, and efficiency results are in
Appendix~\ref{app:extra}.

\FloatBarrier
\subsection{Where pooling pays}\label{sec:pays}

\begin{figure}[!htbp]
\centering
\includegraphics[width=0.98\textwidth]{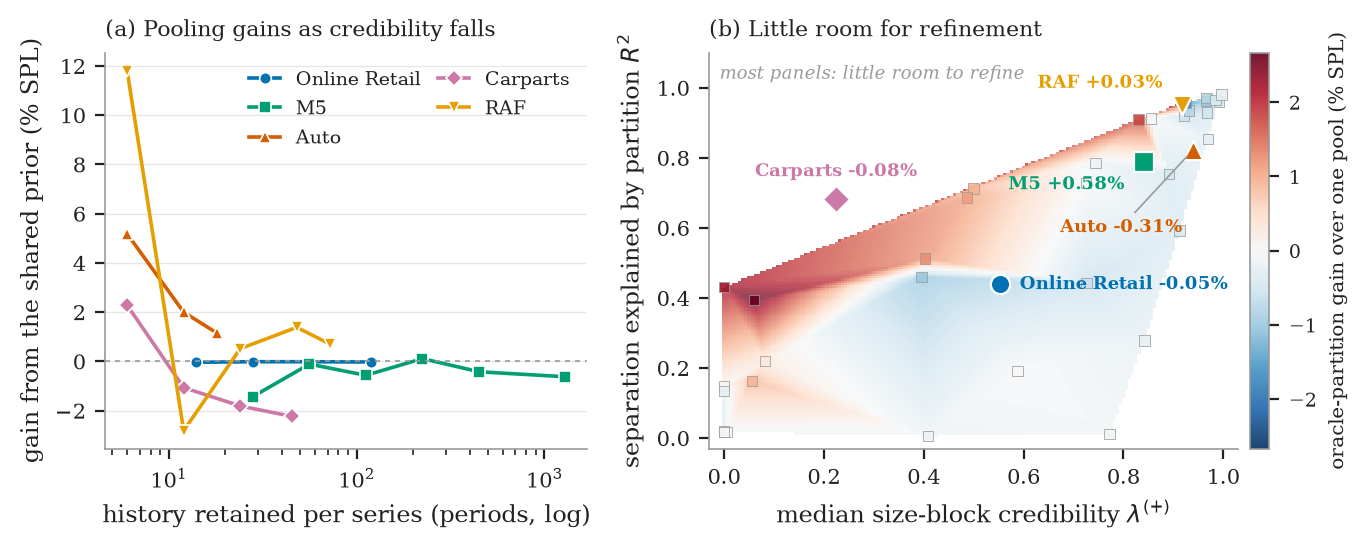}
\caption{\textbf{Forgetting creates leverage for pooling while reducing
the resolution of finer pooling structure.}
\textbf{(a)} Gain in mean SPL from the shared prior, \EBB{} against the
same model with its prior switched off, as the history retained per
series at initialization is shortened (evaluation span fixed, audited
pooling rule and discount, labels relearned on each truncated window;
Table~\ref{tab:coldstart-nopool}). The open marker is RAF at $L=12$,
where the prior-off arm is refit on twelve observations and the
comparison is unstable.
\textbf{(b)} Controlled surface
over median size-block credibility and the share of item log-size
variance explained by the true partition ($R^2$), colored by the true
partition's SPL gain over a single pool; small squares are the 36
simulated cells and the field is a linear interpolation between them,
drawn only where cells exist. The five real panels are projected onto
the same coordinates using their learned partition; labels give the
realized refinement gain. Appendix~\ref{app:demoted} shows the taxonomy
projection and the separation sweep.}
\label{fig:pays}
\end{figure}

\begin{table}[!htbp]
\centering
\caption{\textbf{Mean scaled pinball loss (lower is better) under
fixed-origin and strict walk-forward evaluation}
(protocols in Section~\ref{sec:setup}). Panels are ordered by median
occurrence credibility at the selected discount, given in the first
row. Underline marks the lowest value per column; bold marks it
together with every method a paired per-series test does not
distinguish from it (Holm-adjusted, $5\%$; Appendix~\ref{app:extra}).
The all-zero reference is unranked: on RAF it is competitive at these
quantile levels and falls behind every intensity-bearing method from
$q=0.95$ on (Appendix~\ref{app:fartail}). Adaptive rows coincide with
their base method at
fixed origin; $^{\dagger}$applies ACI only where the
initialization-window tail selects it. Cost is the walk-forward
wall-clock range over panels (TweedieGP twelve workers, \EBB{} one
core); $^{\ddagger}$M5 projected at about $51$\,h, not run. Full
results in Appendices~\ref{app:demoted} and~\ref{app:extra}.}
\label{tab:main}
\small
\setlength{\tabcolsep}{2.5pt}
\begin{adjustbox}{max width=\linewidth}
\begin{tabular}{lrrrrrrrrrrl}
\toprule
 & \multicolumn{5}{c}{Fixed origin} & \multicolumn{5}{c}{Walk-forward} & \\
\cmidrule(lr){2-6}\cmidrule(lr){7-11}
Method & RAF & Auto & Carp. & OR & M5 & RAF & Auto & Carp. & OR & M5 & Cost\\
\emph{Median} $\lambda^{(o)}$ & 0.11 & 0.42 & 0.63 & 0.90 & 0.93 & 0.11 & 0.42 & 0.63 & 0.90 & 0.93 & \\
\midrule
\multicolumn{12}{l}{\emph{Conformal wrappers on classical forecasters}}\\
CP-Croston & 0.332 & 0.326 & 0.401 & 1.550 & 1.690 & 0.330 & 0.318 & 0.395 & 1.372 & 1.567 & $\approx$\,CP-ADIDA\\
CP-SBA & 0.329 & 0.326 & 0.398 & 1.549 & 1.694 & 0.327 & 0.318 & 0.393 & 1.377 & 1.576 & $\approx$\,CP-ADIDA\\
CP-TSB & 0.381 & 0.346 & 0.378 & 1.174 & 1.879 & 0.384 & 0.330 & 0.384 & 1.373 & 1.662 & $\approx$\,CP-ADIDA\\
CP-ADIDA & 0.310 & 0.328 & 0.357 & 1.124 & 1.715 & 0.309 & 0.318 & 0.351 & 1.158 & 1.554 & 4\,s--7.8\,min\\
CP-IMAPA & 0.312 & 0.328 & 0.356 & 1.128 & 1.719 & 0.313 & 0.318 & 0.349 & 1.177 & 1.546 & $\approx$\,CP-ADIDA\\
ACI-ADIDA & \multicolumn{5}{c}{= CP-ADIDA} & 0.308 & 0.315 & 0.342 & 0.911 & 1.419 & 7\,s--9.7\,min\\
\multicolumn{12}{l}{\emph{Model-based forecasters}}\\
AutoARIMA & 0.402 & 0.319 & 0.408 & 1.007 & 1.792 & 0.408 & 0.314 & 0.419 & 1.063 & --- & 13\,min--2.7\,h\\
AutoTheta & 0.410 & 0.340 & 0.393 & 1.006 & 1.795 & 0.431 & 0.322 & 0.406 & 1.059 & --- & 13\,min--2.7\,h\\
iETS & 0.349 & 0.345 & 0.456 & 0.985 & 2.028 & 0.354 & 0.329 & 0.436 & 1.109 & --- & 3.6--30\,min\\
DeepAR & 0.273 & 0.328 & \best{\secondbest{0.317}} & --- & --- & 0.289 & 0.299 & \best{0.313} & \best{\secondbest{0.820}} & \best{\secondbest{1.263}} & 76\,s--68\,min\\
TweedieGP & 0.276 & 0.298 & \best{0.321} & \best{\secondbest{0.884}} & \best{\secondbest{1.623}} & 0.280 & \best{0.292} & \best{\secondbest{0.312}} & 0.853 & --- & 36\,min--2.4\,h$^{\ddagger}$\\
\multicolumn{12}{l}{\emph{Reference (unranked)}}\\
All-zero forecast & 0.267 & 0.561 & 0.374 & 0.916 & 1.959 & 0.267 & 0.561 & 0.374 & 0.916 & 1.959 & ---\\
\multicolumn{12}{l}{\emph{Ours}}\\
\EBB{} & \best{\secondbest{0.268}} & \best{\secondbest{0.294}} & 0.323 & \best{0.885} & 1.671 & \best{\secondbest{0.268}} & \best{0.291} & \best{0.315} & 0.854 & 1.485 & 3--37\,s\\
\EBB{} (rule)$^{\dagger}$ & \multicolumn{5}{c}{= \EBB{}} & \best{\secondbest{0.268}} & \best{\secondbest{0.290}} & \best{0.315} & 0.854 & 1.379 & 3\,s--3.6\,min\\
\bottomrule
\end{tabular}
\end{adjustbox}

\end{table}

The theory poses two questions that aggregate results confound: as
credibility falls, does the shared prior's added leverage help, and
when there is room to refine the pool, is it collected?
(Figure~\ref{fig:pays}.)

\paragraph{Pooling pays where credibility is low.}
We truncate each initialization window to its last $L$ observations per
series, leaving the evaluation span fixed, and compare \EBB{} with the
same model with its shared prior switched off, so that every item with
data keeps $\lambda_i=1$. At $L=6$ the prior is worth $5.2\%$ of mean
SPL on Auto, $11.8\%$ on RAF, and $2.3\%$ on Carparts. On Auto the gain
falls monotonically to $2.0\%$ and $1.2\%$ as history returns; on RAF
it is unstable at $L=12$ and between $0.5\%$ and $1.4\%$ thereafter.
Switching one block's prior off at a time locates the gain in the
occurrence block: at $L=6$ the occurrence prior is worth $4.3\%$ on Auto
and $14.1\%$ on RAF, the size prior within $0.3\%$ on both. Online
Retail and M5 are the controls: on M5 both median credibility weights
stay above $0.8$ even with 28 days of history, and the prior is worth
between $-1.4\%$ and $+0.1\%$ at every length. The gain therefore
follows credibility, not history length, and \EBB{}'s advantage over
TweedieGP tracks it, reaching $6.6\%$ and $11.3\%$ on Auto and RAF at
$L=6$ (Table~\ref{tab:coldstart-nopool},
Figure~\ref{fig:prior-vs-tg}).

\paragraph{Room to refine exists on one panel and is not collected.}
Reading the controlled surface at each panel's selected forgetting
regime (Table~\ref{tab:separation}) gives Online Retail, M5, Auto, and
RAF between $-0.5\%$ and $+0.6\%$ of oracle room, and their realized
refinement gains are within $0.6\%$ of zero: there was little to
collect. Carparts has $1.7$--$2.0\%$ of room and returns $-0.08\%$. The
simulation reproduces the failure: at Carparts's coordinates the same
mixture objective loses $0.7$--$0.8\%$, recovering the oracle gain only
when group centers are strongly separated and histories long, and
neither label sample-splitting nor extra shrinkage of the group
variance helps (Appendix~\ref{app:demoted}). Carparts is also the one
panel where the single pool itself costs accuracy at full length:
switching the prior off improves mean SPL by $2.2\%$, with both blocks
contributing ($1.2\%$ occurrence, $1.5\%$ size;
Table~\ref{tab:coldstart-nopool}). A no-pooling candidate added to the
selection family is not selected there (third of 32 on the validation
tail) and changes the selection only on Online Retail, where external
mean SPL moves by at most $0.13\%$ (Table~\ref{tab:none-candidate}).
The diagnostic thus separates \emph{lack of room}, on four panels, from
\emph{failure to learn available structure}, on Carparts, and places
the open problem in the partition objective.

\paragraph{Forgetting is an order of magnitude larger.}
On M5 both choices sit on one selection surface. At the selected
discount, refining the pool improves validation SPL by $0.58\%$;
introducing the discount at fixed structure improves it by $8.95\%$ for
the global pool and $9.46\%$ for the mixture.

\paragraph{Room can be screened before fitting.}
The surface answers ex post, with true labels; the screen answers ex
ante, from the single-pool fit alone. On an expanded controlled grid of
$2{,}700$ panels, a closed-form score,
$\mathrm{med}_i(1-\lambda_i^{(+)})^2\,\hat\tau^2$, ranks the room a true
partition could earn with held-out AUC $0.60$--$0.71$ for gains above
one percent, and a logistic screen on the same fitting-window quantities
reaches $0.82$--$0.84$ when separation levels, lengths, or random folds
are held out ($0.68$ when a whole discount is held out;
Appendix~\ref{app:room}). Ruling out is the useful direction: it is the
common case, four panels of five here, and what it saves is the
partition search itself. On real series, with a known partition
injected into Online Retail, Carparts, and M5 (group offsets up to two
log-units), the oracle partition never earns more than $0.4\%$, because
the single pool absorbs the heterogeneity into $\hat\tau^2$, and the
screen assigns every such panel a probability below $0.17$ of material
room.

\FloatBarrier
\subsection{Accuracy and cost}\label{sec:accuracy}

\paragraph{Against the per-series specialist.}
At fixed origin \EBB{} has the lowest mean SPL on Auto and RAF,
improving on TweedieGP by $1.6\%$ and $3.0\%$ (both significant);
TweedieGP leads on Online Retail, Carparts, and M5 by $0.05\%$, $0.5\%$,
and $2.9\%$, only the M5 gap significant. \EBB{} is thus statistically
indistinguishable from or better than the specialist on four of five
panels. Under walk-forward updating the picture holds on the four
panels where both run: \EBB{} leads on Auto and RAF, TweedieGP on Online
Retail and Carparts by $0.2\%$ and $0.8\%$, and paired tests separate
the two on RAF only, in \EBB{}'s favor. The cost gap is two to three
orders of magnitude (Table~\ref{tab:main}, last column): TweedieGP
refits a variational GP per series, whereas \EBB{} completes every
walk-forward panel in under 40 seconds on one core.

\paragraph{Against a neural global model, credibility orders the
outcome.}
Under walk-forward evaluation \EBB{} beats DeepAR by $7.8\%$ on RAF
(median $\lambda^{(o)}=0.11$) and $2.8\%$ on Auto ($0.42$), is
indistinguishable from it on Carparts ($0.63$), and trails it on the
two daily panels, by $4.1\%$ on Online Retail ($0.90$) and on M5
($0.93$) by $9.1\%$ with the validation-selected ACI layer, $17.6\%$
without it. There DeepAR leads the entire field, the GP specialist
included: it is $3.9\%$ ahead of TweedieGP on Online Retail, and on M5
\EBB{} (rule) is the closest method to it, with ACI-ADIDA next at
$12.4\%$. The ordering is monotone across the five panels and holds
within the monthly and within the daily panels separately
(Figure~\ref{fig:deepar-credibility}); at fixed origin the signs agree
on the three monthly panels. This is the pattern
Section~\ref{sec:pays} finds from inside the model: the two panels
where the shared prior earns the most are the two where a seconds-scale
empirical-Bayes fit beats a trained network, and where median
credibility is near or above $0.9$ the prior is worth nothing within
\EBB{} either. What is left to model there is dynamics and calendar
structure, which DeepAR conditions on and \EBB{} omits by design. Read
before any model is trained, occurrence credibility thus indicates
which family to reach for. Among statistical and conformal methods,
\EBB{} with the ACI rule stays within $0.8\%$ of the best on all five
panels.

A matched full-posterior Gibbs control on the same
hierarchy changes mean SPL by less than $0.1\%$, so the plug-in
approximation is not what separates the methods
(Appendix~\ref{app:mcmc}). Point forecasts are competitive but not
uniformly best (Appendix~\ref{app:point}).

\FloatBarrier
\subsection{Two evaluation cautions}\label{sec:cautions}
First, the conformal construction can matter more than the base
forecaster: making the ADIDA wrapper adaptive lowers its mean SPL on
Online Retail from $1.158$ to $0.911$, a $21\%$ change that exceeds
most differences among base forecasters. Second, absolute-error metrics
favor degenerate forecasts: the all-zero forecast attains the lowest
MAE on three of five panels, and in the controlled study MAE ranks the
known-correct partition as harmful where the likelihood-based scores
rank it as helpful (Table~\ref{tab:sanity}).

\section{Limitations and scope}\label{sec:limitations}

Three limitations qualify the conclusions. First, \EBB{} is not
uniformly the most accurate forecaster: the per-series TweedieGP remains
better on M5 at fixed origin, and model-side extensions---heavier-tailed
size laws, per-item and two-rate discounting---narrow the gap at most
marginally (Appendix~\ref{app:negative}).
It has no calendar component, and a default DeepAR is better on the two
daily panels under walk-forward evaluation.
Second, the resolution theory
is estimator-specific: Corollary~\ref{cor:resolution} rests on the
Gaussian working model, on the pseudo-count variance the estimator
attributes to discounted means, and on the moment estimator that
initializes \EBB{}'s REML fit, so it is a necessary condition for this
estimator rather than an information-theoretic bound. The pre-fit
screen is validated in controlled simulation, where it interpolates
across held-out separation and history regimes but extrapolates poorly
to an unseen discount; on real series it validates rejections, not
detections. Third, the comparison does not exhaust the model class or
the partition objectives. We do not rerun the external hierarchical
Bayesian systems of Section~\ref{sec:related}; a same-model Gibbs
control shows the plug-in EB approximation itself has little effect
(Appendix~\ref{app:mcmc}), but richer hierarchical dynamics,
foundation-model and DeepState baselines under matched protocols
(Appendix~\ref{app:neural}), soft or separation-aware mixture
objectives in place of hard assignments, repeated online re-selection
of the discount, and criteria such as CRPS remain open.

\section{Conclusion}\label{sec:conclusion}

Forgetting and pooling are one decision. Under a shared recency
operator, the credibility an item retains governs both how much a
shared prior can move its forecast and how well the data can resolve
the structure of that prior, and both can be read from the fitting
window before a partition is fitted. On five panels the diagnostic is
right in both directions: the shared prior pays where credibility is
low and pays more as histories shorten, refinement is worth at most
$0.6\%$ on four panels and no more is realized, and on the one panel
with room the mixture objective does not collect it, in data and in
simulation alike. The forecaster that carries the diagnostic matches
the strongest per-series specialist to within a few percent under both
protocols at a small fraction of its cost, and occurrence credibility
orders its wins and losses against a neural global model. What remains
open is the objective, not the
bound: partitions that optimize resolvable separation, and sequential
procedures that reconsider memory and pooling together as data arrive.

\clearpage
\appendix

\section{Notation}\label{app:notation}

\begin{table}[!htbp]
\centering
\small
\begin{adjustbox}{max width=\linewidth}
\begin{tabular}{ll}
\toprule
$y_{i,t}$, $z_{i,t}$, $\ell_{i,t}$ & demand, occurrence indicator, log-size\\
$T_i$ & initialization-window length of item $i$\\
$g(i)$, $K$ & pooling group of item $i$; number of mixture components\\
$\alpha_g,\beta_g$ & Beta occurrence prior; $\phi_g=\alpha_g+\beta_g$ its strength\\
$\mu_{0,g},\tau^2_g,\sigma^2_g$ & size prior location, between/within-item variances\\
$w$; $b$ & recency discount; forgetting strength $b=1-w$\\
$n_i(w), c_i(w)$ & discounted observation / occurrence pseudo-counts\\
$\bar p_i(w)$ & discounted empirical occurrence rate $c_i(w)/n_i(w)$\\
$\lambda_i(w)$ & credibility weight $n_i(w)/(n_i(w)+\phi_g)$\\
$\hat\pi_i,\hat\mu_i,\hat\sigma^2_i$ & posterior means of occurrence, log-size, variance\\
$\alpha_p$ & classical TSB occurrence smoothing constant ($=1-w$)\\
$\delta$ & occurrence drift rate (Proposition~\ref{prop:drift})\\
$M$, $n$ & candidate count and validation size (Prop.~\ref{prop:oracle})\\
$B$ & bootstrap draws for hyperparameter averaging\\
\bottomrule
\end{tabular}
\end{adjustbox}
\end{table}

\section{Supporting Results}\label{app:supporting}

This appendix records four connections that support the construction but
are not the paper's main theoretical contribution: the relation to TSB,
a tracking bound under drift, finite-class selection, and classical
credibility.

\FloatBarrier
\subsection{Relation to the TSB occurrence recursion}

Write $p_t=z_{i,t}$ for one fixed item. Let $\hat p_T$ be classical
TSB's occurrence estimate with smoothing $\alpha_p=1-w$ and
initialization $\hat p_0$.

\begin{proposition}[TSB occurrence as a boundary case]
\label{prop:tsb-limit}
For $w\in(0,1)$:
(i) $\lim_{(\alpha,\beta)\to(0,0)}\hat\pi(w)=\bar p(w)$;
(ii) $\hat p_T=(1-w)\sum_{t=1}^{T}w^{T-t}p_t+w^T\hat p_0$ and
$\bar p(w)=(\hat p_T-w^T\hat p_0)/(1-w^T)$;
(iii) hence $\bar p(w)=\hat p_T$ under the self-consistent initialization
$\hat p_0=\bar p(w)$, and
$|\bar p(w)-\hat p_T|\le w^T$ in general.
\end{proposition}

The equivalence concerns only the occurrence block. Classical TSB uses
arithmetic smoothing for positive sizes, whereas \EBB{} discounts
positive log-sizes under \eqref{eq:model}.

\FloatBarrier
\subsection{Tracking bound under occurrence drift}

\begin{proposition}[Tracking risk under drift]
\label{prop:drift}
If $p_t\sim\mathrm{Bern}(\pi_t)$ independently and
$|\pi_t-\pi_T|\le\delta(T-t)$, then
$|\E\bar p(w)-\pi_T|\le\delta w/[(1-w)(1-w^T)]$ and
$\Var\bar p(w)\le\tfrac14(1-w)(1+w^T)/[(1+w)(1-w^T)]$, with the
corresponding bias--variance decomposition for the posterior mean.
\end{proposition}

\begin{corollary}\label{cor:rate}
Balancing the two bounds gives
$1-w^*\asymp\delta^{2/3}$, worst-case MSE
$O(\delta^{2/3})$, and $w^*\to1$ as $\delta\to0$.
\end{corollary}

This result motivates including $w=1$ in the candidate grid; it does not
claim that chronological validation estimates the bound-optimal $w^*$.

\begin{figure}[!htbp]
\centering
\includegraphics[width=0.93\linewidth]{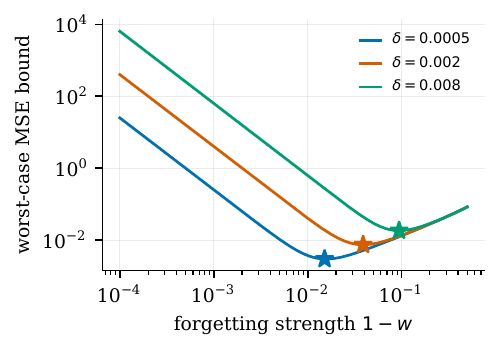}
\caption{Bias--variance bounds in Proposition~\ref{prop:drift}. Stars
mark the minimizing forgetting strength; as drift $\delta$ increases,
the optimum moves at the rate in Corollary~\ref{cor:rate}.}
\label{fig:tradeoff}
\end{figure}

\FloatBarrier
\subsection{Finite candidate selection}

\begin{proposition}[Conditional finite-class bound]
\label{prop:oracle}
Let $R(c)$ be the expected validation risk of candidate $c$ and
$\hat R(c)$ its average over $n$ validation items. If the item losses are
independent conditional on the fitted candidates and lie in $[0,B]$, then
the minimizer $\hat c$ over $M$ candidates satisfies, with probability at
least $1-\eta$,
\[
R(\hat c)\le\min_cR(c)+B\sqrt{\frac{2\log(2M/\eta)}{n}}.
\]
\end{proposition}

The independent units are items, because observations and quantiles are
averaged within item before selection. The selector does not
clip scaled pinball loss, so the bounded-loss assumption is not automatic
and this proposition is not presented as a formal guarantee for the
unclipped implementation. It records the standard finite-class behavior
under an explicit clipping condition.

\FloatBarrier
\subsection{Classical credibility}

\begin{proposition}[Pooling reduces Bayes risk in the stationary model]
\label{prop:credibility}
Under the undiscounted Beta--Binomial occurrence model with known
hyperparameters, the posterior mean is the Bayes estimator of $\pi_i$
under squared loss and has strictly smaller Bayes risk than the per-item
empirical rate whenever $\phi_g>0$. Its optimal weight is the classical
credibility factor $\lambda_i=n_i/(n_i+\phi_g)$
\citep{buhlmann1967}.
\end{proposition}

This is the B\"uhlmann--Straub result specialized to Bernoulli
occurrences. It justifies pooling when the stationary hierarchical model
is correctly specified; it does not assert that a misspecified or stale
group prior helps under drift.

\section{Proofs}\label{app:proofs}

\FloatBarrier
\subsection{Proof of Proposition~\ref{prop:tsb-limit}}
(i) Immediate from the discounted form of \eqref{eq:credibility-pair}
since $c(w),n(w)$ are finite.
(ii) Induction: $\hat p_1=(1-w)p_1+w\hat p_0$. If the closed form holds at
$T-1$, then
$\hat p_T=(1-w)p_T+w\hat p_{T-1}=(1-w)\sum_{t=1}^{T}w^{T-t}p_t+w^{T}\hat p_0$.
With $n(w)=(1-w^{T})/(1-w)$,
\[
\bar p(w)=\frac{(1-w)\sum_t w^{T-t}p_t}{1-w^{T}}
=\frac{\hat p_T-w^{T}\hat p_0}{1-w^{T}}.
\]
(iii) Rearranging, $\bar p(w)-\hat p_T=w^{T}(\bar p(w)-\hat p_0)$, and
$|\bar p(w)-\hat p_0|\le1$. \qed

\FloatBarrier
\subsection{Proof of Proposition~\ref{prop:active}}
$n(w)=(1-w^{T})/(1-w)\le1/(1-w)$ and $x\mapsto x/(x+\phi)$ is increasing,
so $\lambda(w)\le[1/(1-w)]/[1/(1-w)+\phi]=1/(1+(1-w)\phi)$. \qed

\FloatBarrier
\subsection{Proof of Proposition~\ref{prop:drift} and
Corollary~\ref{cor:rate}}
With $k=T-t$ and $\sum_{k\ge0}kw^k=w/(1-w)^2$,
\[
\bigl|\E\bar p(w)-\pi_T\bigr|
=\Bigl|\frac{\sum_t w^{T-t}(\pi_t-\pi_T)}{n(w)}\Bigr|
\le\frac{\delta\,w/(1-w)^2}{(1-w^{T})/(1-w)}
=\delta\,\frac{w}{(1-w)(1-w^{T})}.
\]
By independence and $\pi_t(1-\pi_t)\le\tfrac14$,
\[
\Var\,\bar p(w)\le\frac{\tfrac14\,(1-w^{2T})/(1-w^{2})}
{[(1-w^{T})/(1-w)]^{2}}
=\frac14\cdot\frac{(1-w)(1+w^{T})}{(1+w)(1-w^{T})}.
\]
Decomposing
$\hat\pi-\pi_T=\lambda(\bar p-\E\bar p)+\lambda(\E\bar p-\pi_T)+(1-\lambda)(\mu_\pi-\pi_T)$,
the first term is zero-mean and independent of the deterministic
remainder, giving
$\E(\hat\pi-\pi_T)^2\le[\lambda\mathrm{B}(w)+(1-\lambda)|\mu_\pi-\pi_T|]^2
+\lambda^2\Var\bar p(w)$ with $\mathrm{B}(w)$ the bias bound. For the
corollary set $b=1-w$ and let $T\to\infty$ (so $w^{T}\to0$): the squared
bias bound behaves as $\delta^2(1-b)^2/b^2\sim\delta^2/b^2$ and the
variance bound becomes $\tfrac14\, b/(2-b)$, which is $b/8$ to leading
order as $b\to0$. Minimizing $f(b)=\delta^2/b^2+b/8$ gives
$f'(b)=-2\delta^2/b^3+1/8=0$, i.e.,
$b^{*}=(16\delta^{2})^{1/3}=2^{4/3}\delta^{2/3}$, and
$f(b^{*})=O(\delta^{2/3})$. \qed

\FloatBarrier
\subsection{Proof of Proposition~\ref{prop:oracle}}
For each candidate $c$, Hoeffding's inequality for independent scores in
$[0,B]$ gives
$\Pr\bigl(|\hat R(c)-R(c)|>t\bigr)\le2\exp(-2nt^2/B^2)$. A union bound
over the $M$ candidates makes
$\sup_c|\hat R(c)-R(c)|\le t$ hold with probability at least
$1-2M\exp(-2nt^2/B^2)$. Setting the right side to $\eta$ yields
$t=B\sqrt{\log(2M/\eta)/(2n)}$. On this event,
$R(\hat c)\le\hat R(\hat c)+t\le\hat R(c^{*})+t\le R(c^{*})+2t$ with
$c^{*}=\arg\min_c R(c)$, which is the claim with constant
$2t=B\sqrt{2\log(2M/\eta)/n}$. \qed

\FloatBarrier
\subsection{Proof of Proposition~\ref{prop:credibility}}
With known $(\alpha_g,\beta_g)$ and $s_i\mid\pi_i\sim
\mathrm{Bin}(n_i,\pi_i)$, $\pi_i\sim\mathrm{Beta}(\alpha_g,\beta_g)$,
the posterior mean $\hat\pi_i=\lambda_i\bar p_i+(1-\lambda_i)\mu_\pi$
with $\lambda_i=n_i/(n_i+\phi_g)$ minimizes Bayes risk under squared
loss among all estimators (tower property). Restricting to linear
combinations $a\bar p_i+b$, the Bayes risk is
$a^2\,\E[\Var(\bar p_i\mid\pi_i)] + (1-a)^2\Var(\pi_i)$ at the optimal
intercept, minimized at
$a^{*}=\Var(\pi_i)/\bigl(\Var(\pi_i)+\E[\Var(\bar p_i\mid\pi_i)]\bigr)
=\lambda_i$, the B\"uhlmann credibility weight. The minimum is strictly
below the risk of $a=1$ (the empirical rate) whenever
$\Var(\pi_i)<\infty$ and $\phi_g>0$. \qed

\FloatBarrier
\subsection{Proof of Proposition~\ref{prop:twosided} and
Corollary~\ref{cor:resolution}}

\textbf{(i)} $\lambda_i=n_i/(n_i+\kappa_g)$ is increasing in $n_i$, and
$n_i\le\sum_{k\ge0}w^{k}=(1-w)^{-1}$ for $w<1$. Substituting the bound,
$\lambda_i\le\frac{(1-w)^{-1}}{(1-w)^{-1}+\kappa_g}
=\bigl[1+(1-w)\kappa_g\bigr]^{-1}$, which is $<1$ whenever $w<1$ and
$\kappa_g>0$.

\textbf{(ii)} Write $\bar y_i=\mu_i+e_i$ with
$\mu_i\sim N(\mu_g,\tau_g^2)$ and $e_i\sim N(0,s_i)$ independent,
$s_i=\sigma_g^2/n_i$, so $\Var(\bar y_i)=\tau_g^2+s_i$. For the sample
variance $S_g^2=(m_g-1)^{-1}\sum_{i\in g}(\bar y_i-\bar{\bar y})^2$,
independence gives
$\Var(\bar{\bar y})=m_g^{-2}\sum_i(\tau_g^2+s_i)$ and hence
\[
\E[S_g^2]=\frac{\sum_i(\tau_g^2+s_i)-m_g\Var(\bar{\bar y})}{m_g-1}
=\frac{(1-m_g^{-1})\sum_i(\tau_g^2+s_i)}{m_g-1}
=\tau_g^2+\bar s_g ,
\]
exactly, with no balance assumption. The estimator truncates precisely
when $S_g^2\le\bar s_g$. Under the working model $S_g^2$ is
approximately $(\tau_g^2+\bar s_g)\chi^2_{m_g-1}/(m_g-1)$, so
$\Var(S_g^2)\simeq\frac{2}{m_g-1}(\tau_g^2+\bar s_g)^2$ and the
standardized distance from the truncation boundary is
\[
\frac{\E[S_g^2]-\bar s_g}{\sqrt{\Var(S_g^2)}}
=\frac{\tau_g^2}{\sqrt{2/(m_g-1)}\,(\tau_g^2+\bar s_g)}
=\sqrt{\tfrac{m_g-1}{2}}\;\rho_g ,
\]
giving the stated normal approximation for
$\Pr[S_g^2\le\bar s_g]$. The discount bound follows from
$n_i\le(1-w)^{-1}$: $s_i=\sigma_g^2/n_i\ge(1-w)\sigma_g^2$ for every
$i$, hence $\bar s_g\ge(1-w)\sigma_g^2$, and $\rho_g$ is decreasing in
$\bar s_g$. Finally $\hat\tau_g^2=0$ gives $\kappa_g=\infty$ and
$\lambda_i=0$.

\textbf{Corollary.} Requiring the collapse probability to stay below a
fixed level is, by (ii), the condition $\rho_g\sqrt{(m_g-1)/2}\ge c$ for
a constant $c$ of order one. Solving
$\rho_g=\tau_g^2/(\tau_g^2+\bar s_g)\ge c\sqrt{2/(m_g-1)}$ for
$\tau_g^2$ gives
$\tau_g^2\ge\bar s_g\,c\sqrt{2}/(\sqrt{m_g-1}-c\sqrt2)$, which is
\eqref{eq:resolution-condition} with $c_\delta=c\sqrt2$ and
$c=\Phi^{-1}(1-\delta)$. Refining a
partition decreases $m_g$ and, by construction of the mixture objective,
$\tau_g^2$, while decreasing $w$ raises the floor $(1-w)\sigma_g^2$ on
$\bar s_g$. Each shrinks the left side or raises the right side. \qed

\FloatBarrier
\subsection{An estimator-free noise floor}

\begin{lemma}[Sampling variance of a discounted mean]
\label{lem:floor}
Let $x_1,\dots,x_n$ be uncorrelated with common variance $\sigma^2$, and
let $a_j=w^{k_j}$ with $w\in(0,1)$ and integers $0\le k_1<\dots<k_n$.
Then
\[
\Var\Bigl(\frac{\sum_j a_jx_j}{\sum_j a_j}\Bigr)
=\sigma^2\frac{\sum_j a_j^2}{(\sum_j a_j)^2}
\;\ge\;\sigma^2\,\frac{1-w}{1+w},
\]
for every $n$ and every choice of the $k_j$, and the variance is at most
$\sigma^2/\sum_j a_j$.
\end{lemma}

\emph{Proof.} The equality is the variance of a weighted mean. For the
lower bound write $c=(1-w)/(1+w)$ and argue by induction on $n$; the
ratio is invariant to rescaling the weights, so take $a_1=1$. For $n=1$
the ratio is $1\ge c$. For $n>1$ let $r=\sum_{j\ge2}a_j$ and
$q=\sum_{j\ge2}a_j^2$. The tail weights are again powers of $w$ with
increasing exponents, so $q\ge c\,r^2$ by the induction hypothesis, and
$a_j\le w^{\,j-1}$ gives $r\le w/(1-w)$. Hence the ratio is at least
$g(r)=(1+c\,r^2)/(1+r)^2$, whose derivative has the sign of $c\,r-1<0$ on
$0\le r\le w/(1-w)<1/c$. Evaluating at the right end point,
$g\{w/(1-w)\}=(1-w)^2+c\,w^2=(1-w)\{(1-w)(1+w)+w^2\}/(1+w)=c$. The upper
bound follows from $a_j^2\le a_j$. \qed

The lemma is applied to the positive log-sizes of one item, conditionally
on its occurrence pattern: whatever the length of the history, the mean
of exponentially discounted observations retains sampling variance of
order $(1-w)\sigma^2$, between $\sigma^2(1-w)/(1+w)$ and the working
value $s_i=\sigma^2/n_i^+$ used by the estimator. The noise floor in
Proposition~\ref{prop:twosided} is therefore a property of the
discounted statistic and not of the estimator built on it; since the
working value is the larger of the two, $\E[S_g^2]\le\tau_g^2+\bar s_g$
under the sampling model and the resolution condition is necessary for
the moment estimator.

The algebraic steps, the exact unbiasedness in (ii), and the monotonicity
claims are verified with a computer algebra system in
the authors' test suite. The $\chi^2$ approximation in (ii) is checked
against Monte Carlo simulation, agreeing to within $0.05$ in probability
across the regimes we use.

\section{Implementation details}\label{app:implementation}

\paragraph{EB fitting.} Group Beta--Binomial hyperparameters maximize the
(weighted) marginal likelihood by L-BFGS-B with positivity bounds; the
size block uses pooled within-item variance for $\sigma^2_g$ and an
intercept-only restricted marginal likelihood in $\tau^2_g$ with profiled
$\mu_{0,g}$; item-level process variances use the conjugate
scaled-inv-$\chi^2$ update with $\nu{=}20$. With discounting, every count
is replaced by its weighted analogue \eqref{eq:discounted-stats}; the
Beta--Binomial marginal extends to real pseudo-counts via log-Gamma
functions. The predictive law is zero-inflated log-normal with predictive
variance $\hat\sigma^2_{\mathrm{proc}}+\hat v_\mu$. The experimental
configuration averages quantiles over $B{=}20$ bootstrap resamples of the
item-statistics rows. An optional location-scale calibration layer exists
but is \emph{off} in all headline results; the adaptive conformal
comparison is reported as a separate baseline rather than used to modify
\EBB{} after fitting.

\paragraph{Mixture EM.} Deterministic quantile initialization over
(occurrence rate, mean log-size); degenerate components (mass $<2$)
frozen to global fallback priors; tolerance $10^{-5}$, max 60 iterations;
$K\in\{1,\dots,6\}$ by BIC with $5K+(K-1)$ parameters.

\paragraph{Selection.} Internal chronological split (first 80\% / last
20\%, min head 8); score = mean pinball over $q\in\{0.5,0.75,0.9\}$,
per-series scaled by head mean absolute demand, averaged within item and then
uniformly across items; grid over
$w\in\{1,0.999,0.997,0.995,0.99,0.98,0.95,0.90\}$ and partitions
\{global, taxonomy, mixture(BIC)\}. Selected configurations per dataset
are reported beneath Table~\ref{tab:point}.

\paragraph{Environment and timing methodology.} All experiments ran on one
Windows~11 workstation (Intel Core Ultra~9 275HX, 32\,GB RAM). Core
determinism tests are bit-identical within this environment;
cross-environment bit identity is not claimed. Timing is wall-clock and
excludes data loading. Joint selection over the 24 candidates takes
$51$--$243$ seconds per panel, with one deterministic six-$K$ EM/BIC
search for the mixture on the fitting head. Final model-only point plus
five-quantile prediction with $B=20$ takes 6.2 seconds on
Carparts, 38.0 on Online Retail, and 124.0 on M5, whose
external span has 3.235 million observations. GPU is not used in \EBB{}.

\paragraph{Baselines.} StatsForecast defaults; TSB
$(\alpha_d,\alpha_p)=(0.5,0.45)$; AutoARIMA/AutoTheta season 7 (daily) /
12 (monthly) for intervals. TSB-tuned: $5\times5$ grid
$\alpha_d\in\{0.05,0.1,0.2,0.35,0.5\}$,
$\alpha_p\in\{0.01,0.05,0.1,0.25,0.45\}$ (25 candidates, the same order
as \EBB{}'s 24), scored on the identical 80/20 chronological split by
per-series scaled MAE (so neither method is scored on its home
criterion), refit on the full window. Selected $(\alpha_d,\alpha_p)$:
Online Retail $(0.10,0.25)$; M5 $(0.20,0.01)$; Auto $(0.20,0.10)$;
Carparts $(0.50,0.45)$, i.e.\ the textbook constants; RAF
$(0.50,0.25)$. Static conformal wrappers use a chronological 80/20 split
and a \emph{panel-pooled} signed-residual distribution; quantiles are
zero-truncated and monotonized. The ACI variants retain that residual
pool and adapt the requested level for every series--quantile pair with
$\gamma=0.05$. Tweedie-GLM is a per-series autoregressive GLM with log
link, $p{=}1.5$, 14 lags (daily) / 6 (monthly), and $\ell_2{=}0.1$.
TweedieGP uses the authors' official implementation with Tweedie
likelihood, median-demand scaling, RBF kernel, and 50{,}000 predictive
samples; zero-only training series receive a zero fallback so that every
series is scored. iETS uses \texttt{smooth::adam} with automatic
occurrence modeling.

\section{Main-text material in full}\label{app:demoted}

This appendix keeps, verbatim, the tables, figures, and paragraphs of the
long-form manuscript that the main text summarizes.

\FloatBarrier
\subsection{The two coupled choices in real data}
\begin{figure}[!htbp]
\centering
\includegraphics[width=0.98\textwidth]{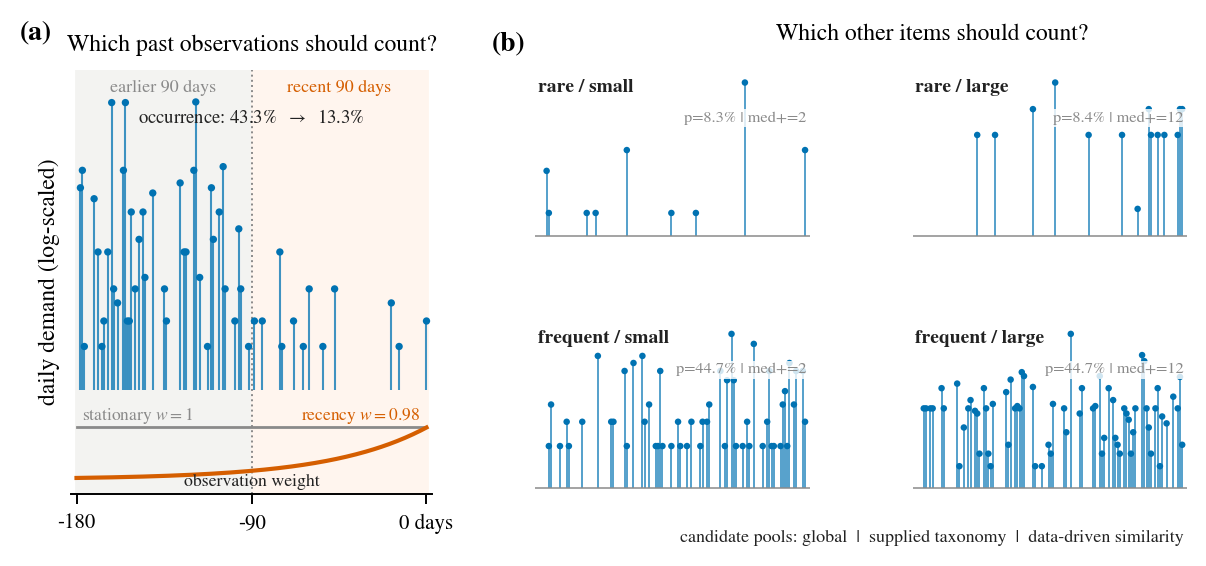}
\caption{The two coupled choices, in real intermittent demand.
\textbf{(a)} One item's daily history: occurrence falls from $43.3\%$ to
$13.3\%$ between the earlier and the recent ninety days, so a stationary
weighting ($w=1$) and a recency weighting ($w=0.98$) summarize the same
history differently---which past observations should count?
\textbf{(b)} Four items from the same panel, spanning rare and frequent
occurrence at small and large sizes: any pooled prior must decide which
of these heterogeneous neighbors an item should borrow strength
from---which other items should count? Candidate pools are a global
pool, a supplied taxonomy, and data-driven similarity.}
\label{fig:problem}
\end{figure}

\FloatBarrier
\subsection{Persistent prior leverage}

Fix an item and suppress its subscripts. Because
$n(w)=\sum_{t=1}^{T}w^{T-t}\le(1-w)^{-1}$ for $w<1$, the occurrence
credibility weight cannot converge to one.

\begin{proposition}[Forgetting keeps pooling relevant]
\label{prop:active}
For every horizon $T$,
$\lambda(w)\le\{1+(1-w)\phi\}^{-1}<1$ when $w<1$, whereas
$\lambda(1)\to1$ as $T\to\infty$. Hence the pooled occurrence prior
retains positive weight indefinitely under forgetting.
\end{proposition}

The statement is about leverage, not benefit: it does not guarantee
that a particular prior is a good shrinkage target. It says that the
choice of target can continue to affect forecasts even for old series.

\FloatBarrier
\subsection{Resolution diagnostics recorded by the selector}
Fitting each candidate already produces everything the resolution
analysis of Section~\ref{sec:resolution} needs, so the selector records it
rather than recomputing it later. For every $(r,w)$ and every resulting
group $g$ we store, from the fitting head only, the median occurrence and
positive-size credibility weights $\lambda_i^{(o)}$ and $\lambda_i^{(+)}$,
the share of items with $\lambda_i^{(+)}<0.1$, the group sizes $m_g$, and
the resolution statistic
\[
\hat z_g=\frac{S_g^2-\bar s_g}{\sqrt{2/(m_g-1)}\,S_g^2},
\]
whose nonpositive values coincide with the collapse event
$\hat\tau_g^2=0$ of Proposition~\ref{prop:twosided}. These quantities cost
nothing beyond the fit that produced them, they are computed before any
external observation is seen, and they answer a question the validation
score cannot: not which candidate scores best on the tail, but whether the
discounted data contain enough information to estimate the candidate's
group structure at all. Section~\ref{sec:main-results} uses them to
separate the two.

\FloatBarrier
\subsection{Pooling candidates and computational profile}
Let $\mathcal R$ contain three ways to construct $g(\cdot)$ from the
fitting data. Table~\ref{tab:pool-candidates} orders them from coarse to
fine: one prior for the panel, one prior per fixed demand-pattern class,
and one prior per learned component. The ladder is deliberate. The
question this paper asks is not which rung is best in general---that
depends on the panel---but whether the discounted data support moving up
it at all. Section~\ref{sec:resolution} gives the condition under which
they do, and Section~\ref{sec:main-results} reports what happens when they
do not.

\begin{table}[!htbp]
\centering
\caption{Pooling candidates used by \EBB{}, ordered coarse to fine.
Every data-dependent label is computed from the selector's fitting head,
never its validation tail.}
\label{tab:pool-candidates}
\small
\setlength{\tabcolsep}{2.5pt}
\begin{adjustbox}{max width=\linewidth}
\begin{tabular}{lp{0.70\linewidth}}
\toprule
Candidate & Shared-prior groups\\
\midrule
Global & All items share one prior.\\
Taxonomy & Four ADI/CV$^2$ demand-pattern classes (smooth, intermittent,
erratic, lumpy) using the classical thresholds $1.32$ and $0.49$
\citep{syntetos2005categorization}.\\
Learned & Hard assignments from a mixture of occurrence--size prior
components; $K\in\{1,\ldots,6\}$ is selected by BIC.\\
\bottomrule
\end{tabular}
\end{adjustbox}
\end{table}

For the learned candidate, component $k$ carries
$\eta_k=(\alpha_k,\beta_k,\mu_{0,k},\tau_k^2,\sigma_k^2)$. Its item
marginal is the product of Beta--Binomial occurrence evidence and Normal
random-effects size evidence. EM has an analytic responsibility update;
its M-step maximizes responsibility-weighted EB objectives. A
deterministic quantile initialization makes repeated fits seed-free.
Mixture labels are built once from the fitting head and shared across the
discount grid, with $K\in\{1,\dots,6\}$ selected by BIC at that fit.
Re-learning the partition separately at every discount is a natural
variant---changing $w$ changes both the item evidence and the components
it can support---but we do not evaluate it here. The
selected responsibilities are converted to hard labels so all three
candidates share the same downstream forecaster.

Two properties of this cost profile matter for the comparisons that
follow. First, the whole pipeline is CPU-only and closed form apart from
the mixture search: fitting and predicting a panel costs on the order of
milliseconds per series, so the full 24-candidate joint selection
completes in minutes on a laptop-class machine and no GPU, tuning budget,
or convergence diagnostic is required. Second, estimation is exact rather
than approximate at every stage, so repeated end-to-end runs are
bit-identical in the authors' test suite. Neither property is an accuracy
claim; both are the terms on which the accuracy results in
Section~\ref{sec:experiments} should be read, since several of the
strongest comparators there cost two to three orders of magnitude more per
series and none is reproducible to the bit. Measured wall-clock figures
for every method are in Appendix~\ref{app:implementation}.

\FloatBarrier
\subsection{Fixed-origin results in full}
\label{sec:main-results}

\begin{table}[!htbp]
\centering
\caption{\textbf{Probabilistic forecasting is the primary comparison.}
Scaled pinball loss (SPL; lower is better) is reported as the mean over
$q\in\{0.1,0.25,0.5,0.75,0.9\}$ and at q90. Bold and underline mark the
best and second-best non-degenerate methods. M5 reports seed 42 for direct
comparability; three-seed robustness is summarized in the text. \EBB{}
uses its raw predictive law, with no post-hoc calibration. TweedieGP is the
official implementation of Damato et al.\ at its released defaults on
all five panels; Appendix~\ref{app:efficiency} reports measured cost.
The all-zero row is a metric diagnostic and
is excluded from ranking.}
\label{tab:prob}
\small
\setlength{\tabcolsep}{2.5pt}
\begin{adjustbox}{max width=\linewidth}
\begin{tabular}{lcc@{\hspace{9pt}}cc@{\hspace{9pt}}cc@{\hspace{9pt}}cc@{\hspace{9pt}}cc}
\toprule
 & \multicolumn{2}{c}{Online Retail} & \multicolumn{2}{c}{M5} & \multicolumn{2}{c}{Auto} & \multicolumn{2}{c}{Carparts} & \multicolumn{2}{c}{RAF}\\
\cmidrule(lr){2-3}\cmidrule(lr){4-5}\cmidrule(lr){6-7}\cmidrule(lr){8-9}\cmidrule(lr){10-11}
Model & Mean & q90 & Mean & q90 & Mean & q90 & Mean & q90 & Mean & q90\\
\midrule
AutoARIMA & 1.0069 & 1.6297 & 1.7919 & 2.4148 & 0.3189 & 0.2659 & 0.4084 & 0.4897 & 0.4020 & 0.5910\\
AutoTheta & 1.0055 & 1.6056 & 1.7954 & 2.3123 & 0.3403 & 0.2760 & 0.3925 & \secondbest{0.4697} & 0.4104 & 0.5956\\
CP-Croston & 1.5502 & 2.7686 & 1.6902 & \best{2.1551} & 0.3260 & 0.3105 & 0.4012 & 0.5433 & 0.3322 & 0.4927\\
CP-SBA & 1.5492 & 2.8601 & 1.6939 & 2.1736 & 0.3256 & 0.3153 & 0.3982 & 0.5418 & 0.3290 & 0.4920\\
CP-TSB & 1.1744 & 2.6746 & 1.8794 & 2.3148 & 0.3459 & 0.3150 & 0.3783 & 0.4842 & 0.3814 & 0.4986\\
CP-ADIDA & 1.1239 & 2.5409 & 1.7150 & 2.1949 & 0.3275 & 0.3082 & 0.3571 & 0.4982 & 0.3097 & 0.4905\\
CP-IMAPA & 1.1280 & 2.5506 & 1.7190 & 2.1904 & 0.3279 & 0.3084 & 0.3561 & 0.4908 & 0.3116 & 0.4907\\
Tweedie-GLM & 1.1477 & 1.7458 & 4.0693 & 5.2620 & 0.3221 & 0.2783 & 0.4776 & 0.6897 & 0.3121 & 0.5312\\
iETS & 0.9846 & 1.7544 & 2.0282 & 2.9772 & 0.3448 & 0.3252 & 0.4561 & 0.5646 & 0.3494 & \secondbest{0.4900}\\
TweedieGP & \best{0.8845} & \best{1.5218} & \best{1.6228} & \secondbest{2.1723} & \secondbest{0.2985} & \best{0.2598} & \best{0.3214} & \best{0.4600} & \secondbest{0.2763} & 0.5219\\
\midrule
Zero (diagnostic) & 0.9161 & 1.6490 & 1.9590 & 3.5262 & 0.5612 & 1.0102 & 0.3736 & 0.6725 & 0.2669 & 0.4805\\
\midrule
\EBB{} (ours) & \secondbest{0.8849} & \secondbest{1.5268} & \secondbest{1.6707} & 2.4758 & \best{0.2937} & \secondbest{0.2599} & \secondbest{0.3229} & 0.4817 & \best{0.2680} & \best{0.4856}\\
\bottomrule
\end{tabular}
\end{adjustbox}
\end{table}

Table~\ref{tab:prob} sets the empirical position, and it is a narrow one.
\EBB{} attains the lowest mean SPL on Auto and RAF, by $1.6\%$ and
$3.0\%$ over the official TweedieGP implementation, both significant
(Table~\ref{tab:spl-significance}); TweedieGP attains it on Online Retail,
Carparts, and M5, by $0.05\%$, $0.5\%$, and $2.9\%$, of which only the M5
margin is significant. Against every other comparator in the table
\EBB{} is best on all five panels, by $10.1\%$ over iETS on Online
Retail and $1.2\%$ over CP-Croston on M5; DeepAR, run at fixed origin
on the three monthly panels, is reported in Table~\ref{tab:deepar}.
Across three independent
5{,}000-series M5 samples the selected configuration is identical---
$(\text{mixture},\,w{=}0.95,\,K{=}6)$---and \EBB{} has mean SPL
$1.7501\pm0.084$ against $1.7547\pm0.065$ for CP-Croston, its closest
comparator: ahead on two samples, behind by $1.0\%$ on the third, and
ahead on average by $0.3\%$, well within one standard deviation. We report
the spread rather than an aggregate win because it is the quantity
Section~\ref{sec:leverage} explains. The explanation is not that these
panels leave no room for a hierarchical construction to work; on three of
them the credibility weights say otherwise. It is that the room which
exists is not converted by a learned partition, which is why the
achievable margin is set by the shared predictive law rather than by the
pooling structure.

The degenerate row exposes what the central quantiles cannot see. On RAF,
the all-zero forecast is marginally better through q90 because the true
quantiles of most series are zero. At $q\in\{0.95,0.975,0.99\}$ it falls
behind every intensity-bearing method, and intensity-bearing models become
distinguishable from one another (Appendix~\ref{app:fartail}). TweedieGP
is strongest throughout the Carparts far tail and remains competitive on
RAF. The probabilistic claim is therefore not that one distribution wins
every quantile, but that \EBB{} is the most consistent aggregate
performer across panels and remains informative in the region where a zero
forecast ceases to be useful.

Point forecasts tell a deliberately weaker story. \EBB{} is best on six
of ten MAE/RMSSE cells and top two on seven, but the zero forecast has the
lowest MAE on three panels and classical methods lead several remaining
cells. We report the complete point table and horizon decomposition in
Appendix~\ref{app:point}: they establish competitiveness and rule out a
long-horizon averaging artifact, but do not support uniform point-forecast
superiority.

\FloatBarrier
\subsection{Walk-forward results in full}
\label{sec:walkforward}

\begin{table}[!htbp]
\centering
\caption{\textbf{Strict probabilistic walk-forward evaluation on all five
panels.} Online Retail and M5 use seven-day blocks; Auto, Carparts, and
RAF use monthly blocks. Mean and q90 are scaled pinball losses; all five
quantiles, coverage, and interval width are reported in
Appendix~\ref{app:walkforward}. Classical wrappers and ACI refit each
block. AutoARIMA, AutoTheta, iETS, and TweedieGP refit every four blocks
on Online Retail and every block on the monthly panels; on M5 these four
were not run under this protocol on cost grounds (dashes). \EBB{} updates
sufficient statistics each block; ACI-\EBB{} applies the ACI-ADIDA
calibration layer to \EBB{}'s predictive quantiles, and \EBB{} (rule)
applies it only where the validation-tail rule selects it, so it equals
one of the two rows above it on each panel and is not ranked. Bold and underline
rank non-degenerate methods by SPL; on RAF the all-zero forecast is
marginally lower than every method through q90. On M5, the longest daily
panel, the adaptive conformal layer decides the ranking: ACI-\EBB{} leads,
ACI-ADIDA is second, and uncalibrated \EBB{} is third.}
\label{tab:wfprob}
\small
\setlength{\tabcolsep}{2.5pt}
\begin{adjustbox}{max width=\linewidth}
\begin{tabular}{lrrrrrrrrrr}
\toprule
 & \multicolumn{2}{c}{Online Retail} & \multicolumn{2}{c}{Carparts} & \multicolumn{2}{c}{Auto} & \multicolumn{2}{c}{RAF} & \multicolumn{2}{c}{M5}\\
\cmidrule(lr){2-3}\cmidrule(lr){4-5}\cmidrule(lr){6-7}\cmidrule(lr){8-9}\cmidrule(lr){10-11}
Model & Mean & q90 & Mean & q90 & Mean & q90 & Mean & q90 & Mean & q90\\
\midrule
AutoARIMA & 1.0627 & 1.5225 & 0.4195 & 0.4839 & 0.3139 & 0.2632 & 0.4083 & 0.5920 & --- & ---\\
AutoTheta & 1.0586 & 1.4835 & 0.4056 & 0.4743 & 0.3224 & 0.2619 & 0.4308 & 0.6233 & --- & ---\\
iETS & 1.1092 & 1.9404 & 0.4362 & 0.5154 & 0.3286 & 0.3009 & 0.3537 & 0.4902 & --- & ---\\
CP-TSB & 1.3733 & 2.8422 & 0.3836 & 0.4758 & 0.3297 & 0.3009 & 0.3842 & 0.4941 & 1.6616 & 2.0422\\
CP-ADIDA & 1.1576 & 2.5143 & 0.3509 & 0.4880 & 0.3175 & 0.2949 & 0.3095 & \secondbest{0.4894} & 1.5537 & 1.9629\\
CP-IMAPA & 1.1773 & 2.5224 & 0.3492 & 0.4789 & 0.3176 & 0.2948 & 0.3127 & 0.4903 & 1.5457 & 1.9490\\
ACI-ADIDA & 0.9108 & 1.6681 & 0.3417 & 0.4767 & 0.3147 & 0.2900 & 0.3085 & 0.5598 & \secondbest{1.4192} & \secondbest{1.8064}\\
TweedieGP & \best{0.8526} & \best{1.3791} & \best{0.3121} & \best{0.4326} & 0.2924 & \best{0.2539} & 0.2801 & 0.5261 & --- & ---\\
Zero (diagnostic) & 0.9161 & 1.6490 & 0.3736 & 0.6725 & 0.5612 & 1.0102 & 0.2669 & 0.4805 & 1.9590 & 3.5262\\
\midrule
\EBB{} & \secondbest{0.8540} & \secondbest{1.4002} & 0.3145 & 0.4578 & \secondbest{0.2910} & 0.2563 & \best{0.2681} & \best{0.4864} & 1.4851 & 1.8797\\
ACI-\EBB{} & 0.9082 & 1.5623 & \secondbest{0.3128} & \secondbest{0.4521} & \best{0.2901} & \secondbest{0.2560} & \secondbest{0.2743} & 0.5175 & \best{1.3785} & \best{1.6438}\\
\EBB{} (rule) & 0.8540 & 1.4002 & 0.3145 & 0.4578 & 0.2901 & 0.2560 & 0.2681 & 0.4864 & 1.3785 & 1.6438\\
\bottomrule
\end{tabular}
\end{adjustbox}
\end{table}

Table~\ref{tab:wfprob} is the deployment-facing test, and it reproduces
the fixed-origin picture for the uncalibrated methods. \EBB{} leads on
Auto and RAF, by $0.5\%$ and $4.3\%$ over TweedieGP; TweedieGP leads on
Online Retail and Carparts by $0.2\%$ and $0.8\%$. Per-series paired tests
(Table~\ref{tab:wf-significance}) cannot separate the two on Online
Retail, Carparts, or Auto, and favor \EBB{} on RAF. Over adaptive
conformal inference on ADIDA, \EBB{} gains $6.2\%$ on Online Retail and
$8.0\%$ on Carparts, beats every static conformal wrapper on all five
panels, and loses M5 by $4.4\%$.

The M5 loss is a calibration effect, and the same layer is available to
\EBB{}. ACI-\EBB{} reads \EBB{}'s predictive quantile at the adaptive
level instead of the nominal one. On M5 it lowers mean SPL from $1.4851$
to $1.3785$, $2.9\%$ below ACI-ADIDA and significant; it also leads Auto
($0.2901$) and brings Carparts to within $0.2\%$ of TweedieGP. On Online
Retail and RAF the same layer hurts, by $6.3\%$ and $2.3\%$, because the
per-series coverage signal is too noisy on those panels to improve an
already calibrated forecaster. Whether to apply it is therefore a
selection problem, and the initialization window already answers it: on
the same 80/20 chronological split used to select the discount, \EBB{} is
walked forward over the validation tail with and without the layer, and
the layer is applied iff its tail pinball is lower
(Appendix~\ref{tab:aci-rule}). The rule applies the layer on M5 and Auto
and withholds it elsewhere, matching the ex-post better variant on
4 of five panels (Carparts is the exception, where the two
are within $0.5\%$ and statistically undecided), and the tail gains it
sees track the external ones ($+7.8\%$ versus $+7.2\%$ on M5, $-7.9\%$
versus $-6.3\%$ on Online Retail). The resulting row, \EBB{} (rule), is
within 0.8\% ({Carparts}) of the best statistical or conformal method
on every panel under walk-forward updating (DeepAR is compared
separately in Table~\ref{tab:deepar}). Two further facts survive the mixed
result: under equal calibration \EBB{} beats the ADIDA base on all five
panels (undecided only on Online Retail), and the adaptive layer's benefit
is a property of the panel, not of the forecaster it wraps.

\begin{figure}[!htbp]
\centering
\includegraphics[width=0.96\textwidth]{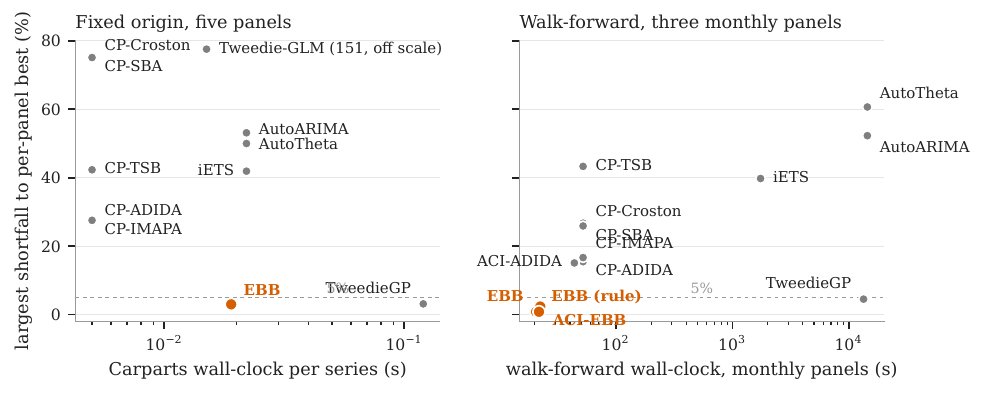}
\caption{\textbf{Largest shortfall against cost.} Each point is one
method; $y$ is its largest mean-SPL shortfall to the per-panel best
(Table~\ref{tab:regret}), $x$ is measured wall-clock on a log scale.
Left: fixed origin over five panels against Carparts cost per series
(Table~\ref{tab:efficiency}). Right: walk-forward over the three monthly
panels, on which every method has a measured wall-clock, against the total
walk-forward wall-clock on those panels (Table~\ref{tab:wfcost}; TweedieGP
uses twelve workers, every other method one process). The dashed line
marks $5\%$.}
\label{fig:frontier-full}
\end{figure}

Table~\ref{tab:regret} and Figure~\ref{fig:frontier-full} summarize both
protocols by the largest shortfall to the per-panel best among the
statistical and conformal comparators; DeepAR is not part of this
summary, and with it included the largest walk-forward shortfalls of
\EBB{} and \EBB{} (rule) are $17.6\%$ and $9.1\%$, both on M5
(Table~\ref{tab:deepar}). Among the methods of
Table~\ref{tab:regret}, \EBB{} and TweedieGP are the only ones within
five percent everywhere (\EBB{} at most $3.0\%$ at fixed origin and $4.6\%$
under walk-forward updating; TweedieGP $3.1\%$ and $4.5\%$), and every
other method is at least $15\%$ behind on some panel. Counting our own
wrapper as a competitor, uncalibrated \EBB{} trails ACI-\EBB{} by $7.7\%$
on M5, and \EBB{} (rule) closes that to 0.8\%. The two leaders differ in cost. TweedieGP refits per-series
variational GPs at every update, so its walk-forward runs took between
$36$ minutes (Auto) and $2.4$ hours (RAF) with twelve workers and project
to about $51$ hours on M5 at the Online Retail cadence; \EBB{} updates
sufficient statistics in closed form and completes each of the five
walk-forward runs in under $40$ seconds on one core. With ACI, total
wall-clock time is $52.3$ seconds on Online Retail and $217$ seconds on
M5, while the three monthly panels remain below $14$ seconds
(Table~\ref{tab:wfcost}).

\begin{table}[!htbp]
\centering
\caption{\textbf{Largest shortfall to the per-panel best.} For each
method, the maximum over panels of its mean-SPL regret to the best
non-degenerate method on that panel (percent; the panel attaining it in
parentheses; Carp.\ is Carparts), and the number of panels on which the method is within
$5\%$ of the best. A superscript gives the number of panels evaluated
when fewer than five. ACI-\EBB{} and the rule row exist only under walk-forward updating; the
rule row is excluded from the per-panel best. DeepAR is not included
(Table~\ref{tab:deepar}).
Per-panel regrets are in Table~\ref{tab:regret-full}.}
\label{tab:regret}
\small
\setlength{\tabcolsep}{2.5pt}
\begin{adjustbox}{max width=\linewidth}
\begin{tabular}{lcccc}
\toprule
 & \multicolumn{2}{c}{Fixed origin} & \multicolumn{2}{c}{Walk-forward}\\
\cmidrule(lr){2-3}\cmidrule(lr){4-5}
Method & Max (panel) & $\le5\%$ & Max (panel) & $\le5\%$\\
\midrule
AutoARIMA & 50.0 (RAF) & 0/5 & 52.3 (RAF)$^{4}$ & 0/4\\
AutoTheta & 53.1 (RAF) & 0/5 & 60.7 (RAF)$^{4}$ & 0/4\\
Tweedie-GLM & 150.8 (M5) & 0/5 & --- & ---\\
iETS & 41.9 (Carp.) & 0/5 & 39.8 (Carp.)$^{4}$ & 0/4\\
CP-Croston & 75.3 (OR) & 1/5 & 60.9 (OR) & 0/5\\
CP-SBA & 75.1 (OR) & 1/5 & 61.5 (OR) & 0/5\\
CP-TSB & 42.3 (RAF) & 0/5 & 61.1 (OR) & 0/5\\
CP-ADIDA & 27.1 (OR) & 0/5 & 35.8 (OR) & 0/5\\
CP-IMAPA & 27.5 (OR) & 0/5 & 38.1 (OR) & 0/5\\
ACI-ADIDA & --- & --- & 15.1 (RAF) & 1/5\\
TweedieGP & 3.1 (RAF) & 5/5 & 4.5 (RAF)$^{4}$ & 4/4\\
\midrule
\EBB{} & 3.0 (M5) & 5/5 & 7.7 (M5) & 4/5\\
ACI-\EBB{} & --- & --- & 6.5 (OR) & 4/5\\
\EBB{} (rule) & --- & --- & 0.8 (Carp.) & 5/5\\
\bottomrule
\end{tabular}
\end{adjustbox}
\end{table}

The configuration is selected once at the origin and then updated online;
repeated online re-selection is outside the present experiment.

\FloatBarrier
\subsection{Mechanism figure and full separation surface}

\begin{figure}[!htbp]
\centering
\includegraphics[width=0.92\textwidth]{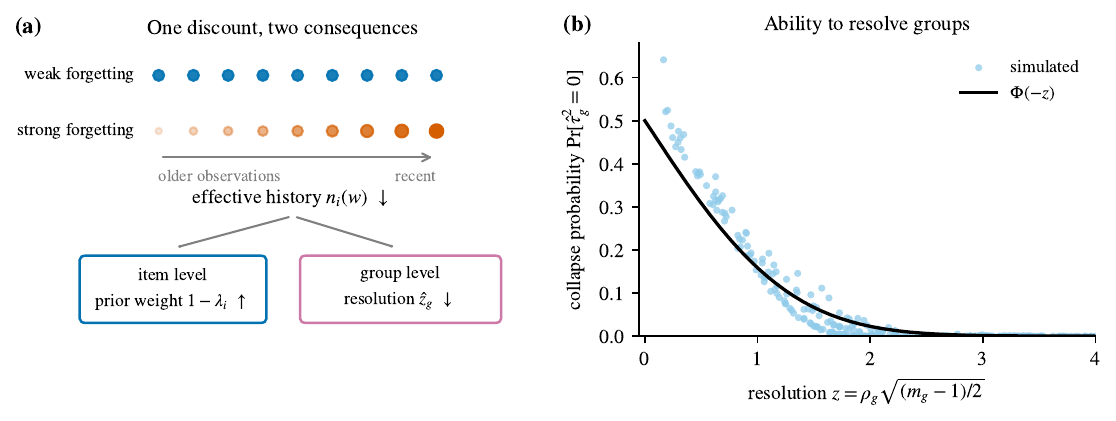}
\caption{\textbf{One discount creates two opposing effects.}
\textbf{(a)} The shared recency operator reduces item effective history,
increasing the leverage of the shared prior, while also reducing the
effective information available to estimate group-level structure.
\textbf{(b)} Under the working approximation, the probability that the
group variance estimate reaches its boundary decreases with the
population resolution coordinate
$z_g=\rho_g\sqrt{(m_g-1)/2}$; the curve is
$\Phi(-z_g)$ from \eqref{eq:collapse-prob}.}
\label{fig:mechanism}
\end{figure}

\begin{figure}[!htbp]
\centering
\includegraphics[width=0.98\textwidth]{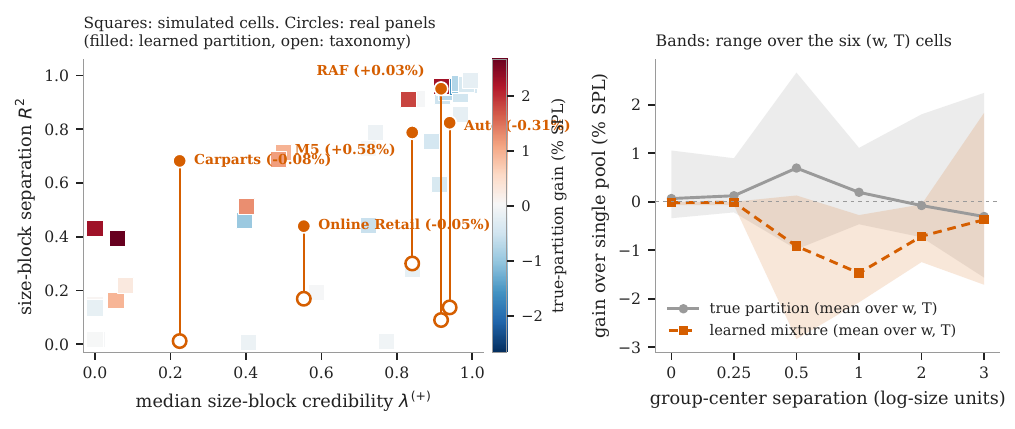}
\caption{\textbf{Refinement room depends on both leverage and separation,
and is small on four of five real panels} (full version of
Figure~\ref{fig:pays}b).
Left: each square is a controlled simulation cell, positioned by its
median size-block credibility and the share of item log-size variance
explained by the true partition ($R^2$), and colored by the true
partition's SPL gain over a single pool. Real panels are projected onto
the same coordinate system using $R^2$ from the learned partition
(filled) and the objective-free ADI/CV$^2$ taxonomy (open); labels show
the learned partition's realized refinement gain.
Right: oracle true-partition gain and learned-mixture gain versus
group-center separation, averaged over the six $(w,T)$ settings, with
the range across settings shaded. Table~\ref{tab:separation} reports the
simulation-implied oracle room corresponding to each real panel's
selected forgetting regime.}
\label{fig:separation}
\end{figure}

\FloatBarrier
\subsection{Controlled test of the resolution mechanism}

\label{sec:synthetic}

Real data never reveal the true pooling partition, so on their own they
cannot distinguish a partition that is not worth having from one we failed
to find. Both appear as a small number. We therefore construct panels
where the partition is correct by definition, and use them as a reference
against which the learned partitions of Section~\ref{sec:leverage} are
measured: what a good partition is worth at a given level of credibility
leverage, rather than what our estimator obtains there.

We simulate from the hierarchical hurdle model of \eqref{eq:model} with
group centers arranged symmetrically around a fixed grand mean, so that
increasing separation does not change the panel scale. Each cell compares
a single pool against the known true partition under the same predictive
representation, with nine paired replicates, and sweeps the two quantities
that Section~\ref{sec:resolution} identifies as controlling credibility:
the discount $w$ and the series length $T$. The resulting grid spans the
leverage range the real panels occupy, from the near-fully-shrunk regime
of Carparts to the item-dominated regime of Auto and RAF.

\begin{table}[!htbp]
\centering
\caption{Advantage of the \emph{true} partition over a single pool on
simulated panels (mean percentage improvement over nine paired replicates;
standard error in parentheses). Positive favors the partition. $T$ is
series length.}
\label{tab:sanity}
\small
\setlength{\tabcolsep}{2.5pt}
\begin{adjustbox}{max width=\linewidth}
\begin{tabular}{llrrrr}
\toprule
$T$ & $w$ & NLL & log-size MSE & pinball & MAE\\
\midrule
$30$ & $0.90$ & $2.40\;(0.18)$ & $17.15\;(2.39)$ & $2.25\;(0.61)$ & $-1.72\;(0.73)$\\
$30$ & $0.95$ & $1.12\;(0.39)$ & $9.00\;(2.45)$ & $-0.44\;(0.58)$ & $-0.17\;(0.51)$\\
$30$ & $1.00$ & $1.02\;(0.08)$ & $8.37\;(1.53)$ & $-0.69\;(0.32)$ & $-1.95\;(0.59)$\\
$120$ & $0.90$ & $1.65\;(0.22)$ & $18.36\;(1.56)$ & $1.57\;(0.64)$ & $-3.34\;(1.37)$\\
$120$ & $0.95$ & $0.96\;(0.11)$ & $3.65\;(0.77)$ & $0.23\;(0.28)$ & $-1.32\;(0.73)$\\
$120$ & $1.00$ & $0.23\;(0.04)$ & $1.13\;(0.31)$ & $-0.32\;(0.13)$ & $-0.80\;(0.11)$\\
\bottomrule
\end{tabular}
\end{adjustbox}
\end{table}

Table~\ref{tab:sanity} gives the result. The true partition improves NLL
and log-size MSE in every cell, and pinball loss whenever forgetting is
strong; in a separate length sweep its pinball advantage decays from
$1.53\%$ at $T=12$ to $0.01\%$ at $T=400$. Both patterns follow the
credibility mechanism directly. Discounting or short histories reduce an
item's own weight $\lambda^{(+)}_i$ and open room for a prior to act;
long undiscounted histories drive $\lambda^{(+)}_i\to1$ and make every
partition irrelevant. Stronger forgetting simultaneously makes group
variance components harder to estimate, as
Proposition~\ref{prop:twosided} predicts, which is why the gains in the
$w=0.90$ rows are bought against a noisier $\hat\tau_g^2$ and do not
continue to grow without limit. Extending the grid with a separation axis
(Appendix~\ref{app:simulation-details}; group centers $0$ to $3$
log-units apart) shows that the return is governed by the forgetting
regime as much as by leverage: with no separation in either block the return is within noise of
zero; with $w=0.90$ a true partition earns $1$--$2.7\%$ from separation
$0.5$ upward, with $w=0.95$ it exceeds $1\%$ only at separation $0.5$
with long histories ($\lambda^{(+)}\approx0.4$), and with $w=1$ it never
does. The learned mixture, run on the same simulated panels, is negative
at every moderate separation, and the two remedies already in the code
base---label sample-splitting and credibility shrinkage of the group
variance---do not recover the loss (grid means $-0.58\%$, $-0.79\%$, $-0.66\%$, and $-0.56\%$ combined).

MAE moves the other way in all six cells. A known-correct partition, on
data generated from the model it describes, is reported as harmful by the
metric most commonly used in this literature. We take this as the cleanest
available statement of why Section~\ref{sec:main-results} treats
absolute-error criteria as diagnostics rather than as the headline.

\paragraph{What these panels are for.}
Plotted against median $\lambda^{(+)}$, the cells of
Table~\ref{tab:sanity} and the length sweep trace the grey curves of
Figure~\ref{fig:refinement-leverage}: the return to \emph{correct}
structure as a function of leverage. Because the partition is supplied
rather than estimated, that return is an upper envelope---no procedure
that must also learn the grouping can exceed it, and any procedure that
falls below it is losing something to estimation rather than to a lack of
structure. This is the comparison Section~\ref{sec:leverage} draws, and it
is the reason the controlled study is necessary rather than decorative:
the real panels supply the leverage coordinates, and only a known-partition
experiment can supply the ordinate they should be judged against.

Two boundaries apply. The experiment isolates the value of correct
structure and says nothing about whether the mixture recovers it---indeed
Section~\ref{sec:leverage} shows it does not, and the gap between the two
is precisely what the comparison measures. And it simulates from the model
that \EBB{} fits, so the mechanism is tested under correct specification
only; a misspecified size law would shift the envelope, though it is hard
to see how it would raise it. Full diagnostics, including the growth of
$\mathrm{RMSE}(\hat\tau_g^2)$ as $w$ falls, are in
Appendix~\ref{app:simulation-details}.

\FloatBarrier
\subsection{Leverage, selection, and ablation}
\label{sec:leverage}

The results so far are consistent and small. This section asks whether
their size is a property of the data or of our estimator, and shows that
the two can be told apart with quantities the selector already computes on
the fitting head (Section~\ref{sec:select}).

\paragraph{What gets selected, and how weakly.}
All five panels select $w<1$: Carparts $0.90$, Online Retail and M5
$0.95$, Auto $0.99$, and RAF $0.997$. Across three independent
5{,}000-series M5 samples the selected configuration is identical in
structure, discount, and mixture order---$(\text{mixture},\,w{=}0.95,\,
K{=}6)$ in every case.

The pooling axis behaves differently, and the honest summary is that it
barely resolves. On three panels the spread between the best and worst
pooling structure at the selected discount is under a tenth of a percent
of validation SPL: $0.05\%$ on Online Retail, $0.08\%$ on Carparts,
$0.02\%$ on RAF. Only two panels separate at all---M5, where the learned
mixture beats the global pool by $0.58\%$, and Auto, where the global pool
beats every refinement by $0.31\%$. Announcing a winner on the other three
would report selector noise. What the selector reliably does is choose a
discount and avoid poor structural corners; it does not discover demand
taxonomies, and we do not claim it does.

\begin{table}[!htbp]
\centering
\caption{\textbf{Credibility leverage under a single pool.} Median
credibility weight per panel at no forgetting and at the selected
discount, for both blocks of the hurdle model:
$\lambda^{(o)}_i=n_i/(n_i+\phi_g)$ for occurrence and
$\lambda^{(+)}_i=n_i^{+}/(n_i^{+}+\kappa_g)$ for positive size, the
latter being the quantity bounded by Proposition~\ref{prop:twosided}.
A weight near one means the item's own history dominates its posterior,
bounding how far any shared prior can move it. ``$<0.1$'' is the share of
items whose size-block weight falls below $0.1$. The final column is the
external MAE change of a separately selected learned partition relative to
the single pool; negative favors the single pool.}
\label{tab:leverage}
\small
\setlength{\tabcolsep}{2.5pt}
\begin{adjustbox}{max width=\linewidth}
\begin{tabular}{lccccccc}
\toprule
 & & \multicolumn{3}{c}{$\lambda^{(+)}$ (size)} & \multicolumn{2}{c}{$\lambda^{(o)}$ (occ.)} & learned\\
\cmidrule(lr){3-5}\cmidrule(lr){6-7}
Panel & $w^{*}$ & $w{=}1$ & $w^{*}$ & $<0.1$ & $w{=}1$ & $w^{*}$ & vs.\ pool\\
\midrule
Carparts & $0.90$ & 0.802 & \textbf{0.224} & 22.5\% & 0.882 & 0.628 & $-1.2\%$\\
Online Retail & $0.95$ & 0.897 & \textbf{0.553} & 14.6\% & 0.969 & 0.895 & $-2.9\%$\\
M5 & $0.95$ & 0.996 & 0.841 & 12.9\% & 0.998 & 0.931 & $-2.3\%$\\
Auto & $0.99$ & 0.944 & 0.940 & 0.0\% & 0.525 & \textbf{0.422} & $-3.6\%$\\
RAF & $0.997$ & 0.926 & 0.917 & 0.0\% & 0.085 & \textbf{0.106} & $-31.5\%$\\
\bottomrule
\end{tabular}
\end{adjustbox}
\end{table}

\paragraph{Leverage is an upper bound, not a prediction.}
Table~\ref{tab:leverage} reports the diagnostic. Read as
Corollary~\ref{cor:resolution} intends, it does not say which partition
wins; it says how much any partition could win by. Two panels have almost
no room on the size block: Auto and RAF retain median
$\lambda^{(+)}\approx0.92$--$0.94$ even at their selected discounts, with
no item below $0.1$, so their sub-percent pooling differences are
explained before any partition is fitted.
The remaining three retain
leverage: Carparts is extreme---median $0.224$, with $22.5\%$ of items
essentially fully shrunk---and Online Retail sits at $0.553$ under the
selected discount. Leverage alone does not fix the prize, however. The
controlled study shows that the true partition's return at a given
$\lambda^{(+)}$ depends on the forgetting regime that produced it and on
how far the groups are apart, so the panels are placed on the full
surface below.

\begin{figure}[!htbp]
\centering
\includegraphics[width=0.98\linewidth,height=0.37\textheight,keepaspectratio]{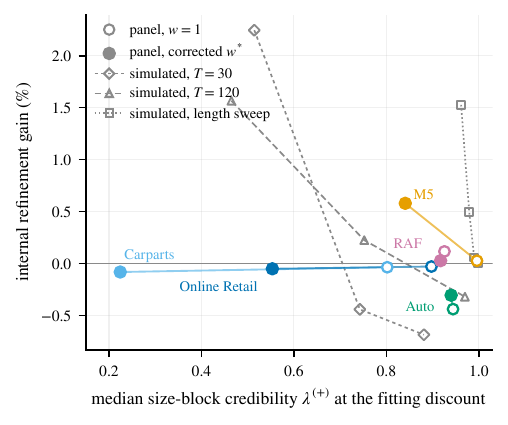}\\[0.5em]
\includegraphics[width=0.98\linewidth,height=0.37\textheight,keepaspectratio]{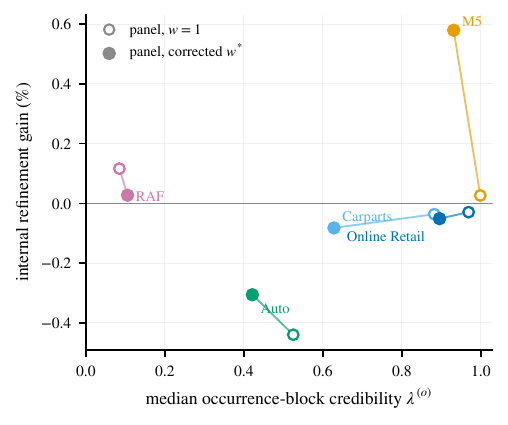}
\caption{\textbf{The simulated curves are an upper envelope that the real
panels do not reach.} Internal refinement gain (best refined structure
versus the global pool, validation SPL) against median credibility weight,
on the size block (top) and the occurrence block (bottom). Open markers
are $w{=}1$, filled markers the selected discount, and the connecting
segment is the movement forgetting induces. Grey curves are the controlled
panels of Section~\ref{sec:synthetic}, where the partition supplied is
correct by construction.
The grey curves pool the three separations of the
controlled grid; Figure~\ref{fig:separation} resolves them. Read against
the pooled envelope, every real panel lies below it at its own leverage.}
\label{fig:refinement-leverage}
\end{figure}

\paragraph{High leverage is an amplifier with no sign attached.}
The clearest illustration is RAF, which combines high size-block
credibility ($\lambda^{(+)}=0.917$) with low
occurrence credibility ($\lambda^{(o)}=0.106$), and loses $31.5\%$ of
external MAE to a learned partition. The two facts are one fact. The point
forecast is $\hat y_i=\hat\pi_i\exp(\hat\mu_i+\hat\sigma_i^2/2)$; when
$89\%$ of $\hat\pi_i$ comes from the group prior, replacing the group is
close to replacing the forecast. This is the empirical content of the
caveat attached to Proposition~\ref{prop:active}: persistent prior leverage
guarantees that the choice of target keeps mattering, not that a
particular target is good. Read together with the previous paragraph, low
leverage caps the upside and high leverage uncaps the downside, and
neither makes a learned partition worth using here.

\begin{table}[!htbp]
\centering
\caption{Online Retail distributional ablation. Each row is an
independent hierarchical run; ``auto'' denotes its own internal discount
selection. The raw predictive law is used and lower is better. Full
point metrics are in Appendix Table~\ref{tab:ablation-full}.}
\label{tab:ablation}
\small
\begin{tabular*}{\columnwidth}{@{\extracolsep{\fill}}llrrr@{}}
\toprule
Pooling & Forget & SPL & SPL$_{90}$ & AIW\\
\midrule
global & off & 0.8899 & 1.5306 & 10.51\\
taxonomy & off & 0.8897 & 1.5299 & 10.56\\
learned & off & 0.8905 & 1.5336 & 10.79\\
global & auto ($.95$) & \secondbest{0.8849} & \secondbest{1.5268} & \best{9.25}\\
taxonomy & auto ($.95$) & \best{0.8846} & \best{1.5254} & 9.43\\
learned & auto ($.95$) & 0.8869 & 1.5343 & 10.69\\
\bottomrule
\end{tabular*}
\end{table}

\paragraph{What the ablation confirms.}
Table~\ref{tab:ablation} is the external counterpart on Online Retail: the
three pooling candidates differ by less than $0.1\%$ in mean SPL without
forgetting and by less than $0.3\%$ at the selected discount, matching the
internal picture. Discounting shows up
elsewhere, narrowing the global and taxonomy $80\%$ intervals by
$11$--$12\%$ while improving mean SPL by about half a percent; the full
ablation
(Appendix Table~\ref{tab:ablation-full}) shows it improves external MAE by
$3.5$--$4.0\%$ and slightly worsens RMSE. The cross-panel result in
Table~\ref{tab:prob} should therefore be attributed to the shared
hurdle-EB predictive law and to the temporal half of the selection, not to
a learned-partition gain. That is not a concession made after the fact:
the leverage diagnostic states the ceiling in advance, and
Figure~\ref{fig:refinement-leverage}\ shows we do not reach it.

\FloatBarrier
\subsection{Pre-fit screen: protocol, metrics, and real-series transfer}
\label{app:room}

\begin{figure}[!htbp]
\centering
\includegraphics[width=0.96\textwidth]{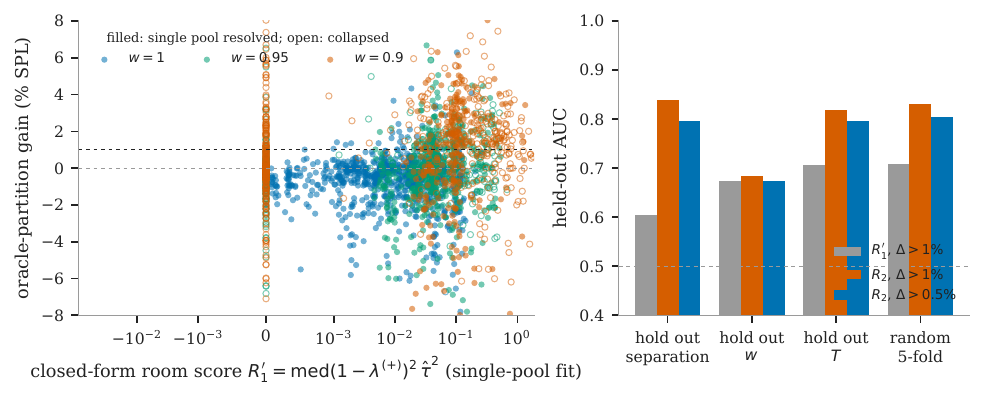}
\caption{\textbf{Screening for pooling room before any partition is
fitted.} Left: the closed-form score computed from the single-pool fit
against the oracle partition's realized gain on $2{,}700$ simulated
panels, colored by discount; open markers are panels whose single-pool
variance component collapsed. Right: held-out AUC for the event that the
oracle gain exceeds one percent (or half a percent), when whole
separation levels, discounts, or lengths are held out, or under random
folds; $R_1'$ is the closed-form score and $R_2$ a logistic screen on the
same fitting-window quantities.}
\label{fig:room}
\end{figure}

\paragraph{Protocol.} The grid crosses separation $\{0,.25,.5,1,2\}$,
discount $\{1,.95,.90\}$, length $\{30,60,120\}$, occurrence mean
$\{.1,.25,.5\}$, items per group $\{20,60\}$, and balanced versus
$70/10/10/10$ group sizes, with four groups, $\tau^2=0.15$, $\sigma^2=1$,
occurrence separation scaled as $0.8\min(\mathrm{sep},1)$, five
replicates per cell, and seed 20260915. Per panel, only the single-pool
fit is used for features: median retained leverage on the size and
occurrence blocks, $\hat\tau^2$, the Beta concentration, the resolution
statistic, effective counts, and panel size. The target is the oracle
partition's paired pinball gain over the single pool (estimated
hyperparameters, true labels), which can be negative. The pre-registered
closed-form score $R_1$ used an unweighted mean of sampling variances
that is dominated by items whose discounted positive count approaches
zero; it has no screening power (AUC $0.45$--$0.64$) . $R_1'$ replaces that term by the pool's own
weighted estimate $\hat\tau^2$; $R_2$ is a logistic regression on
$\log(1+\cdot)$ of the features, fitted on the training folds only. In $39\%$ of the simulated
panels the single-pool variance component collapses ($\hat\tau^2=0$),
and $52\%$ of the panels whose oracle gain exceeds one percent are among
them: those gains are escapes from a collapsed fit, which the resolution
statistic flags, and Table~\ref{tab:room} therefore reports the resolved
panels separately.

\begin{table}[!htbp]
\centering
\caption{Held-out screening performance on the simulated grid for the
event that the oracle partition's gain exceeds one percent (base rate
$22\%$). Columns: AUC of the pre-registered $R_1$, of $R_1'$, and of the
logistic screen $R_2$ on all panels; the same for $R_1'$ and $R_2$ on the
panels whose single-pool variance component did not collapse; precision
and recall of $R_2$ at its best-$F_1$ threshold.}
\label{tab:room}
\small
\setlength{\tabcolsep}{2.5pt}
\begin{adjustbox}{max width=\linewidth}
\begin{tabular}{lcccccc}
\toprule
 & \multicolumn{3}{c}{All panels} & \multicolumn{2}{c}{Resolved} & \\
\cmidrule(lr){2-4}\cmidrule(lr){5-6}
Held out & $R_1$ & $R_1'$ & $R_2$ & $R_1'$ & $R_2$ & $R_2$ P / R\\
\midrule
hold out a separation level & 0.45 & 0.60 & 0.84 & 0.72 & 0.85 & 0.54 / 0.77\\
hold out a discount & 0.64 & 0.67 & 0.68 & 0.66 & 0.68 & 0.51 / 0.49\\
hold out a length & 0.51 & 0.71 & 0.82 & 0.79 & 0.83 & 0.54 / 0.67\\
random 5-fold & 0.51 & 0.71 & 0.83 & 0.78 & 0.84 & 0.55 / 0.65\\
\bottomrule
\end{tabular}
\end{adjustbox}
\end{table}

\begin{figure}[!htbp]
\centering
\includegraphics[width=0.9\linewidth]{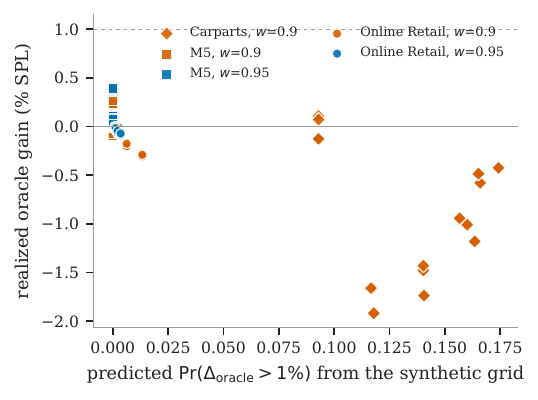}
\caption{Real-series transfer of the screen: predicted probability of
material room (from the simulated grid) against the realized oracle gain
of the injected partition, by panel and discount.}
\label{fig:room-transfer}
\end{figure}

\paragraph{Real-series transfer.} Online Retail, Carparts, and the
5{,}000-series M5 sample keep their zeros, noise, lengths, and scales; a
known partition is injected by multiplying every positive value of item
$i$ by $\exp(\delta_{g(i)})$ with symmetric offsets
$\{\pm0.5,\pm1.5\}\times\delta$, $\delta\in\{0,.25,.5,1,2\}$, over four
randomly assigned groups (three assignments), at the selected discount and
again at $w=0.90$. The oracle gain of the injected partition never exceeds
0.4\% on any panel: the single pool absorbs the injected
heterogeneity ($\hat\tau^2$ rises from $0.04$ to $5.1$ on Carparts as
$\delta$ goes from $0$ to $2$) and the median size-block credibility
rises toward one, so the items' own evidence, not the prior, carries the
forecast. $R_2$, fitted on the simulated grid and applied unchanged,
assigns every injected panel a probability below 0.17 of material
room (Figure~\ref{fig:room-transfer}). The transfer test therefore checks
the screen's rejections, not its detections; a real-series test of
detections would require panels whose items lack positive observations,
which is the short-history regime of Section~\ref{sec:pays}.

\section{Additional results}\label{app:extra}

\FloatBarrier
\subsection{Point forecasting and horizon decomposition}
\label{app:point}

Table~\ref{tab:point} gives the complete point comparison retained for
continuity with the intermittent-demand literature. Rankings exclude the
deliberately degenerate Zero and Naive-last rows. \EBB{} takes six of
ten MAE/RMSSE cells and is top two on seven, but the result is not uniform:
classical methods retain several point-metric wins, and Zero attains the
lowest MAE on Online Retail, Carparts, and RAF. This is why the main text
uses point accuracy as a secondary diagnostic.

\begin{table}[!htbp]
\centering
\caption{Point forecasting across the five panels (lower is better;
bold best and underline second among non-degenerate methods). Degenerate
rows are shown as metric diagnostics and excluded from ranking.}
\label{tab:point}
\small
\setlength{\tabcolsep}{2.5pt}
\begin{adjustbox}{max width=\linewidth}
\begin{tabular}{lcc@{\hspace{10pt}}cc@{\hspace{10pt}}cc@{\hspace{10pt}}cc@{\hspace{10pt}}cc}
\toprule
 & \multicolumn{2}{c}{Online Retail} & \multicolumn{2}{c}{M5} & \multicolumn{2}{c}{Auto} & \multicolumn{2}{c}{Carparts} & \multicolumn{2}{c}{RAF}\\
\cmidrule(lr){2-3}\cmidrule(lr){4-5}\cmidrule(lr){6-7}\cmidrule(lr){8-9}\cmidrule(lr){10-11}
Model & MAE & RMSSE & MAE & RMSSE & MAE & RMSSE & MAE & RMSSE & MAE & RMSSE\\
\midrule
CrostonClassic & 6.3294 & 5.1051 & 1.2345 & 2.3539 & 3.4056 & 1.1488 & 0.6792 & 1.0981 & 2.7315 & 1.6280\\
CrostonSBA & 6.1953 & 5.0692 & 1.2076 & 2.3517 & \secondbest{3.3325} & \secondbest{1.1451} & 0.6628 & 1.0935 & 2.6520 & 1.6272\\
TSB & \secondbest{5.5736} & 4.8031 & 1.2672 & 2.3612 & 3.6587 & 1.2066 & \best{0.5396} & 1.0156 & \best{2.1660} & 1.6570\\
TSB-tuned & 5.6318 & 4.7985 & \best{1.1932} & \secondbest{2.3137} & 3.3917 & 1.1452 & \best{0.5396} & 1.0156 & \secondbest{2.2447} & 1.6408\\
ADIDA & 5.6860 & \secondbest{4.7967} & 1.2274 & 2.3398 & 3.4079 & 1.1545 & 0.5556 & 1.0244 & 2.2977 & \secondbest{1.6245}\\
IMAPA & 5.7000 & 4.7970 & 1.2317 & 2.3280 & 3.4143 & 1.1555 & \secondbest{0.5524} & 1.0126 & 2.2850 & 1.6259\\
AutoARIMA & 6.5494 & 4.8175 & 1.2283 & 2.3169 & 3.4417 & 1.1745 & 0.5822 & \secondbest{1.0105} & 2.3153 & 1.6262\\
AutoTheta & 8.2862 & 4.8550 & 1.2988 & 2.3143 & 3.7490 & 1.2215 & 0.5704 & \best{0.9966} & 2.3687 & 1.6399\\
\midrule
Zero & 4.7549 & 4.8089 & 1.2995 & 2.4629 & 4.1444 & 1.4454 & 0.3867 & 1.0949 & 1.1717 & 1.6345\\
Naive-last & 6.0570 & 4.8324 & 1.3750 & 2.5450 & 4.3529 & 1.3818 & 0.5399 & 1.1408 & 1.8340 & 1.7644\\
\midrule
\EBB{} (ours) & \best{5.5325} & \best{4.7901} & \secondbest{1.2032} & \best{2.2938} & \best{3.2145} & \best{1.1374} & 0.5726 & 1.0120 & 2.3874 & \best{1.6245}\\
\bottomrule
\end{tabular}
\end{adjustbox}
\end{table}

\begin{figure}[!htbp]
\centering
\includegraphics[width=0.93\linewidth]{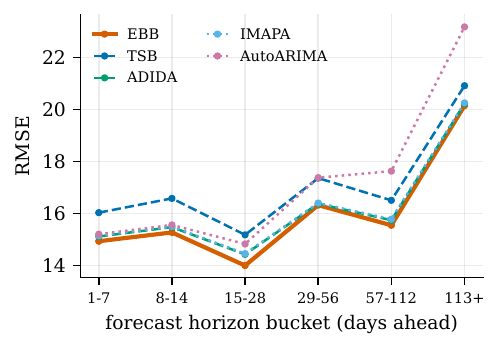}
\caption{Online Retail RMSE by fixed-origin forecast-horizon bucket.
\EBB{} is lowest in every bucket, so the aggregate result is not
created by averaging only over distant horizons. It is also lowest in
MAE in every bucket, by $0.2$--$3.2\%$ over the next method.}
\label{fig:horizon}
\end{figure}

The selected $(g,w)$ pairs are: Online Retail (global, $0.95$), M5
(mixture, $0.95$), Auto (global, $0.99$), Carparts (global, $0.90$), and
RAF (mixture, $0.997$). On three of the five panels the margins between
pooling structures on the internal validation surface are under a tenth
of a percent, so the structure component of these pairs should be read as
Section~\ref{sec:leverage} reads it: the selector avoids poor structural
corners, and no winner is announced where the surface does not resolve
one. The criterion \eqref{eq:selection-risk} weights every validation
observation equally. Averaging within items first selects the same pair
on four panels, where all validation tails have equal length; on Online
Retail, where they range from 2 to 24 days, it selects the same discount
with the taxonomy in place of the single pool, two candidates whose
validation losses differ by less than a tenth of a percent.

\FloatBarrier
\subsection{Selector diagnostics}
\label{sec:selector-diagnostics}
\label{app:selector-diagnostics}

The selection surface is locally flat near its optimum: on Online Retail
the best seven candidates span all three pooling structures and lie within
$0.51\%$ of one another in validation SPL. Section~\ref{sec:leverage}
reports the credibility leverage and refinement gains that accompany each
selected pair, and Figure~\ref{fig:refinement-leverage} places them
against the controlled panels. The selector is useful for choosing a
discount and for avoiding poor structural corners; the winning pooling
structure on a panel whose candidates lie within a tenth of a percent
should not be read as a scientific taxonomy. 

\begin{table}[!htbp]
\centering
\caption{Full Online Retail ablation corresponding to
Table~\ref{tab:ablation}. Each row is an independent run; ``auto'' denotes
its own internal discount selection. TSB has no native predictive
distribution. Lower is better.}
\label{tab:ablation-full}
\small
\setlength{\tabcolsep}{2.5pt}
\begin{adjustbox}{max width=\linewidth}
\begin{tabular}{llrrrrrr}
\toprule
Pooling & Forget & MAE & RMSE & RMSSE & SPL & SPL$_{90}$ & AIW\\
\midrule
none (TSB) & fixed & 5.5736 & 18.7272 & 4.8031 & --- & --- & ---\\
global & off & 5.7623 & \best{17.6892} & \best{4.7851} & 0.8899 & 1.5306 & 10.51\\
taxonomy & off & 5.7663 & \secondbest{17.6930} & \secondbest{4.7875} & 0.8897 & 1.5299 & 10.56\\
learned & off & 5.8126 & 17.7064 & 4.7876 & 0.8905 & 1.5336 & 10.79\\
global & auto ($.95$) & \best{5.5325} & 17.8386 & 4.7901 & \secondbest{0.8849} & \secondbest{1.5268} & \best{9.25}\\
taxonomy & auto ($.95$) & \secondbest{5.5624} & 17.8421 & 4.7897 & \best{0.8846} & \best{1.5254} & 9.43\\
learned & auto ($.95$) & 5.6925 & 17.8765 & 4.7902 & 0.8869 & 1.5343 & 10.69\\
\bottomrule
\end{tabular}
\end{adjustbox}
\end{table}

\FloatBarrier
\subsection{Walk-forward point results and protocol}
\label{app:walkforward}

\begin{table}[!htbp]
\centering
\caption{Online Retail point forecasting under strict seven-day
walk-forward updates. The hierarchical rows differ in their
initialization discount and pooling structure.}
\label{tab:walkforward}
\small
\setlength{\tabcolsep}{2.5pt}
\begin{adjustbox}{max width=\linewidth}
\begin{tabular}{lccc}
\toprule
Model & MAE & RMSE & RMSSE\\
\midrule
CrostonClassic & 5.692 & 16.937 & 4.854\\
CrostonSBA & 5.586 & 16.931 & 4.836\\
TSB & 5.434 & 17.754 & 4.815\\
ADIDA & \best{5.320} & \best{16.728} & \secondbest{4.637}\\
IMAPA & \secondbest{5.332} & \secondbest{16.745} & \best{4.632}\\
AutoARIMA & 5.482 & 16.904 & 5.105\\
AutoTheta & 5.394 & 16.836 & 4.735\\
\midrule
Stationary hierarchy ($w=1$) & 5.586 & 17.260 & 4.725\\
Forgetting hierarchy (tax., $.95$) & 5.411 & 17.059 & 4.682\\
\EBB{} (selected global, $.95$) & 5.421 & 17.188 & 4.696\\
\bottomrule
\end{tabular}
\end{adjustbox}
\end{table}

\begin{table}[!htbp]
\centering
\caption{Full Online Retail walk-forward SPL at all five quantiles and on average,
with marginal (Cov@80) and positive-demand (Cov$^{+}$@80) coverage and
average interval width (AIW). CP and ACI wrappers refit each of the 36 seven-day blocks; AutoARIMA, AutoTheta, iETS, and TweedieGP refit every four blocks; \EBB{} updates sufficient statistics each block. The zero row is excluded from ranking.}
\label{tab:wfprob-full}
\small
\setlength{\tabcolsep}{2.5pt}
\begin{adjustbox}{max width=\linewidth}
\begin{tabular}{lrrrrrrrrr}
\toprule
Model & q10 & q25 & q50 & q75 & q90 & Mean & Cov@80 & Cov$^{+}$@80 & AIW\\
\midrule
AutoARIMA & 0.259 & 0.572 & 1.197 & 1.763 & 1.522 & 1.063 & 0.933 & 0.762 & 16.58\\
AutoTheta & 0.233 & 0.584 & 1.276 & 1.717 & 1.484 & 1.059 & 0.931 & 0.765 & 16.43\\
iETS & 0.236 & 0.505 & 1.302 & 1.562 & 1.940 & 1.109 & 0.598 & 0.166 & 19.74\\
CP-Croston & 0.341 & 0.778 & 1.364 & 1.745 & 2.632 & 1.372 & 0.793 & 0.478 & 5.66\\
CP-SBA & 0.334 & 0.762 & 1.343 & 1.725 & 2.722 & 1.377 & 0.793 & 0.481 & 5.75\\
CP-TSB & 0.451 & 0.899 & 1.289 & 1.385 & 2.842 & 1.373 & 0.795 & 0.468 & 5.71\\
CP-ADIDA & 0.277 & 0.627 & 1.050 & \secondbest{1.319} & 2.514 & 1.158 & 0.791 & 0.473 & 5.47\\
CP-IMAPA & 0.292 & 0.660 & 1.083 & 1.329 & 2.522 & 1.177 & 0.791 & 0.471 & 5.47\\
ACI-ADIDA & 0.186 & 0.468 & \best{0.909} & 1.324 & 1.668 & 0.911 & 0.904 & 0.646 & 19.67\\
ACI-TSB & 0.189 & 0.482 & 0.948 & 1.384 & 1.710 & 0.943 & 0.903 & 0.641 & 21.21\\
TweedieGP & \secondbest{0.183} & \secondbest{0.458} & 0.917 & 1.326 & \best{1.379} & \best{0.853} & 0.924 & 0.709 & 13.03\\
Zero (diagnostic) & 0.183 & 0.458 & 0.916 & 1.374 & 1.649 & 0.916 & 0.740 & 0.000 & 0.00\\
\midrule
\EBB{} & 0.183 & 0.458 & \secondbest{0.910} & \best{1.319} & \secondbest{1.400} & \secondbest{0.854} & 0.901 & 0.621 & 10.31\\
ACI-\EBB{} & \best{0.183} & \best{0.458} & 0.915 & 1.423 & 1.562 & 0.908 & 0.913 & 0.665 & 19.55\\
\EBB{} (rule) & 0.183 & 0.458 & 0.910 & 1.319 & 1.400 & 0.854 & 0.901 & 0.621 & 10.31\\
\bottomrule
\end{tabular}
\end{adjustbox}
\end{table}

\begin{table}[!htbp]
\centering
\caption{Full Carparts walk-forward SPL at all five quantiles and on average,
with marginal (Cov@80) and positive-demand (Cov$^{+}$@80) coverage and
average interval width (AIW). Every baseline refits each of the six monthly blocks; \EBB{} updates sufficient statistics. The zero row is excluded from ranking.}
\label{tab:wf-carparts-full}
\small
\setlength{\tabcolsep}{2.5pt}
\begin{adjustbox}{max width=\linewidth}
\begin{tabular}{lrrrrrrrrr}
\toprule
Model & q10 & q25 & q50 & q75 & q90 & Mean & Cov@80 & Cov$^{+}$@80 & AIW\\
\midrule
AutoARIMA & 0.150 & 0.282 & 0.537 & 0.643 & 0.484 & 0.419 & 0.886 & 0.620 & 1.63\\
AutoTheta & 0.120 & 0.253 & 0.523 & 0.657 & 0.474 & 0.406 & 0.902 & 0.678 & 1.80\\
iETS & 0.272 & 0.368 & 0.493 & 0.532 & 0.515 & 0.436 & 0.442 & 0.411 & 1.17\\
CP-Croston & 0.099 & 0.245 & 0.480 & 0.614 & 0.539 & 0.395 & 0.847 & 0.580 & 1.39\\
CP-SBA & 0.098 & 0.242 & 0.475 & 0.611 & 0.538 & 0.393 & 0.847 & 0.581 & 1.37\\
CP-TSB & 0.141 & 0.301 & 0.487 & 0.513 & 0.476 & 0.384 & 0.834 & 0.551 & 1.27\\
CP-ADIDA & 0.096 & 0.230 & 0.411 & 0.529 & 0.488 & 0.351 & 0.829 & 0.593 & 1.28\\
CP-IMAPA & 0.101 & 0.235 & 0.413 & 0.518 & 0.479 & 0.349 & 0.830 & 0.581 & 1.27\\
ACI-ADIDA & 0.088 & 0.221 & 0.406 & 0.517 & 0.477 & 0.342 & 0.870 & 0.604 & 1.38\\
TweedieGP & 0.075 & 0.188 & 0.375 & 0.490 & \best{0.433} & \best{0.312} & 0.931 & 0.662 & 1.46\\
Zero (diagnostic) & 0.075 & 0.187 & 0.374 & 0.560 & 0.672 & 0.374 & 0.795 & 0.000 & 0.00\\
\midrule
\EBB{} & \secondbest{0.075} & \secondbest{0.187} & \secondbest{0.372} & \secondbest{0.481} & 0.458 & 0.315 & 0.932 & 0.669 & 1.42\\
ACI-\EBB{} & \best{0.075} & \best{0.187} & \best{0.371} & \best{0.480} & \secondbest{0.452} & \secondbest{0.313} & 0.934 & 0.676 & 1.41\\
\EBB{} (rule) & 0.075 & 0.187 & 0.372 & 0.481 & 0.458 & 0.315 & 0.932 & 0.669 & 1.42\\
\bottomrule
\end{tabular}
\end{adjustbox}
\end{table}

\begin{table}[!htbp]
\centering
\caption{Full Auto walk-forward SPL at all five quantiles and on average,
with marginal (Cov@80) and positive-demand (Cov$^{+}$@80) coverage and
average interval width (AIW). Every baseline refits each of the six monthly blocks; \EBB{} updates sufficient statistics. The zero row is excluded from ranking.}
\label{tab:wf-auto-full}
\small
\setlength{\tabcolsep}{2.5pt}
\begin{adjustbox}{max width=\linewidth}
\begin{tabular}{lrrrrrrrrr}
\toprule
Model & q10 & q25 & q50 & q75 & q90 & Mean & Cov@80 & Cov$^{+}$@80 & AIW\\
\midrule
AutoARIMA & 0.141 & 0.313 & 0.447 & 0.405 & 0.263 & 0.314 & 0.778 & 0.799 & 8.73\\
AutoTheta & 0.152 & 0.320 & 0.460 & 0.418 & 0.262 & 0.322 & 0.780 & 0.825 & 9.64\\
iETS & 0.172 & 0.312 & 0.435 & 0.423 & 0.301 & 0.329 & 0.564 & 0.690 & 7.33\\
CP-Croston & 0.148 & 0.307 & 0.437 & 0.402 & 0.296 & 0.318 & 0.801 & 0.802 & 5.97\\
CP-SBA & 0.146 & 0.304 & 0.434 & 0.404 & 0.301 & 0.318 & 0.800 & 0.803 & 6.01\\
CP-TSB & 0.155 & 0.321 & 0.458 & 0.414 & 0.301 & 0.330 & 0.808 & 0.799 & 5.97\\
CP-ADIDA & 0.149 & 0.307 & 0.437 & 0.400 & 0.295 & 0.318 & 0.802 & 0.802 & 5.92\\
CP-IMAPA & 0.149 & 0.307 & 0.437 & 0.400 & 0.295 & 0.318 & 0.802 & 0.802 & 5.92\\
ACI-ADIDA & 0.139 & 0.305 & 0.438 & 0.401 & 0.290 & 0.315 & 0.830 & 0.822 & 6.74\\
TweedieGP & 0.122 & 0.279 & \secondbest{0.418} & \best{0.389} & \best{0.254} & 0.292 & 0.834 & 0.828 & 8.81\\
Zero (diagnostic) & 0.112 & 0.281 & 0.561 & 0.842 & 1.010 & 0.561 & 0.262 & 0.000 & 0.00\\
\midrule
\EBB{} & \best{0.112} & \secondbest{0.275} & 0.418 & 0.393 & 0.256 & \secondbest{0.291} & 0.906 & 0.873 & 9.17\\
ACI-\EBB{} & \secondbest{0.113} & \best{0.275} & \best{0.416} & \secondbest{0.391} & \secondbest{0.256} & \best{0.290} & 0.905 & 0.876 & 9.41\\
\EBB{} (rule) & 0.113 & 0.275 & 0.416 & 0.391 & 0.256 & 0.290 & 0.905 & 0.876 & 9.41\\
\bottomrule
\end{tabular}
\end{adjustbox}
\end{table}

\begin{table}[!htbp]
\centering
\caption{Full RAF walk-forward SPL at all five quantiles and on average,
with marginal (Cov@80) and positive-demand (Cov$^{+}$@80) coverage and
average interval width (AIW). Every baseline refits each of the twelve monthly blocks; \EBB{} updates sufficient statistics. The zero row is excluded from ranking.}
\label{tab:wf-raf-full}
\small
\setlength{\tabcolsep}{2.5pt}
\begin{adjustbox}{max width=\linewidth}
\begin{tabular}{lrrrrrrrrr}
\toprule
Model & q10 & q25 & q50 & q75 & q90 & Mean & Cov@80 & Cov$^{+}$@80 & AIW\\
\midrule
AutoARIMA & 0.066 & 0.151 & 0.490 & 0.742 & 0.592 & 0.408 & 0.944 & 0.352 & 8.76\\
AutoTheta & 0.066 & 0.158 & 0.497 & 0.809 & 0.623 & 0.431 & 0.948 & 0.409 & 10.03\\
iETS & 0.220 & 0.270 & 0.353 & 0.435 & 0.490 & 0.354 & 0.495 & 0.019 & 0.53\\
CP-Croston & 0.090 & 0.216 & 0.381 & 0.471 & 0.492 & 0.330 & 0.828 & 0.045 & 1.03\\
CP-SBA & 0.088 & 0.212 & 0.375 & 0.468 & 0.491 & 0.327 & 0.828 & 0.041 & 0.98\\
CP-TSB & 0.160 & 0.317 & 0.462 & 0.488 & 0.494 & 0.384 & 0.836 & 0.025 & 0.37\\
CP-ADIDA & 0.078 & 0.187 & 0.343 & 0.449 & \secondbest{0.489} & 0.309 & 0.823 & 0.021 & 0.72\\
CP-IMAPA & 0.082 & 0.192 & 0.348 & 0.452 & 0.490 & 0.313 & 0.824 & 0.019 & 0.69\\
ACI-ADIDA & 0.057 & 0.154 & 0.326 & 0.446 & 0.560 & 0.308 & 0.909 & 0.083 & 1.85\\
TweedieGP & 0.053 & 0.134 & 0.270 & 0.417 & 0.526 & 0.280 & 0.935 & 0.198 & 4.07\\
Zero (diagnostic) & 0.053 & 0.134 & 0.267 & 0.400 & 0.480 & 0.267 & 0.919 & 0.000 & 0.00\\
\midrule
\EBB{} & \secondbest{0.053} & \secondbest{0.134} & \secondbest{0.267} & \best{0.400} & \best{0.486} & \best{0.268} & 0.922 & 0.041 & 0.95\\
ACI-\EBB{} & \best{0.053} & \best{0.134} & \best{0.267} & \secondbest{0.401} & 0.517 & \secondbest{0.274} & 0.927 & 0.101 & 2.07\\
\EBB{} (rule) & 0.053 & 0.134 & 0.267 & 0.400 & 0.486 & 0.268 & 0.922 & 0.041 & 0.95\\
\bottomrule
\end{tabular}
\end{adjustbox}
\end{table}

\begin{table}[!htbp]
\centering
\caption{Full M5 walk-forward SPL at all five quantiles and on average,
with marginal (Cov@80) and positive-demand (Cov$^{+}$@80) coverage and
average interval width (AIW). Static CP wrappers and ACI-ADIDA refit each of the 93 seven-day blocks on the 5{,}000-series sample; AutoARIMA, AutoTheta, iETS, and TweedieGP were not run under this protocol on cost grounds; \EBB{} updates sufficient statistics each block. The zero row is excluded from ranking.}
\label{tab:wf-m5-full}
\small
\setlength{\tabcolsep}{2.5pt}
\begin{adjustbox}{max width=\linewidth}
\begin{tabular}{lrrrrrrrrr}
\toprule
Model & q10 & q25 & q50 & q75 & q90 & Mean & Cov@80 & Cov$^{+}$@80 & AIW\\
\midrule
CP-Croston & 0.502 & 1.150 & 1.917 & 2.343 & 1.923 & 1.567 & 0.812 & 0.679 & 2.26\\
CP-SBA & 0.495 & 1.137 & 1.909 & 2.385 & 1.956 & 1.576 & 0.812 & 0.684 & 2.28\\
CP-TSB & 0.531 & 1.263 & 2.094 & 2.377 & 2.042 & 1.662 & 0.831 & 0.685 & 2.37\\
CP-ADIDA & 0.479 & 1.108 & 1.889 & 2.330 & 1.963 & 1.554 & 0.827 & 0.690 & 2.28\\
CP-IMAPA & 0.480 & 1.110 & 1.889 & 2.301 & 1.949 & 1.546 & 0.828 & 0.691 & 2.28\\
ACI-ADIDA & 0.402 & \secondbest{0.960} & \best{1.718} & \secondbest{2.211} & \secondbest{1.806} & \secondbest{1.419} & 0.883 & 0.718 & 3.48\\
Zero (diagnostic) & 0.392 & 0.980 & 1.959 & 2.938 & 3.526 & 1.959 & 0.606 & 0.000 & 0.00\\
\midrule
\EBB{} & \secondbest{0.391} & 0.967 & 1.852 & 2.336 & 1.880 & 1.485 & 0.875 & 0.690 & 2.77\\
ACI-\EBB{} & \best{0.389} & \best{0.951} & \secondbest{1.768} & \best{2.140} & \best{1.644} & \best{1.379} & 0.891 & 0.730 & 3.00\\
\EBB{} (rule) & 0.389 & 0.951 & 1.768 & 2.140 & 1.644 & 1.379 & 0.891 & 0.730 & 3.00\\
\bottomrule
\end{tabular}
\end{adjustbox}
\end{table}

\begin{table}[!htbp]
\centering
\caption{Per-panel mean-SPL regret (percent) to the best non-degenerate
method on that panel, under the fixed-origin and walk-forward protocols.
Zero regret marks the panel leader. Dashes mark methods not run under
that protocol on that panel.}
\label{tab:regret-full}
\small
\setlength{\tabcolsep}{2.5pt}
\begin{adjustbox}{max width=\linewidth}
\begin{tabular}{lrrrrrrrrrr}
\toprule
 & \multicolumn{5}{c}{Fixed origin} & \multicolumn{5}{c}{Walk-forward}\\
\cmidrule(lr){2-6}\cmidrule(lr){7-11}
Method & OR & Carparts & Auto & RAF & M5 & OR & Carparts & Auto & RAF & M5\\
\midrule
AutoARIMA & 13.8 & 27.1 & 8.6 & 50.0 & 10.4 & 24.6 & 34.4 & 8.2 & 52.3 & ---\\
AutoTheta & 13.7 & 22.1 & 15.9 & 53.1 & 10.6 & 24.2 & 30.0 & 11.1 & 60.7 & ---\\
Tweedie-GLM & 29.8 & 48.6 & 9.7 & 16.5 & 150.8 & --- & --- & --- & --- & ---\\
iETS & 11.3 & 41.9 & 17.4 & 30.4 & 25.0 & 30.1 & 39.8 & 13.3 & 31.9 & ---\\
CP-Croston & 75.3 & 24.8 & 11.0 & 24.0 & 4.2 & 60.9 & 26.7 & 9.7 & 23.1 & 13.7\\
CP-SBA & 75.1 & 23.9 & 10.9 & 22.8 & 4.4 & 61.5 & 25.9 & 9.5 & 21.9 & 14.4\\
CP-TSB & 32.8 & 17.7 & 17.8 & 42.3 & 15.8 & 61.1 & 22.9 & 13.7 & 43.3 & 20.5\\
CP-ADIDA & 27.1 & 11.1 & 11.5 & 15.6 & 5.7 & 35.8 & 12.4 & 9.4 & 15.4 & 12.7\\
CP-IMAPA & 27.5 & 10.8 & 11.6 & 16.3 & 5.9 & 38.1 & 11.9 & 9.5 & 16.6 & 12.1\\
ACI-ADIDA & --- & --- & --- & --- & --- & 6.8 & 9.5 & 8.5 & 15.1 & 3.0\\
TweedieGP & 0.0 & 0.0 & 1.6 & 3.1 & 0.0 & 0.0 & 0.0 & 0.8 & 4.5 & ---\\
\midrule
\EBB{} & 0.0 & 0.5 & 0.0 & 0.0 & 3.0 & 0.2 & 0.8 & 0.3 & 0.0 & 7.7\\
ACI-\EBB{} & --- & --- & --- & --- & --- & 6.5 & 0.2 & 0.0 & 2.3 & 0.0\\
\EBB{} (rule) & --- & --- & --- & --- & --- & 0.2 & 0.8 & 0.0 & 0.0 & 0.0\\
\bottomrule
\end{tabular}
\end{adjustbox}
\end{table}

\begin{table}[!htbp]
\centering
\caption{Short-history scaling: initialization window truncated to its
last $L$ observations per series (evaluation span unchanged; SPL scale
from the full window). Median credibility weights at the audited discount,
mean SPL of \EBB{}, TweedieGP, and CP-ADIDA, and \EBB{}'s advantage over
TweedieGP. Mixture labels are re-learned on the truncated window.}
\label{tab:coldstart}
\small
\setlength{\tabcolsep}{2.5pt}
\begin{adjustbox}{max width=\linewidth}
\begin{tabular}{llrrrrrrr}
\toprule
Panel & $L$ & median $n$ & $\lambda^{(o)}$ & $\lambda^{(+)}$ & \EBB{} & TweedieGP & CP-ADIDA & Adv.\ (\%)\\
\midrule
Online Retail & 14 & 14 & 0.823 & 0.405 & 0.8894 & 0.8920 & 1.0106 & +0.3\\
 & 28 & 28 & 0.870 & 0.506 & 0.8862 & 0.8879 & 1.1519 & +0.2\\
 & 56 & 56 & 0.893 & 0.546 & 0.8852 & 0.8859 & 1.1247 & +0.1\\
 & full & 119 & 0.895 & 0.553 & 0.8849 & 0.8845 & 1.1239 & -0.0\\
\addlinespace
M5 & 28 & 28 & 0.910 & 0.812 & 1.6925 & 1.6626 & 1.7182 & -1.8\\
 & 56 & 56 & 0.928 & 0.839 & 1.6645 & 1.6774 & 1.7082 & +0.8\\
 & 112 & 112 & 0.930 & 0.844 & 1.6697 & 1.6558 & 1.6912 & -0.8\\
 & 224 & 224 & 0.931 & 0.842 & 1.6637 & 1.5995 & 1.7092 & -4.0\\
 & 448 & 448 & 0.931 & 0.842 & 1.6670 & 1.6063 & 1.7024 & -3.8\\
 & full & 1294 & 0.931 & 0.841 & 1.6707 & 1.6228 & 1.7150 & -3.0\\
\addlinespace
Auto & 6 & 6 & 0.278 & 0.867 & 0.3009 & 0.3221 & 0.3363 & +6.6\\
 & 12 & 12 & 0.377 & 0.920 & 0.2948 & 0.3017 & 0.3317 & +2.3\\
 & full & 18 & 0.421 & 0.940 & 0.2937 & 0.2985 & 0.3275 & +1.6\\
\addlinespace
Carparts & 6 & 6 & 0.585 & 0.180 & 0.3199 & 0.3279 & 0.3858 & +2.4\\
 & 12 & 12 & 0.560 & 0.231 & 0.3197 & 0.3173 & 0.3628 & -0.8\\
 & 24 & 24 & 0.596 & 0.245 & 0.3206 & 0.3145 & 0.3585 & -1.9\\
 & full & 45 & 0.628 & 0.224 & 0.3229 & 0.3214 & 0.3571 & -0.5\\
\addlinespace
RAF & 6 & 6 & 0.037 & 0.000 & 0.2730 & 0.3079 & 0.4071 & +11.3\\
 & 12 & 12 & 0.075 & 0.682 & 0.2696 & 0.2811 & 0.3506 & +4.1\\
 & 24 & 24 & 0.135 & 0.790 & 0.2669 & 0.2788 & 0.3208 & +4.3\\
 & 48 & 48 & 0.152 & 0.891 & 0.2674 & 0.2769 & 0.3141 & +3.4\\
 & full & 72 & 0.106 & 0.917 & 0.2680 & 0.2763 & 0.3097 & +3.0\\
\bottomrule
\end{tabular}
\end{adjustbox}
\end{table}

\begin{table}[!htbp]
\centering
\caption{Within-model value of the shared prior under history truncation.
Both arms use the audited structure and discount, labels relearned on the
truncated window, and no bootstrap ensemble; ``prior off'' scales the
prior strength $(\alpha_g,\beta_g,\kappa_g)$ by $10^{-6}$, so every item
with data keeps $\lambda_i=1$ and only items without a positive
observation fall back to the group mean. Gain is the percentage by which
switching the prior off raises the loss (positive: pooling helps):
``both'' switches off both blocks, ``occ.''\ and ``size'' one block at a
time with the other left on, so the two need not sum to the total. The
last column repeats \EBB{}'s advantage over TweedieGP from
Table~\ref{tab:coldstart}. On RAF at $L=12$ the mixture labels are
relearned on twelve observations and the comparison is unstable.}
\label{tab:coldstart-nopool}
\small
\setlength{\tabcolsep}{2.5pt}
\begin{adjustbox}{max width=\linewidth}
\begin{tabular}{llrrrrrrrrr}
\toprule
 & & & & \multicolumn{2}{c}{mean SPL} & \multicolumn{3}{c}{gain from prior, mean SPL (\%)} & q90 (\%) & vs.\ TweedieGP\\
\cmidrule(lr){5-6}\cmidrule(lr){7-9}
Panel & $L$ & $\lambda^{(o)}$ & $\lambda^{(+)}$ & prior on & prior off & both & occ. & size & both & (\%)\\
\midrule
Online Retail & 14 & 0.823 & 0.405 & 0.8894 & 0.8891 & -0.0 & -0.0 & -0.2 & -0.3 & +0.3\\
 & 28 & 0.870 & 0.506 & 0.8862 & 0.8861 & -0.0 & +0.0 & -0.1 & -0.1 & +0.2\\
 & 56 & 0.893 & 0.546 & 0.8852 & 0.8851 & -0.0 & +0.0 & -0.1 & -0.0 & +0.1\\
 & full & 0.895 & 0.553 & 0.8849 & 0.8847 & -0.0 & +0.0 & -0.1 & -0.0 & -0.0\\
\addlinespace
M5 & 28 & 0.910 & 0.812 & 1.6938 & 1.6696 & -1.4 & -0.6 & -0.8 & -4.2 & -1.8\\
 & 56 & 0.928 & 0.839 & 1.6649 & 1.6631 & -0.1 & +0.4 & -0.8 & -3.4 & +0.8\\
 & 112 & 0.930 & 0.844 & 1.6718 & 1.6624 & -0.6 & +0.0 & -0.7 & -0.6 & -0.8\\
 & 224 & 0.931 & 0.842 & 1.6605 & 1.6624 & +0.1 & +0.7 & -0.7 & -2.5 & -4.0\\
 & 448 & 0.931 & 0.842 & 1.6690 & 1.6620 & -0.4 & +0.1 & -0.7 & -0.7 & -3.8\\
 & full & 0.931 & 0.841 & 1.6730 & 1.6626 & -0.6 & +0.0 & -1.0 & -1.8 & -3.0\\
\addlinespace
Auto & 6 & 0.278 & 0.867 & 0.3010 & 0.3166 & +5.2 & +4.3 & +0.3 & +1.1 & +6.6\\
 & 12 & 0.377 & 0.920 & 0.2948 & 0.3007 & +2.0 & +1.8 & -0.1 & +0.3 & +2.3\\
 & full & 0.421 & 0.940 & 0.2937 & 0.2971 & +1.2 & +1.0 & -0.1 & +0.1 & +1.6\\
\addlinespace
Carparts & 6 & 0.585 & 0.180 & 0.3199 & 0.3272 & +2.3 & +2.3 & -0.4 & +0.0 & +2.4\\
 & 12 & 0.560 & 0.231 & 0.3198 & 0.3164 & -1.1 & -0.5 & -1.0 & -4.4 & -0.8\\
 & 24 & 0.596 & 0.245 & 0.3207 & 0.3149 & -1.8 & -0.9 & -1.3 & -6.6 & -1.9\\
 & full & 0.628 & 0.224 & 0.3230 & 0.3158 & -2.2 & -1.2 & -1.5 & -7.5 & -0.5\\
\addlinespace
RAF & 6 & 0.037 & 0.000 & 0.2677 & 0.2993 & +11.8 & +14.1 & -0.1 & +24.0 & +11.3\\
 & 12 & 0.075 & 0.682 & 0.2847 & 0.2767 & -2.8 & -1.9 & -1.9 & -7.1 & +4.1\\
 & 24 & 0.135 & 0.790 & 0.2706 & 0.2720 & +0.5 & +0.9 & -0.3 & +1.3 & +4.3\\
 & 48 & 0.152 & 0.891 & 0.2671 & 0.2708 & +1.4 & +1.6 & +0.0 & +3.8 & +3.4\\
 & full & 0.106 & 0.917 & 0.2681 & 0.2700 & +0.7 & +0.8 & -0.0 & +1.9 & +3.0\\
\bottomrule
\end{tabular}
\end{adjustbox}
\end{table}

\begin{table}[!htbp]
\centering
\caption{A no-pooling candidate in the selection family. The family is extended from 24 to
32 candidates by adding \emph{none} (the model of Table~\ref{tab:coldstart-nopool} with
its prior switched off) at each of the eight discounts, scored on the same validation tail
by the same criterion. ``Validation'' is the best \emph{none} candidate's validation loss
relative to the 24-candidate winner, ``External'' the full-length mean SPL of the model
with its prior off relative to the prior on (both without the bootstrap ensemble);
negative favors \emph{none}. The selection changes on Online Retail only, where
(none, 0.95) gives $0.8847$ at fixed origin and
$0.8529$ under walk-forward against
$0.8849$ and $0.8540$ for the
configuration used in the paper. On Carparts and M5 the validation tail and the external
evaluation disagree in sign. Headline results use the 24-candidate family.}
\label{tab:none-candidate}
\small
\setlength{\tabcolsep}{2.5pt}
\begin{adjustbox}{max width=\linewidth}
\begin{tabular}{lllrrrr}
\toprule
 & \multicolumn{2}{c}{Selected configuration} & \multicolumn{2}{c}{Best \emph{none}} & \multicolumn{2}{c}{\emph{none} vs.\ pooled (\%)}\\
\cmidrule(lr){2-3}\cmidrule(lr){4-5}\cmidrule(lr){6-7}
Panel & 24 candidates & 32 candidates & $w$ & rank of 32 & Validation & External\\
\midrule
Online Retail & (global, 0.95) & (none, 0.95) & 0.95 & 1 & -0.39 & -0.02\\
M5 & (mixture, 0.95) & (mixture, 0.95) & 0.95 & 6 & +0.67 & -0.62\\
Auto & (global, 0.99) & (global, 0.99) & 0.98 & 10 & +0.74 & +1.16\\
Carparts & (global, 0.90) & (global, 0.90) & 0.90 & 3 & +0.49 & -2.23\\
RAF & (mixture, 0.997) & (mixture, 0.997) & 1.00 & 22 & +0.31 & +0.71\\
\bottomrule
\end{tabular}
\end{adjustbox}
\end{table}

\begin{figure}[!htbp]
\centering
\includegraphics[width=0.98\linewidth]{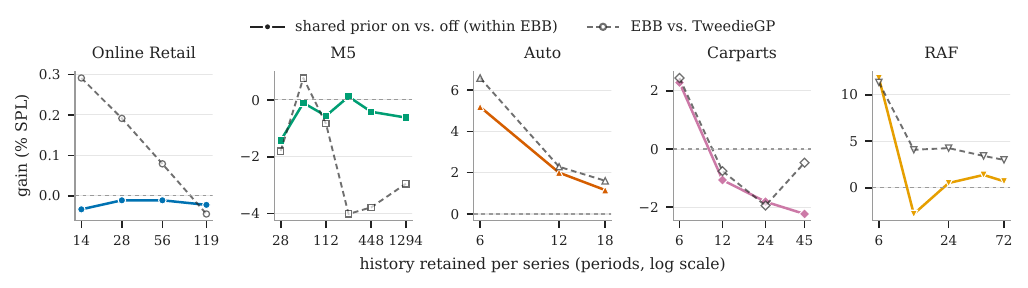}
\caption{The two readings of the short-history experiment side by side.
Solid: gain in mean SPL from the shared prior within \EBB{} (prior on
against prior off; Table~\ref{tab:coldstart-nopool}), the curve shown in
Figure~\ref{fig:pays}(a). Dashed: \EBB{}'s advantage over TweedieGP at
the same truncation (Table~\ref{tab:coldstart}). On Auto and RAF the
two move together and most of the advantage at $L=6$ is the prior's
($5.2$ of $6.6\%$ on Auto, $11.8$ against $11.3\%$ on RAF).}
\label{fig:prior-vs-tg}
\end{figure}

\EBB{} applies the selected discount once to the initialization
sufficient statistics and then accumulates revealed blocks without
compounding it. The per-quantile tables above predate the DeepAR run
and do not include it; the statements that follow are about the methods
they list. On Online Retail, ACI-ADIDA improves the static conformal
wrapper substantially and wins q50 by a hair; TweedieGP takes the mean
and q90, and \EBB{} takes q75 and is second elsewhere. On Carparts the
first two quantiles saturate at zero, \EBB{} leads from q10 through q75,
and TweedieGP takes q90 and the mean. On Auto \EBB{} leads q10, q25, and
the mean while TweedieGP leads q50 through q90; on RAF \EBB{} leads every
quantile, coinciding with the zero forecast through q75; on M5 ACI-\EBB{}
leads every quantile from q25 up and the mean, ACI-ADIDA is second, and
uncalibrated \EBB{} leads q10. ACI-\EBB{} improves on \EBB{} on M5, Auto,
and Carparts and degrades it on Online Retail and RAF
(Table~\ref{tab:wf-significance}).
Table~\ref{tab:regret-full} gives the per-panel regrets behind
Table~\ref{tab:regret}. Re-running joint selection at every block is not
evaluated.

\FloatBarrier
\subsection{DeepAR: configuration, protocol, and paired tests}
\label{app:deepar}

DeepAR \citep{salinas2020} is run from GluonTS~0.11.12 (MXNet~1.7, CPU)
with library defaults for the architecture (two LSTM layers of 40 cells,
dropout $0.1$, default lags and time features for the frequency), a
negative-binomial output, 100 epochs of 50 batches at batch size 32 and
learning rate $10^{-3}$, seed 42, and 200 sample paths per forecast. The
configuration, the protocols, and the rule that every result is reported
were fixed before any DeepAR number under these protocols was seen; nothing was tuned. One global model
is trained per panel on the initialization window. At fixed origin it
forecasts the whole evaluation span once; this is run on the three
monthly panels and not on Online Retail, whose evaluation span exceeds
the initialization window, nor on M5, where it would require a 647-step
decoder. Under walk-forward evaluation the trained model is never
refitted and forecasts each block from the history extended with
revealed targets only, the same information \EBB{} receives through its
sufficient-statistic updates. Scores use the functions and the scale
used for every other method.

\begin{figure}[!htbp]
\centering
\includegraphics[width=0.46\linewidth]{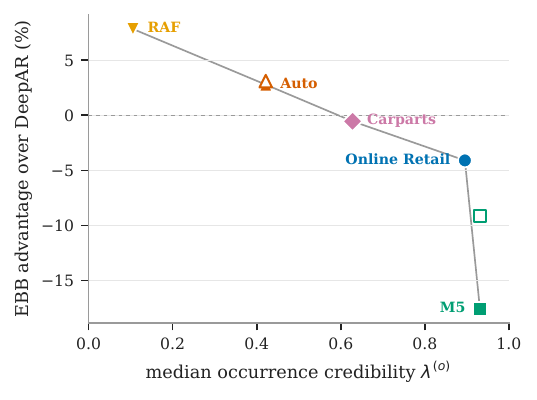}
\caption{\EBB{}'s walk-forward advantage over DeepAR against the median
occurrence credibility of each panel (values in Table~\ref{tab:deepar}).
Filled markers are uncalibrated \EBB{}; the open marker on M5 is
\EBB{} with the validation-selected ACI layer.}
\label{fig:deepar-credibility}
\end{figure}

\paragraph{The outcome against occurrence credibility.}
Table~\ref{tab:deepar} lists the panels in increasing order of median
occurrence credibility at the selected discount, the quantity reported
in Table~\ref{tab:main}. Under walk-forward evaluation \EBB{}'s advantage
over DeepAR falls monotonically along that order
(Figure~\ref{fig:deepar-credibility}), and the order is preserved
within the three monthly panels and within the two daily panels, so it
is not an artifact of sampling frequency alone. At fixed origin, run on
the monthly panels only, the signs agree with the walk-forward ones but
the two leads are not ordered: the lead on Auto is the larger. The
credibility weights were computed for Table~\ref{tab:leverage} before
DeepAR was run; the comparison with them was made afterwards and is
reported as an observation on five panels, not as a test. Every paired
decision in Table~\ref{tab:deepar} is unchanged by a Holm adjustment
over its 15 tests.

\begin{table}[!htbp]
\centering
\caption{DeepAR against \EBB{} and TweedieGP, panels in increasing order
of median occurrence credibility $\lambda^{(o)}$. $h$ is the prediction
length per forecast. Adv.\ is the mean SPL of the worse of DeepAR and
\EBB{} relative to the better, positive when \EBB{} is the better;
$p$-values are from paired per-series
$t$-tests of DeepAR against each method. Wall-clock covers training and
all forecasts on one machine without a GPU.}
\label{tab:deepar}
\small
\setlength{\tabcolsep}{2.5pt}
\begin{adjustbox}{max width=\linewidth}
\begin{tabular}{llrrrrrrr}
\toprule
 & & & & \multicolumn{2}{c}{DeepAR} & & & \\
\cmidrule(lr){5-6}
Protocol & Panel & $\lambda^{(o)}$ & $h$ & mean & q90 & \EBB{} & \EBB{} (rule) & TweedieGP\\
\midrule
Fixed & RAF & 0.11 & 12 & 0.2727 & 0.5091 & 0.2680 & --- & 0.2763\\
 & Auto & 0.42 & 6 & 0.3279 & 0.3050 & 0.2937 & --- & 0.2985\\
 & Carparts & 0.63 & 6 & 0.3169 & 0.4610 & 0.3229 & --- & 0.3214\\
\addlinespace
Walk-fwd. & RAF & 0.11 & 1 & 0.2891 & 0.5844 & 0.2681 & 0.2681 & 0.2801\\
 & Auto & 0.42 & 1 & 0.2991 & 0.2779 & 0.2910 & 0.2901 & 0.2924\\
 & Carparts & 0.63 & 1 & 0.3128 & 0.4507 & 0.3145 & 0.3145 & 0.3121\\
 & Online Ret. & 0.90 & 7 & 0.8204 & 1.3017 & 0.8540 & 0.8540 & 0.8526\\
 & M5 & 0.93 & 7 & 1.2631 & 1.3927 & 1.4851 & 1.3785 & ---\\
\bottomrule
\end{tabular}
\end{adjustbox}
\par\medskip
\begin{adjustbox}{max width=\linewidth}
\begin{tabular}{llrrrr}
\toprule
 & & \EBB{} adv. & \multicolumn{2}{c}{$p$ vs.} & Wall\\
\cmidrule(lr){4-5}
Protocol & Panel & (\%) & \EBB{} & TweedieGP & (min)\\
\midrule
Fixed & RAF & +1.8 & $<10^{-3}$ & $<10^{-3}$ & 5.5\\
 & Auto & +11.6 & $<10^{-3}$ & $<10^{-3}$ & 2.4\\
 & Carparts & -1.9 & $0.008$ & $0.274$ & 2.4\\
\addlinespace
Walk-fwd. & RAF & +7.8 & $<10^{-3}$ & $<10^{-3}$ & 2.5\\
 & Auto & +2.8 & $<10^{-3}$ & $<10^{-3}$ & 1.6\\
 & Carparts & -0.5 & $0.830$ & $0.744$ & 1.3\\
 & Online Ret. & -4.1 & $<10^{-3}$ & $<10^{-3}$ & 19.4\\
 & M5 & -17.6 & $<10^{-3}$ & --- & 67.8\\
\bottomrule
\end{tabular}
\end{adjustbox}
\end{table}

\FloatBarrier
\subsection{Three model-side remedies that do not close the gap}
\label{app:negative}

Section~\ref{sec:main-results} locates the remaining shortfall to
TweedieGP on Online Retail, Carparts, and M5. Three remedies within the
\EBB{} family are the obvious candidates: a heavier-tailed positive-size
law, an item-specific forgetting rate, and a separate forgetting rate for
the occurrence block. Both were evaluated on the
fixed-origin protocol with the audited pooling structure and, for the
size law, the identical fitted posterior; neither changes any panel's
leader.

\textbf{Predictive-law shape.} Table~\ref{tab:shape} replaces the
log-normal predictive law of \eqref{eq:predictive-law} by log-$t$,
moment-matched Gamma, and inflated-scale alternatives computed from the
same item posteriors. Mean SPL moves by at most $0.003$ on every panel and
in no case reaches TweedieGP.

\begin{table}[!htbp]
\centering
\caption{Mean SPL of alternative positive-size predictive laws computed
from the same fitted \EBB{} posterior (single fit, no bootstrap averaging),
against TweedieGP.}
\label{tab:shape}
\small
\setlength{\tabcolsep}{2.5pt}
\begin{adjustbox}{max width=\linewidth}
\begin{tabular}{lrrrrr}
\toprule
Size law & OR & Carparts & Auto & RAF & M5\\
\midrule
log-normal & 0.8849 & 0.3230 & 0.2937 & 0.2681 & 1.6730\\
log-$t$, $\nu{=}10$ & 0.8849 & 0.3228 & 0.2936 & 0.2680 & 1.6735\\
log-$t$, $\nu{=}5$ & 0.8850 & 0.3226 & 0.2940 & 0.2678 & 1.6744\\
Gamma (moment-matched) & 0.8878 & 0.3227 & 0.2937 & 0.2673 & 1.6735\\
\midrule
TweedieGP & 0.8845 & 0.3214 & 0.2985 & 0.2763 & 1.6228\\
\bottomrule
\end{tabular}
\end{adjustbox}
\end{table}

\textbf{Item-specific forgetting.} Each item receives its own discount
$w_i$, chosen on an inner chronological split of the fitting head by
minimizing $n_i^{V}R_i(w)+\kappa\bar R(w)$, where $R_i$ is the item's
validation pinball curve over the grid of Section~\ref{sec:select},
$\bar R$ the panel average, and $\kappa$ an empirical-Bayes shrinkage
strength selected on the outer validation tail;
$\kappa{=}\infty$ recovers the panel-wide $w^{\star}$ and $\kappa{=}0$
leaves each item unshrunk. Table~\ref{tab:peritem} reports the result.
The pre-registered adoption criterion (improve Online Retail or M5 by at
least $1\%$ without degrading Auto or RAF by more than $0.3\%$, and
either recover Online Retail from TweedieGP or bring M5 within $1.5\%$ of
it) fails on its second clause: M5 improves by $1.15\%$ but remains
$1.8\%$ behind, and Online Retail is unchanged. Unshrunk selection
($\kappa{=}0$) is markedly worse on M5, so the shrinkage is necessary; the
selected $\kappa$ is interior on four panels. Item-specific forgetting is evaluated only in this ablation and is
not used in the main benchmark results.

\begin{table}[!htbp]
\centering
\caption{Item-specific forgetting rates with shrinkage strength $\kappa$
selected on the outer validation tail. $w^{\star}$ is the panel-wide
discount; ``share $<w^{\star}$'' is the fraction of items that selected
faster forgetting than the panel rate; $\kappa{=}\infty$ and $\kappa{=}0$
give the fully shrunk and unshrunk endpoints. External mean SPL,
$B{=}20$.}
\label{tab:peritem}
\small
\setlength{\tabcolsep}{2.5pt}
\begin{adjustbox}{max width=\linewidth}
\begin{tabular}{lrrrrrrrrr}
\toprule
Panel & $w^{\star}$ & Scalar & $\kappa^{\star}$ & Per-item & $\Delta$ (\%) & share $<w^{\star}$ & $\kappa{=}\infty$ & $\kappa{=}0$ & TweedieGP\\
\midrule
Online Retail & 0.95 & 0.8849 & 10 & 0.8846 & -0.03 & 0.19 & 0.8849 & 0.8845 & 0.8845\\
Carparts & 0.9 & 0.3229 & $\infty$ & 0.3229 & +0.00 & 0.00 & 0.3229 & 0.3298 & 0.3214\\
Auto & 0.99 & 0.2937 & 30 & 0.2940 & +0.10 & 0.14 & 0.2942 & 0.2931 & 0.2985\\
RAF & 0.997 & 0.2680 & 100 & 0.2678 & -0.07 & 0.16 & 0.2680 & 0.2675 & 0.2763\\
M5 & 0.95 & 1.6707 & 3 & 1.6515 & -1.15 & 0.49 & 1.6751 & 1.8155 & 1.6228\\
\bottomrule
\end{tabular}
\end{adjustbox}
\end{table}

\textbf{Separate occurrence forgetting.} The M5 shortfall sits in the
occurrence block (Section~\ref{sec:main-results}), and TSB itself smooths
occurrence and size with two constants, so a third remedy gives the
occurrence counts their own discount $w^{(o)}$ while the size statistics
keep the selected $w$. $w^{(o)}$ is chosen on the same 80/20 chronological
split and criterion as $w$, from a grid that contains the tied value.
Table~\ref{tab:occdisc} reports the result. The selector keeps the tied
rate on Online Retail, Auto, and RAF, moves M5 by $0.14\%$, and moves
Carparts by $0.77\%$ (to $0.3204$, marginally ahead of TweedieGP); the
pre-registered criterion of a $1\%$ gain on M5 is not met, so the single
rate is retained. A faster occurrence rate ($w^{(o)}{=}0.8$) removes every
zero q90 forecast on M5 and lowers q90 SPL from $2.476$ to $2.452$ without
changing the mean, which locates the M5 tail problem in occurrence
persistence rather than in the size law, but the selector does not choose
it and we do not override the selector.

\begin{table}[!htbp]
\centering
\caption{Separate occurrence discount $w^{(o)}$ with the size discount
fixed at the selected $w$; $w^{(o)\star}$ is chosen on the initialization
window's validation tail. External mean SPL ($B{=}20$) for the tied and
selected rates, the share of evaluation rows whose q90 forecast is exactly
zero, and TweedieGP for reference.}
\label{tab:occdisc}
\small
\setlength{\tabcolsep}{2.5pt}
\begin{adjustbox}{max width=\linewidth}
\begin{tabular}{lccccccc}
\toprule
Panel & $w$ & $w^{(o)\star}$ & Tied & Selected & $\Delta$ (\%) & q90 $=0$ share & TweedieGP\\
\midrule
Online Retail & 0.95 & tied & 0.8849 & 0.8849 & +0.00 & 0.362 $\to$ 0.362 & 0.8845\\
M5 & 0.95 & 0.98 & 1.6707 & 1.6683 & -0.14 & 0.206 $\to$ 0.233 & 1.6228\\
Auto & 0.99 & tied & 0.2937 & 0.2937 & +0.00 & 0.000 $\to$ 0.000 & 0.2985\\
Carparts & 0.9 & 0.8 & 0.3229 & 0.3204 & -0.77 & 0.081 $\to$ 0.000 & 0.3214\\
RAF & 0.997 & tied & 0.2680 & 0.2680 & +0.00 & 0.279 $\to$ 0.279 & 0.2763\\
\bottomrule
\end{tabular}
\end{adjustbox}
\end{table}

\FloatBarrier
\subsection{Significance and interval diagnostics}
\label{sec:significance}

\begin{table}[!htbp]
\centering
\caption{Per-series mean SPL paired tests: \EBB{} versus every
comparator on each panel, Holm-adjusted within panel (paired $t$ and
Wilcoxon signed-rank; the family includes TweedieGP on every
panel). Cells count comparators by outcome at the $5\%$ level; named
entries are the comparators for which \EBB{} is \emph{not}
significantly better.}
\label{tab:spl-significance}
\small
\setlength{\tabcolsep}{2.5pt}
\begin{adjustbox}{max width=\linewidth}
\begin{tabular}{lcccl}
\toprule
Panel & Family & Sig.\ better & Sig.\ worse & Undecided\\
\midrule
Online Retail & 10 & 9 & 0 & TweedieGP\\
M5 & 10 & 5 & 1 (TweedieGP) & CP wrappers (4)\\
Auto & 10 & 10 & 0 & ---\\
Carparts & 10 & 9 & 0 & TweedieGP\\
RAF & 10 & 10 & 0 & ---\\
\bottomrule
\end{tabular}
\end{adjustbox}
\end{table}

\begin{table}[!htbp]
\centering
\caption{Walk-forward per-series paired tests, Holm-adjusted within each
(panel, reference) family; a comparison is significant when both the
paired $t$ and the Wilcoxon adjusted $p$ are below $0.05$. Families
contain every walk-forward method with stored per-series quantiles;
TweedieGP is not in the M5 family and the static conformal wrappers are
not in the Online Retail family.}
\label{tab:wf-significance}
\small
\setlength{\tabcolsep}{2.5pt}
\begin{adjustbox}{max width=\linewidth}
\begin{tabular}{llcccl}
\toprule
Panel & Reference & Family & Sig.\ better & Sig.\ worse & Undecided\\
\midrule
Online Retail & \EBB{} & 7 & 6 & 0 & TweedieGP\\
Carparts & \EBB{} & 11 & 9 & 0 & ACI-\EBB{}, TweedieGP\\
Auto & \EBB{} & 11 & 9 & 0 & ACI-\EBB{}, TweedieGP\\
RAF & \EBB{} & 11 & 11 & 0 & ---\\
M5 & \EBB{} & 7 & 5 & 2 (ACI-ADIDA, ACI-\EBB{}) & ---\\
Online Retail & ACI-\EBB{} & 7 & 4 & 2 (\EBB{}, TweedieGP) & ACI-ADIDA\\
Carparts & ACI-\EBB{} & 11 & 9 & 0 & \EBB{}, TweedieGP\\
Auto & ACI-\EBB{} & 11 & 10 & 0 & TweedieGP\\
RAF & ACI-\EBB{} & 11 & 10 & 1 (\EBB{}) & ---\\
M5 & ACI-\EBB{} & 7 & 7 & 0 & ---\\
\bottomrule
\end{tabular}
\end{adjustbox}
\end{table}

\begin{table}[!htbp]
\centering
\caption{Validation-tail rule for the calibration layer. On the 80/20
chronological split of the initialization window, \EBB{} is initialized on
the head (labels learned on the head only) and walked forward over the
tail with and without the adaptive conformal layer; the layer is applied
in deployment iff its tail scaled pinball ($q\in\{.5,.75,.9\}$) is lower.
External columns are the walk-forward means of Table~\ref{tab:wfprob};
``matches'' marks panels where the rule picks the ex-post better variant.}
\label{tab:aci-rule}
\small
\setlength{\tabcolsep}{2.5pt}
\begin{adjustbox}{max width=\linewidth}
\begin{tabular}{lrrrlrrc}
\toprule
Panel & Tail blocks & Tail plain & Tail ACI & Choice & Ext.\ \EBB{} & Ext.\ ACI-\EBB{} & Matches\\
\midrule
Online Retail & 4 & 0.7382 & 0.7967 & plain & 0.8540 & 0.9082 & yes\\
M5 & 37 & 1.5686 & 1.4457 & ACI & 1.4851 & 1.3785 & yes\\
Auto & 3 & 0.3313 & 0.3304 & ACI & 0.2910 & 0.2901 & yes\\
Carparts & 9 & 0.8975 & 0.9061 & plain & 0.3145 & 0.3128 & no\\
RAF & 14 & 0.9846 & 1.0081 & plain & 0.2681 & 0.2743 & yes\\
\bottomrule
\end{tabular}
\end{adjustbox}
\end{table}

\begin{table}[!htbp]
\centering
\caption{Paired per-series MAE tests on Online Retail: \EBB{} versus
each alternative. Negative mean differences favor \EBB{}; $p$-values
are Holm-adjusted.}
\label{tab:significance}
\small
\setlength{\tabcolsep}{2.5pt}
\begin{adjustbox}{max width=\linewidth}
\begin{tabular}{lccc}
\toprule
vs. & Mean diff & $p_t$ & $p_{\mathrm{Wilcoxon}}$\\
\midrule
Stationary hierarchy & $-0.165$ & $<10^{-29}$ & $<10^{-11}$\\
TSB & $-0.040$ & $0.532$ & $<10^{-155}$\\
TSB-tuned & $-0.090$ & $0.086$ & $<10^{-72}$\\
ADIDA & $-0.036$ & $0.359$ & $<10^{-88}$\\
IMAPA & $-0.049$ & $0.281$ & $<10^{-97}$\\
CrostonSBA & $-0.872$ & $<10^{-26}$ & $<10^{-59}$\\
AutoARIMA & $-0.891$ & $<10^{-14}$ & $<10^{-12}$\\
AutoTheta & $-2.644$ & $<10^{-81}$ & $<10^{-104}$\\
\bottomrule
\end{tabular}
\end{adjustbox}
\end{table}

\begin{table}[!htbp]
\centering
\caption{Raw \EBB{} interval quality across panels at nominal $80\%$
coverage. No post-hoc calibration is used.}
\label{tab:coverage}
\small
\setlength{\tabcolsep}{2.5pt}
\begin{adjustbox}{max width=\linewidth}
\begin{tabular}{lccccc}
\toprule
 & OR & M5 & Auto & Carparts & RAF\\
\midrule
Coverage@80 & 0.892 & 0.846 & 0.907 & 0.931 & 0.921\\
AIW@80 & 9.25 & 2.81 & 9.21 & 1.47 & 0.94\\
\bottomrule
\end{tabular}
\end{adjustbox}
\end{table}

Table~\ref{tab:spl-significance} tests the headline metric directly:
per-series mean SPL, \EBB{} against every comparator of
Table~\ref{tab:prob}, with paired $t$ and Wilcoxon signed-rank tests
Holm-adjusted within each panel and a comparison counted as decided only
when both adjusted tests agree at the $5\%$ level. \EBB{} is
significantly better in 43 of the 50 panel--comparator pairs,
significantly worse in one (TweedieGP on M5, adjusted $p=0.003$ and
$0.035$), and undecided in six. On M5 the mean difference favors
\EBB{} against all four conformal wrappers but the paired $t$ does not
reach significance after correction (the Wilcoxon strongly rejects
symmetry, so the per-series differences are systematic but
heavy-tailed); on Carparts neither \EBB{} nor TweedieGP separates from
the other (adjusted $p\ge0.50$ both ways), consistent with the
half-percent aggregate margin between them; and on Online Retail
TweedieGP is ahead by $0.04\%$ in mean SPL with the paired $t$ at
$p=0.71$. \EBB{} is significantly better than TweedieGP on Auto and
RAF.

The paired MAE tests support a systematic improvement over the stationary
hierarchy. Against TSB, tuned TSB, ADIDA, and IMAPA, mean MAE favors
\EBB{} but the adjusted paired $t$-tests are not significant; the
Wilcoxon tests reverse because those methods win on the typical series
while \EBB{} reduces the difficult-series tail. On per-series RMSE,
both adjusted tests favor \EBB{} against every comparator at $p<0.05$.
Raw intervals mildly over-cover, most strongly on Carparts.

\begin{figure}[!htbp]
\centering
\includegraphics[width=0.93\linewidth]{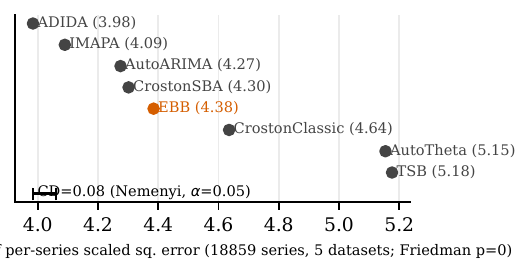}
\caption{Average ranks of per-series scaled squared errors over the
18{,}859 series with complete saved row-level point predictions across
the five panels, under the selected configurations. The critical-difference bar is shown only for scale:
series, rather than datasets, are the blocks, so this is descriptive and
not a valid Friedman--Nemenyi test.}
\label{fig:cd}
\end{figure}

\FloatBarrier
\subsection{Selection objective and efficiency}
\label{sec:criterion}\label{sec:efficiency}

The selector is not mathematically tied to pinball loss; it scores the
criterion supplied on the chronological validation tail. We use
item-averaged SPL because the target is a predictive distribution. The
best candidates are locally close, and a full cross-objective regret
matrix remains future work.

Joint selection takes $51$--$243$ seconds across the five panels,
excluding data loading, and covers 24 forecast candidates with one
deterministic EM/BIC search for the mixture on the fitting head. Final
estimation is analytic apart from low-dimensional group optimization;
selection and point prediction are deterministic, while probabilistic
bootstrap averaging uses a fixed seed.

\FloatBarrier
\subsection{Measured cost of every method}
\label{app:efficiency}

\begin{figure}[!htbp]
\centering
\includegraphics[width=0.95\linewidth]{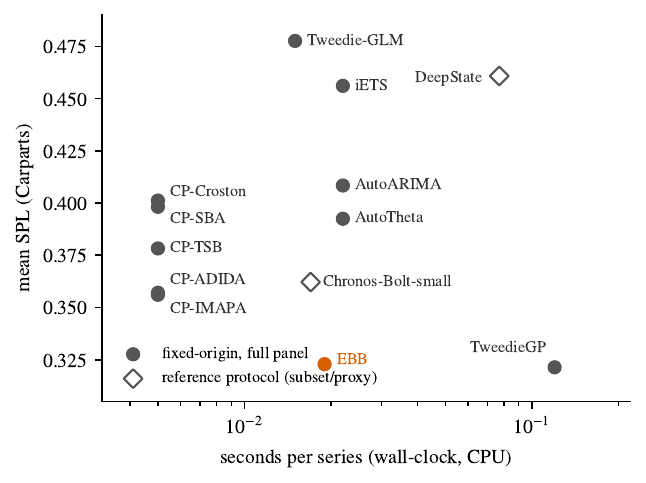}
\caption{Wall-clock cost per series against mean SPL on Carparts.
Filled circles are fixed-origin full-panel runs from
Table~\ref{tab:efficiency}; open diamonds are the reference points with
subset or proxy protocols. All methods run on CPU.}
\label{fig:cost-accuracy}
\end{figure}

Table~\ref{tab:efficiency} reports instrumented wall-clock for every
method of Table~\ref{tab:prob} on Carparts (2{,}509 series), measured on
one CPU workstation in a single session with configurations unchanged.
The \EBB{} row includes its complete 24-candidate joint selection in
addition to the final fit ($B{=}20$) and five-quantile prediction; the
TSB-tuned row likewise includes its 25-candidate selection. No method
uses a GPU. Figure~\ref{fig:cost-accuracy} plots cost against mean SPL.

On the daily panels the official TweedieGP implementation was run in full
under the authors' released defaults (sparse variational GP with 200
inducing points and log-spaced initialization for series longer than 200
observations, all training points otherwise; Adam with learning rate
$0.1$; 50{,}000 predictive samples; median-demand scaling; restart on
numerical failure) and the authors' daily period. Online Retail
(3{,}649 series) completed with no failed fits at a per-series median of
$5.1$\,s; M5 (5{,}000 series, 1{,}294 training observations) completed
with no failed fits at a per-series median of $17.3$\,s, of which roughly
half is the 50{,}000-sample predictive draw over the 647-step external
span (the original protocol forecasts 28 steps), a protocol difference
rather than a method cost. Two caveats govern every timing in
Table~\ref{tab:efficiency}. First, all rows were measured on one machine
in one session, and absolute values are not comparable across platforms.
Second, \citet{damato2025} report $0.24$\,s per Carparts series on an M3
MacBook Pro; on our machine the released defaults give $1.89$\,s (single
thread, median of twelve series) and the paper-stated budget of 100
iterations gives $0.97$\,s, because the released code doubles the
iteration cap for Tweedie likelihoods on series with at most 500
observations.

\begin{table}[!htbp]
\centering
\caption{Instrumented wall-clock on Carparts, one machine, one session.
``Sel.''\ marks totals that include the method's own hyperparameter or
candidate selection. AutoARIMA/AutoTheta share one StatsForecast call
(split across the two); the five conformal wrappers share one
residual-pool fit (split across the five). Rows below the rule are
recorded from prior run metadata rather than re-timed: DeepState is a
full-panel GluonTS/MXNet run, Chronos-Bolt is zero-shot inference on a
500-series subsample, and DRP is a 500-series Online Retail subsample
(no Carparts run exists). All rows are CPU-only.}
\label{tab:efficiency}
\small
\setlength{\tabcolsep}{2.5pt}
\begin{adjustbox}{max width=\linewidth}
\begin{tabular}{lrrccl}
\toprule
Method & Total (s) & s/series & Sel. & Runtime & Bit-repr.\\
\midrule
\EBB{} & 46.7 & 0.019 & yes & Python & Yes\\
TSB-tuned & 3.8 & 0.002 & yes & Python & No\\
CP wrappers (each) & 13.3 & 0.005 & no & Python & No\\
Tweedie-GLM & 37.3 & 0.015 & no & Python & No\\
iETS & 55.3 & 0.022 & no & R & No\\
AutoARIMA & 56.2 & 0.022 & no & Python & No\\
AutoTheta & 56.2 & 0.022 & no & Python & No\\
TweedieGP (12 workers) & 302.1 & 0.120 & no & Python/torch & No\\
\midrule
DeepState & 192.0 & 0.077 & no & Python/MXNet & No\\
Chronos-Bolt-small & 8.3 & 0.017 & no & Python/torch & No\\
DRP (Deep Renewal) & 1200.0 & 2.400 & no & Python/MXNet & No\\
\bottomrule
\end{tabular}
\end{adjustbox}
\end{table}

\begin{table}[!htbp]
\centering
\caption{Walk-forward wall-clock per method and panel (seconds unless
marked in hours), covering every block of the protocol of
Table~\ref{tab:wfprob}: initialization, all per-block refits or updates,
and prediction. StatsForecast, conformal, and ACI rows run
single-process; TweedieGP runs its released code with twelve worker
processes; \EBB{} runs on one core and includes its initialization.
AutoARIMA and AutoTheta share one call. $^{\dagger}$Projected from the
measured $7{,}618$\,s full fit on the 5{,}000-series M5 sample at the
Online Retail cadence of one refit per four blocks ($24$ refits); at one
refit per block the projection is about $197$\,h. Neither was run.}
\label{tab:wfcost}
\small
\setlength{\tabcolsep}{2.5pt}
\begin{adjustbox}{max width=\linewidth}
\begin{tabular}{lrrrrr}
\toprule
Method & Online Retail & Carparts & Auto & RAF & M5\\
\midrule
AutoARIMA & 1169 & 1.1\,h & 763 & 2.7\,h & ---\\
AutoTheta & 1169 & 1.1\,h & 763 & 2.7\,h & ---\\
iETS & 1795 & 270 & 213 & 1254 & ---\\
CP-ADIDA & --- & 9.6 & 4.4 & 38.0 & 465\\
ACI-ADIDA & 126 & 7.4 & 8.9 & 27.3 & 581\\
TweedieGP & 1.7\,h & 2686 & 2139 & 2.4\,h & $\approx$51\,h$^{\dagger}$\\
\midrule
\EBB{} & 22.4 & 3.4 & 3.6 & 13.5 & 36.6\\
ACI-\EBB{} & 52.3 & 4.2 & 4.8 & 13.3 & 217\\
\EBB{} (rule) & 22.4 & 3.4 & 4.8 & 13.5 & 217\\
\bottomrule
\end{tabular}
\end{adjustbox}
\end{table}

\FloatBarrier
\subsection{Controlled-study diagnostics}
\label{app:simulation-details}

In the length sweep used to isolate credibility leverage, the generator
holds $K=2$, 500 items per group, occurrence probabilities $0.05$ and
$0.80$, log-size centers three units apart, and $w=1$. The true
partition's pinball advantage falls monotonically from $1.53\%$ at
$T=12$ (median 4.5 positives per item) to $0.01\%$ at $T=400$ (median
154), with NLL, Brier, and log-size MSE following the same direction.
Across the main grid, the RMSE of $\hat\tau_g^2$ approximately doubles
as $w$ falls from $1.00$ to $0.90$ ($0.037\to0.144$ at $T=120$ and
$0.074\to0.149$ at $T=30$), directly measuring the resolution cost in
Proposition~\ref{prop:twosided}.

The generator centers the groups symmetrically around a fixed grand mean
so that increasing separation does not change total demand scale, and
every arm is evaluated through the same hurdle-lognormal predictive
representation under both likelihood-based proper scores and point
metrics. A four-arm ladder---single pool, true labels with estimated
hyperparameters, true labels with true hyperparameters, and the true
conditional law---checks the implementation; the true conditional law is
best on every metric in every cell.

\paragraph{Separation axis.} The grid above pools the three separations
$\{0.5,1,2\}$ of the original sweep, and the length sweep uses a single
separation of three; the envelope of Figure~\ref{fig:refinement-leverage}
therefore mixes separations. To resolve them we rerun the same generator
on a pre-registered grid: separation $\in\{0,0.25,0.5,1,2,3\}$ log-units
between adjacent group centers, $w\in\{1,0.95,0.90\}$, $T\in\{30,120\}$,
nine paired replicates per cell (seed 20260913), four groups of sixty
items, $\tau^2=0.15$, $\sigma^2=1$, occurrence spread in logit scaled
with the size separation as $0.8\min(\mathrm{sep},1)$ so that zero
separation removes structure from both blocks (a first pass held it fixed
at $0.8$; it is archived and superseded). Each cell records the true partition's paired pinball gain over a
single pool, the gain of the paper's mixture objective run on the same
panel, the median size-block credibility under the single pool, and the
share of item log-size variance explained by the true labels ($R^2$,
computed from the discounted item statistics so it includes sampling
noise). Two remedies for the learned mixture that are implemented are run as extra arms on the same panels: label sample-splitting
(labels from one interleaved half, hyperparameters from the other) and
credibility shrinkage of the group variance component, alone and
combined. Table~\ref{tab:sepsim} summarizes
the surface by separation and Table~\ref{tab:separation} reads it at the
real panels' coordinates. Real-panel $R^2$ uses the same statistic under
the learned mixture labels at the selected discount and under the
ADI/CV$^2$ taxonomy; we do not use the learned $\hat\tau_g^2$ as a scale
because it collapses to its floor under learned partitions
(Section~\ref{sec:leverage}). The simulation was run and summarized
before the real panels were placed.

\begin{table}[!htbp]
\centering
\caption{Separation-resolved controlled study, averaged over the six
$(w,T)$ cells at each separation: true-label $R^2$, the true partition's
gain over a single pool, the learned mixture's gain, the same with label
sample-splitting and credibility shrinkage combined, and the median
number of components the mixture selects.}
\label{tab:sepsim}
\small
\setlength{\tabcolsep}{2.5pt}
\begin{adjustbox}{max width=\linewidth}
\begin{tabular}{lrrrrr}
\toprule
Separation & $R^2$ & True (\%) & Learned (\%) & Learned+fix (\%) & $K$\\
\midrule
0 & 0.01 & +0.07 & -0.02 & -0.36 & 1\\
0.25 & 0.19 & +0.12 & -0.01 & -0.30 & 1\\
0.5 & 0.47 & +0.70 & -0.92 & -0.94 & 2\\
1 & 0.75 & +0.20 & -1.47 & -0.86 & 2\\
2 & 0.93 & -0.08 & -0.71 & -0.37 & 3\\
3 & 0.97 & -0.30 & -0.37 & -0.51 & 4\\
\bottomrule
\end{tabular}
\end{adjustbox}
\end{table}

\begin{table}[!htbp]
\centering
\caption{Real panels placed on the separation-resolved surface. $w$ is the
simulated discount matching the panel's selected one; $R^2$ is the
size-block separation under the learned partition and under the taxonomy,
and ``family $R^2$'' the true-label $R^2$ of the simulated family at the
panel's $\lambda^{(+)}$ ($T{=}120$). ``Room'' is the true partition's gain
interpolated along $\lambda^{(+)}$ within the matching $w$ at each length;
``sim.\ learned'' the learned mixture's gain at the same point
($T{=}120$); ``realized'' the panel's internal refinement gain
(Figure~\ref{fig:refinement-leverage}).}
\label{tab:separation}
\small
\setlength{\tabcolsep}{2.5pt}
\begin{adjustbox}{max width=\linewidth}
\begin{tabular}{lrrrrrrrrr}
\toprule
Panel & $w$ & $\lambda^{(+)}$ & $R^2$ learned & $R^2$ tax. & family $R^2$ & Room $T{=}30$ & Room $T{=}120$ & Sim.\ learned & Realized\\
\midrule
Online Retail & 0.95 & 0.553 & 0.44 & 0.17 & 0.63 & -0.52 & +0.63 & -0.87 & -0.05\\
M5 & 0.95 & 0.841 & 0.79 & 0.30 & 0.86 & -0.34 & -0.45 & -1.21 & +0.58\\
Auto & 1 & 0.940 & 0.82 & 0.14 & 0.72 & -0.50 & -0.33 & -0.45 & -0.31\\
Carparts & 0.9 & 0.224 & 0.68 & 0.01 & 0.55 & +2.01 & +1.73 & -0.75 & -0.08\\
RAF & 1 & 0.917 & 0.95 & 0.09 & 0.62 & -0.48 & -0.41 & -0.58 & +0.03\\
\bottomrule
\end{tabular}
\end{adjustbox}
\end{table}

\FloatBarrier
\subsection{MASE}
For completeness we report MASE (per-series MAE scaled by in-sample
naive MAE), the intermittent-demand literature's traditional point
metric. It inherits the absolute-error pathology quantified in
Section~\ref{sec:main-results}: the all-zero forecast attains the best
MASE on Online Retail ($1.84$ vs.\ TSB $2.08$, \EBB{} $2.41$),
Carparts, and RAF, and zero-leaning methods (TSB, IMAPA) lead among real
forecasters. On Auto, \EBB{} ($0.889$) is second only to CrostonSBA
($0.883$). We treat MASE as descriptive, not
decision-relevant, for the same reason as MAE.

\FloatBarrier
\subsection{Walk-forward results}
The complete point and probabilistic results are in
Tables~\ref{tab:walkforward} and \ref{tab:wfprob}.

\FloatBarrier
\subsection{Neural baselines}
\label{app:neural}

Table~\ref{tab:neural} retains the earlier neural comparisons. Deep
Renewal Processes (DRP) \citep{turkmen2019} are architecturally close to
our occurrence--size decomposition, but the legacy subsample numbers are
indicative only: we rebuilt the Python~3.8/MXNet~1.7 environment and
launched the original code path, yet three configurations diverged to
NaN between epochs 7 and 11 and failed during prediction. DeepAR and an
RNN under prediction-only chaining reach lower MAE in one case but worse
RMSE and RMSSE at both horizons. Neural decoding and a constant
sufficient-statistic forecast are not timing-equivalent, so these rows
should not support a broad efficiency ratio.

Two newly executed reference points provide a cleaner, narrower check.
DeepState---the published Amazon ISSM successor available in GluonTS,
not a reproduction of \citet{seeger2016}---obtains mean SPL $1.0652$ on
a 1{,}000-series Online Retail subset and $0.4608$ on full Carparts.
Chronos-Bolt-small obtains $0.8448$ on an Online Retail 500-series subset,
versus $0.8342$ for \EBB{} on exactly the same subset: a small $1.3\%$
gap. Its Carparts result is only a reference point because the same-source
Monash panel appears in the Chronos data repository and potential training
contamination cannot be excluded. These subset results motivate an
efficiency, determinism, and transparency claim, not a large foundation-
model accuracy advantage.

\begin{table}[!htbp]
\centering
\caption{Neural comparisons on Online Retail. Top: probabilistic
(unscaled pinball, 500-series subsample, fixed protocol). Bottom: point
metrics under prediction-only chaining at two horizons.}
\label{tab:neural}
\small
\setlength{\tabcolsep}{2.5pt}
\begin{adjustbox}{max width=\linewidth}
\begin{tabular}{lcccccc}
\toprule
 & q10 & q25 & q50 & q75 & q90 & Time\\
\midrule
DeepRenewal & 0.358 & 0.895 & 1.790 & 2.684 & 2.941 & $\sim$20\,min\\
Hierarchical EB & 0.387 & 0.914 & 1.772 & 2.390 & 2.186 & 1\,s\\
\midrule
 & \multicolumn{2}{c}{MAE} & \multicolumn{2}{c}{RMSE} & \multicolumn{2}{c}{RMSSE}\\
 & $h{=}10$ & $h{=}100$ & $h{=}10$ & $h{=}100$ & $h{=}10$ & $h{=}100$\\
\midrule
RNN & 3.257 & 3.132 & 11.938 & 12.538 & 0.875 & 8.197\\
DeepAR & 6.245 & 7.288 & 17.221 & 18.623 & 7.151 & 9.986\\
Hierarchical EB & 4.639 & 4.495 & 11.284 & 11.733 & 0.855 & 8.170\\
\bottomrule
\end{tabular}
\end{adjustbox}
\end{table}

\FloatBarrier
\subsection{Far-tail quantiles on the sparsest panels}
\label{app:fartail}

Scaled pinball at $q\in\{0.95,0.975,0.99\}$ on the two panels where the
standard levels discriminate weakly, under the identical fixed protocol
(\EBB{} raw, $B{=}20$; iETS not re-run at these levels). The zero
forecast's scaled pinball is \emph{linear} in $q$ (its pinball at level
$q$ is $q\,\E[y/\mathrm{scale}]$), so the far tail is exactly where its
apparent advantage must collapse; the question the table answers is who
inherits the lead.

\begin{table}[!htbp]
\centering
\small
\setlength{\tabcolsep}{2.5pt}
\begin{adjustbox}{max width=\linewidth}
\begin{tabular}{lccc@{\hspace{10pt}}ccc}
\toprule
 & \multicolumn{3}{c}{Carparts} & \multicolumn{3}{c}{RAF}\\
\cmidrule(lr){2-4}\cmidrule(lr){5-7}
Model & q95 & q97.5 & q99 & q95 & q97.5 & q99\\
\midrule
CP-Croston & 0.413 & 0.290 & 0.168 & 0.488 & 0.537 & 0.526\\
CP-SBA & 0.412 & 0.290 & 0.168 & 0.489 & 0.542 & 0.529\\
CP-TSB & 0.374 & 0.263 & 0.160 & 0.506 & 0.601 & 0.552\\
CP-ADIDA & 0.384 & 0.275 & 0.159 & 0.493 & 0.579 & 0.557\\
CP-IMAPA & 0.378 & 0.270 & 0.156 & 0.493 & 0.581 & 0.564\\
AutoARIMA & 0.385 & 0.332 & 0.301 & \best{0.460} & 0.395 & 0.356\\
AutoTheta & 0.367 & 0.316 & 0.285 & 0.464 & 0.399 & 0.359\\
Tweedie-GLM & 0.581 & 0.445 & 0.305 & 0.497 & 0.439 & 0.367\\
TweedieGP & \best{0.355} & \best{0.222} & \best{0.128} & 0.462 & \best{0.352} & \secondbest{0.247}\\
\midrule
Zero & 0.710 & 0.729 & 0.740 & 0.507 & 0.520 & 0.529\\
\midrule
\EBB{} & \secondbest{0.365} & \secondbest{0.251} & \secondbest{0.144} & \secondbest{0.461} & \secondbest{0.364} & \best{0.246}\\
\bottomrule
\end{tabular}
\end{adjustbox}
\end{table}

Three readings follow. First, \textbf{the saturation breaks as expected}:
on RAF the zero forecast, marginally ahead at $q\le0.9$, falls behind
every intensity-bearing method from $q{=}0.95$ on, and on Carparts it is
last by a factor of two. Second, \textbf{tail leadership is
model-specific}: official TweedieGP wins all three Carparts cells, while
\EBB{} is second in all three; on RAF, AutoARIMA wins q95, TweedieGP
wins q97.5, and \EBB{} wins q99 by a near tie. Third, this table
deliberately narrows the claim: \EBB{}'s most consistent advantage is
aggregate SPL, not universal extreme-quantile dominance. The conformal
wrappers still deteriorate in the RAF far tail, where their per-level
calibration information is sparse.

\FloatBarrier
\subsection{Full-posterior inference on the same hierarchy}
\label{app:mcmc}

To separate the inference method from the model, we paired plug-in EB
with a full-posterior sampler at each discount. Each pair uses the same
single global pool, shared size-variance specification, discounted
sufficient statistics, seed-controlled 500-series Online Retail subsample,
evaluation span, and scoring code. The sampler uses
Metropolis-within-Gibbs for the Beta hyperparameters, four deterministic
chains of 4{,}000 sweeps, 2{,}000 burn-in iterations, and thinning by four.
Both a stationary version ($w{=}1$) and a power-prior discounted version
($w{=}0.98$) are tested. The richer \EBB{} arm runs its joint
selection separately on this subsample and is
reported only as context, not as an inference-method contrast.

\begin{table}[!htbp]
\centering
\small
\begin{adjustbox}{max width=\linewidth}
\begin{tabular}{lccccc}
\toprule
 & SPL mean & SPL q90 & MAE & RMSSE & sec.\\
\midrule
\EBB{} (incl.\ selection) & 0.7942 & 1.3682 & 5.6259 & 4.6712 & 86.4\\
EB global, $w{=}1$ & 0.7919 & 1.3588 & 5.5504 & 4.6684 & 0.2\\
Gibbs global, $w{=}1$ & 0.7919 & 1.3590 & 5.5890 & 4.6696 & 2.7\\
EB global, $w{=}0.98$ & 0.7935 & 1.3658 & 5.4774 & 4.6694 & 0.2\\
Gibbs global, $w{=}0.98$ & 0.7937 & 1.3665 & 5.5549 & 4.6734 & 2.7\\
\bottomrule
\end{tabular}
\end{adjustbox}
\end{table}

Two readings. First, \emph{the empirical-Bayes approximation is not a
distributional-accuracy bottleneck}: paired full-posterior inference changes
mean SPL by $0.007\%/0.019\%$, q90 SPL by $0.018\%/0.048\%$, and RMSSE by
$0.025\%/0.085\%$ at $w=1/0.98$. MAE is somewhat more sensitive but still
changes only $0.70\%/1.42\%$. Second, forgetting helps MAE under both
inference methods, so that finding is not an artifact of the EB plug-in.
The maximum split-$\widehat R$ across $(\alpha,\beta,\mu_0,\tau^2,\sigma^2)$
is $1.0021$, and the minimum effective sample size is $819$ of 2{,}000
retained draws.
What this experiment does \emph{not} measure: the richer external
state-space samplers
\citep{chapados2014,seeger2016} remain unrun
(Section~\ref{sec:limitations}); this
conjugate single-pool Gibbs is far cheaper than those models and is a
control for the inference method, not a stand-in for them. The plug-in
EB path retains bit-reproducibility and requires no convergence
diagnostics, but at this scale raw speed does not separate the two.

\FloatBarrier
\subsection{Datasets}\label{app:datasets}
Online Retail: UCI 352; returns/zero-price removed, 99.5\% quantity cap,
daily aggregation densified with zeros. M5: Kaggle M5-Accuracy;
seed-controlled 5{,}000-series sample. Both M5 protocols use a two-thirds
initialization split; the learned mixture is selected at $w=0.95$ on all
three audited seeds. Auto/RAF: gluon-ts
\texttt{intermittent-datasets} branch. Carparts: Zenodo 4656022, series
without missing values and with at least one zero (2{,}509). Monthly
panels use the last-$h$ holdout of \citet{damato2025}.

\end{document}